\documentclass[10pt,twocolumn,letterpaper]{article}

\usepackage{cvpr}
\usepackage{times}
\usepackage{epsfig}
\usepackage{graphicx}
\usepackage{amsmath}
\usepackage{amssymb}
\usepackage{comment}
\usepackage{color, colortbl}
\usepackage{booktabs}
\usepackage{multirow}
\usepackage{footnote}
\usepackage{xr}
\usepackage{xspace}
\usepackage{caption}
\usepackage{balance}
\definecolor{Gray}{gray}{0.9}

\usepackage[pagebackref=true,breaklinks=true,letterpaper=true,colorlinks,bookmarks=false]{hyperref}

\cvprfinalcopy 

\def\cvprPaperID{6109} 

\ifcvprfinal\fi

\begin{document}
\newcommand{\dbname}{AGORA\xspace}
\newcommand{\scan}{\mathcal{S}}
\newcommand{\numScan}{4240}
\newcommand{\numSub}{350}
\newcommand{{\numval}}{1225}
\newcommand{{\numvalcrop}}{11589}
\newcommand{\numtest}{3387}
\newcommand{\numtestcrop}{30701}
\newcommand{\numtrain}{14529}
\newcommand{\numtraincrop}{131631}
\newcommand{\numtrainscans}{2930}
\newcommand{\numtestscans}{1051}
\newcommand{\numvalscans}{259}
\newcommand{\numBFH}{3161}
\newcommand{\numBody}{1079}
\newcommand{\numtestsub}{105}
\newcommand{\skinerrthresh}{5}
\newcommand{\model}{\mathcal{M}}
\newcommand{\modelpoints}{M}
\newcommand{\clothterm}{E_{\text{cloth}}}
\newcommand{\skinterm}{E_{\text{skin}}}
\newcommand{\dist}{\text{dist}}
\newcommand{\scanpoints}{S}
\newcommand{\newclaim}{\textcolor{blue}}
\newcommand{\updatelater}{\textcolor{red}}
\renewcommand{\etal}{et al.\xspace}
\renewcommand{\ie}{i.e.\xspace}
\renewcommand{\eg}{e.g.\xspace}
\newcommand{\phuang}[1]{\textcolor{red}{[PH: {#1}]}}

\title{{\dbname}: Avatars in Geography Optimized for Regression Analysis}

\author{Priyanka Patel$^1$ \quad
Chun-Hao P. Huang$^1$ \quad
Joachim Tesch$^1$ \quad
David T. Hoffmann$^{2,3}$\thanks{} \quad
\and
Shashank Tripathi$^1$ \quad
Michael J. Black$^1$\\
$^1$Max Planck Institute for Intelligent Systems, T\"ubingen, Germany\\
$^2$University of Freiburg \quad $^3$Bosch Center for Artificial Intelligence\\
{\tt\small \{ppatel, paul.huang, jtesch, dhoffmann, stripathi, black\}@tuebingen.mpg.de}
}

\twocolumn[
{
    \renewcommand\twocolumn[1][]{#1}
   \maketitle
   \thispagestyle{empty}
    \centering
    \vspace{-2ex}
 \begin{minipage}{1.00\textwidth}
	\centerline{\includegraphics[width=0.23\textwidth]{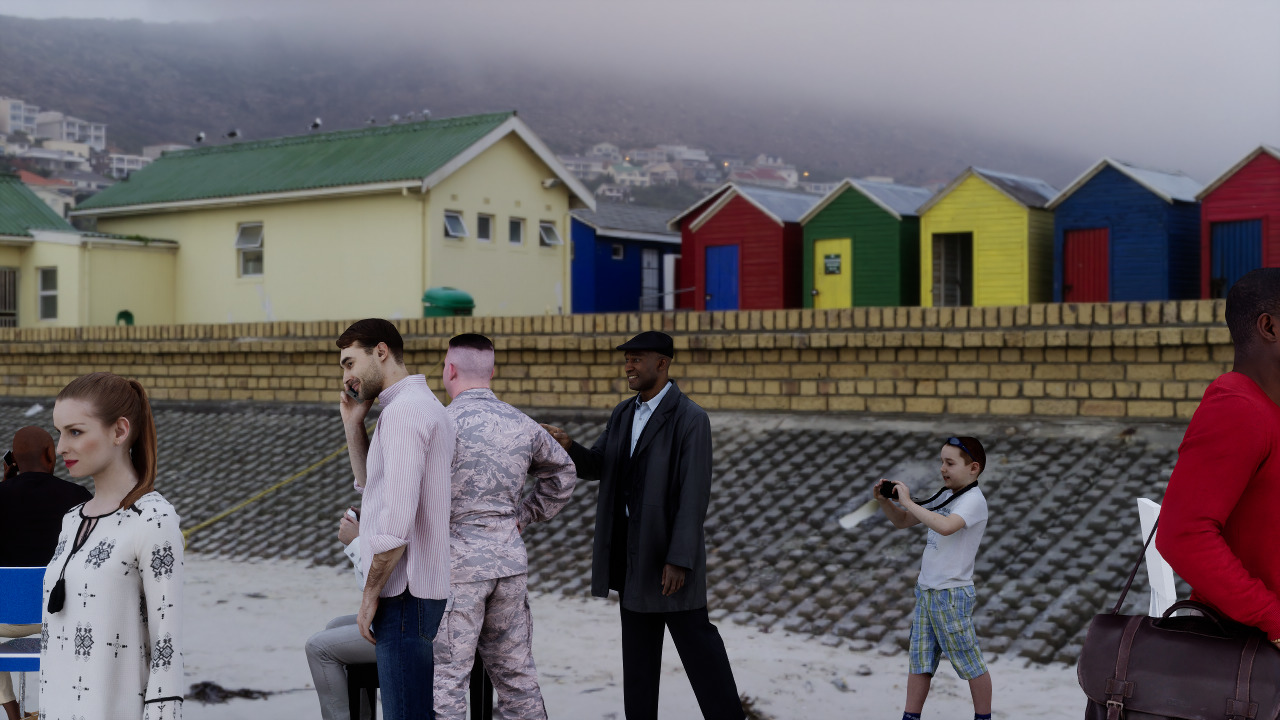}	\includegraphics[width=0.23\textwidth]{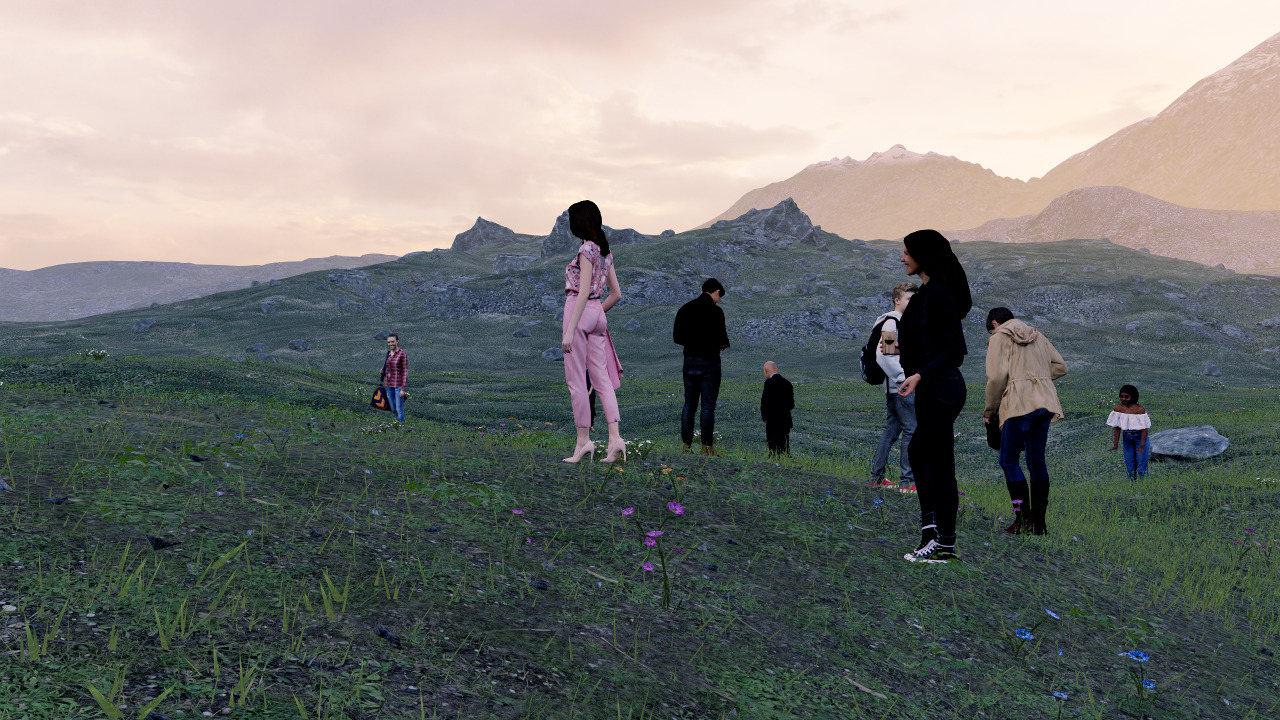}
	\includegraphics[width=0.23\textwidth]{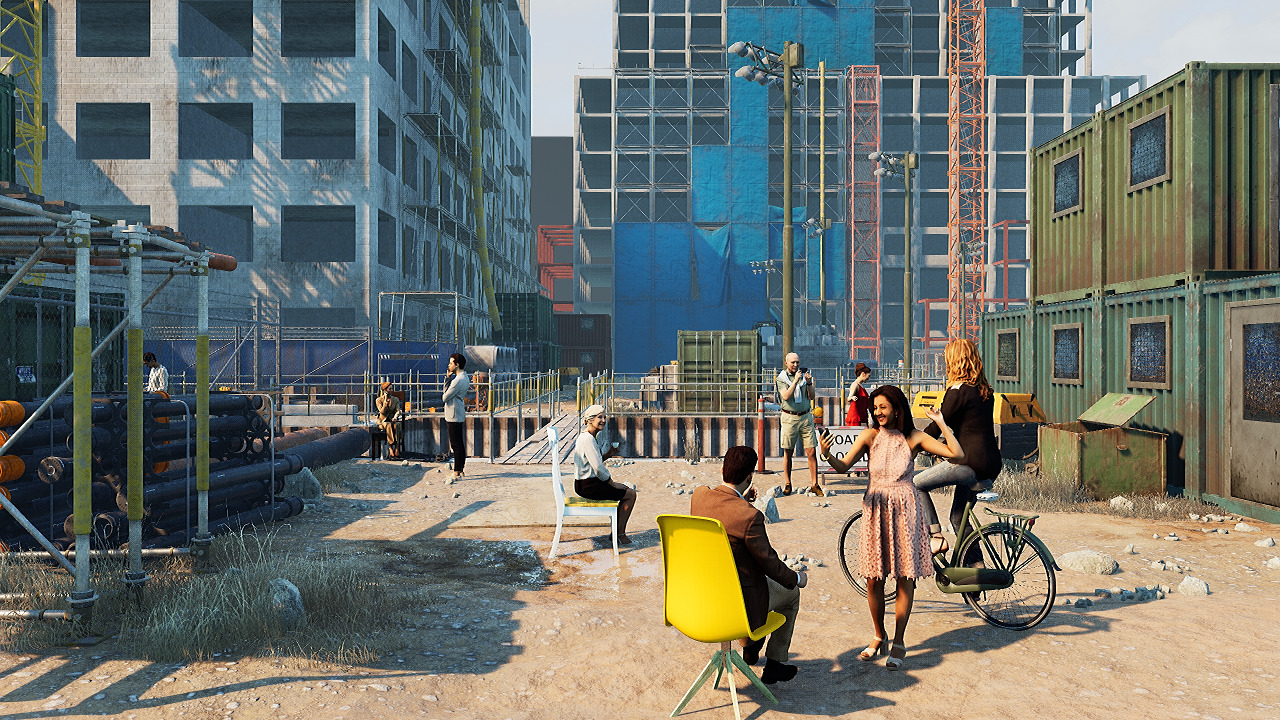} \includegraphics[width=0.23\textwidth]{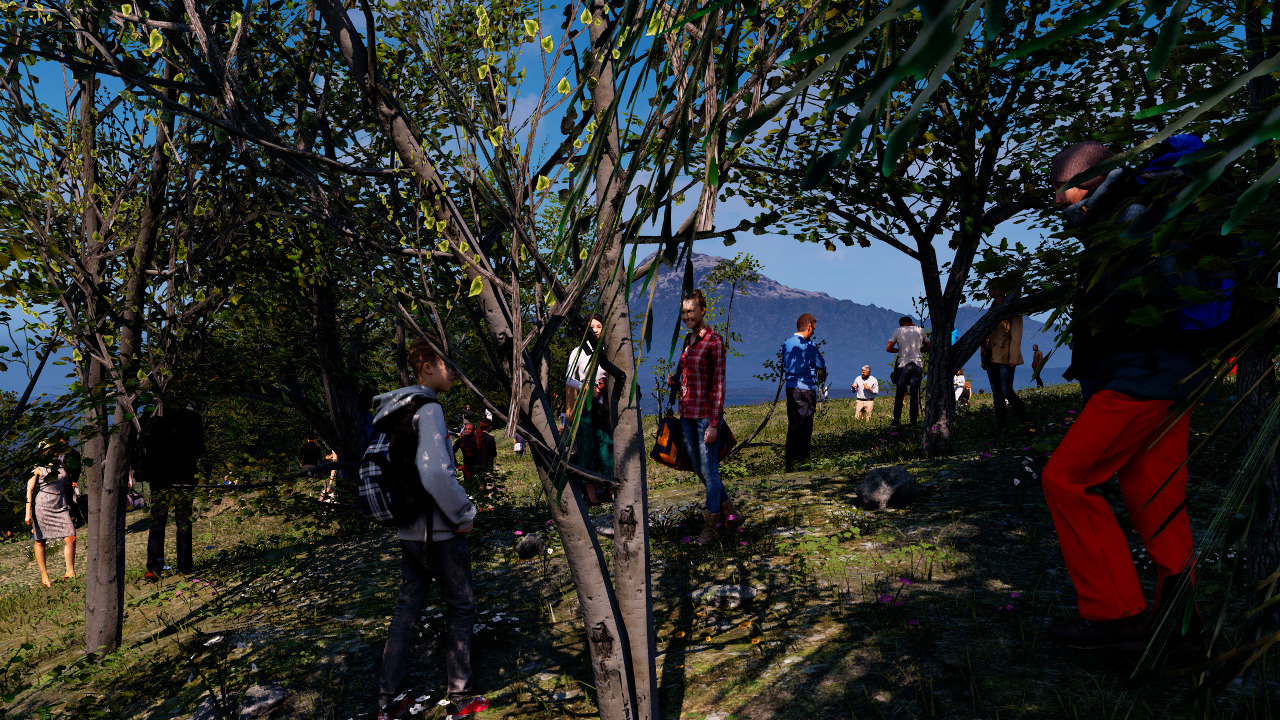}	 }
	\centerline{\includegraphics[width=0.23\textwidth]{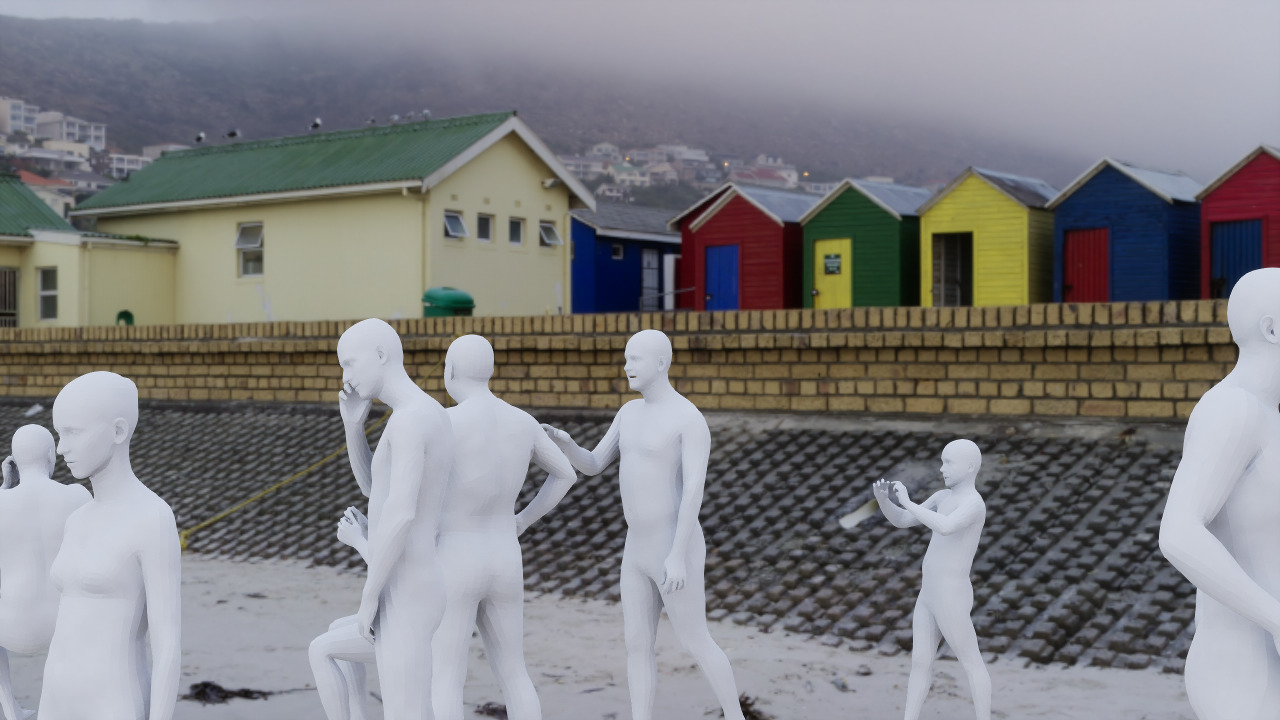}	\includegraphics[width=0.23\textwidth]{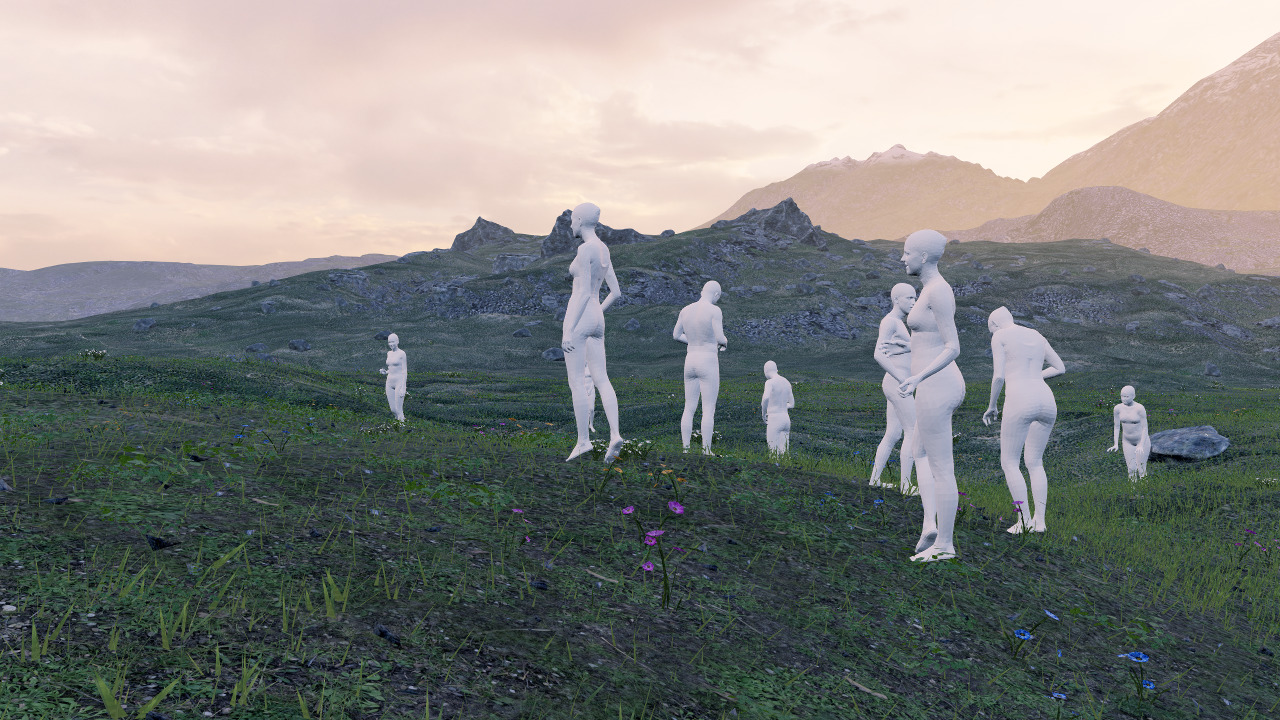}
	\includegraphics[width=0.23\textwidth]{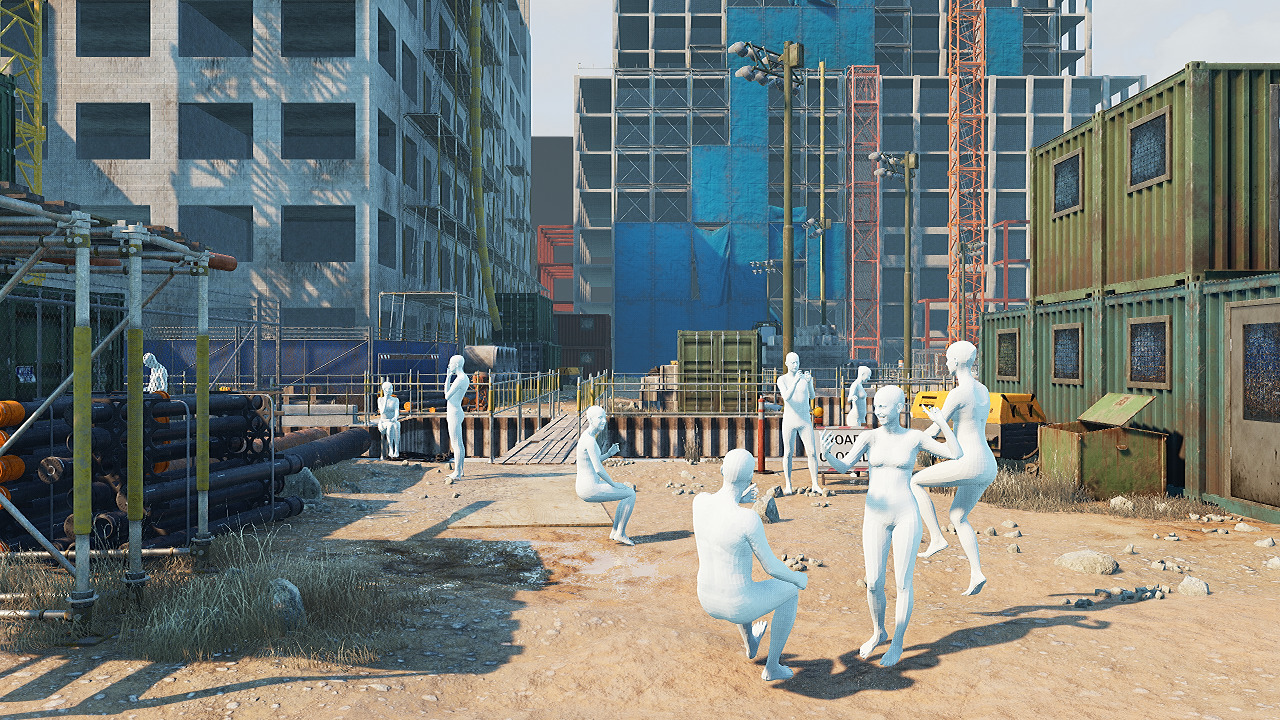} \includegraphics[width=0.23\textwidth]{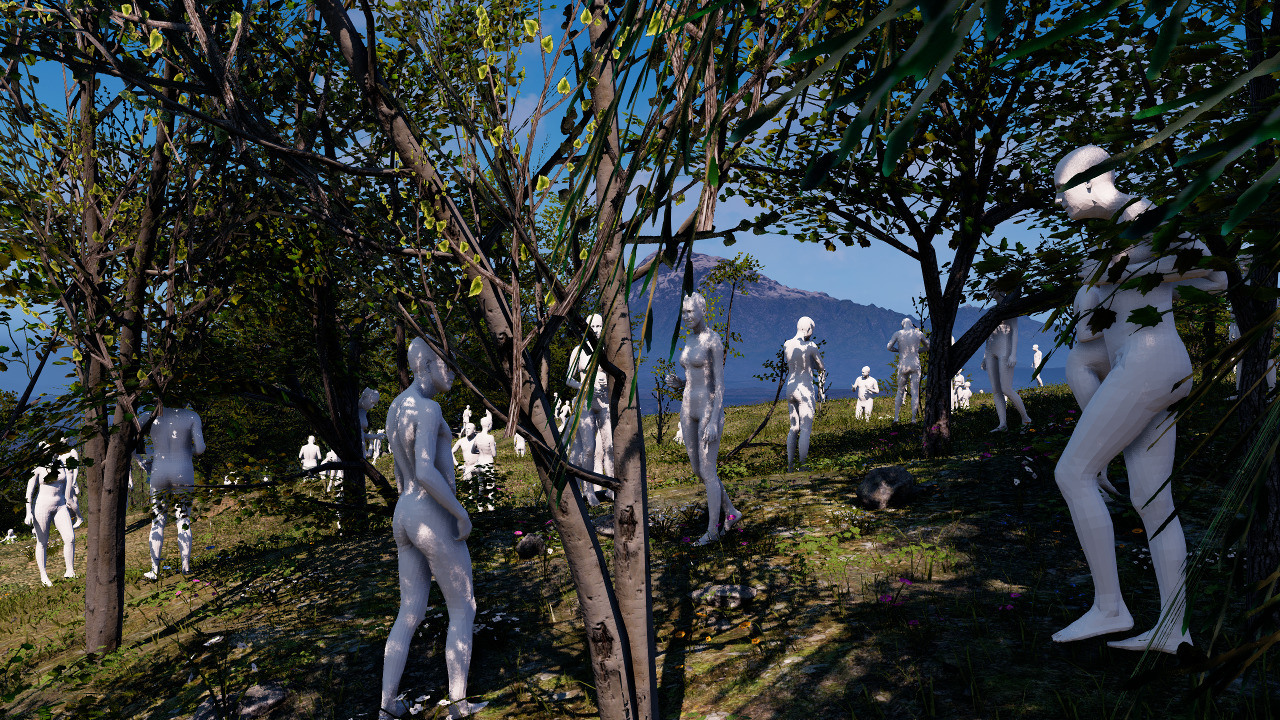}	 	}
	\centerline{\includegraphics[width=0.23\textwidth]{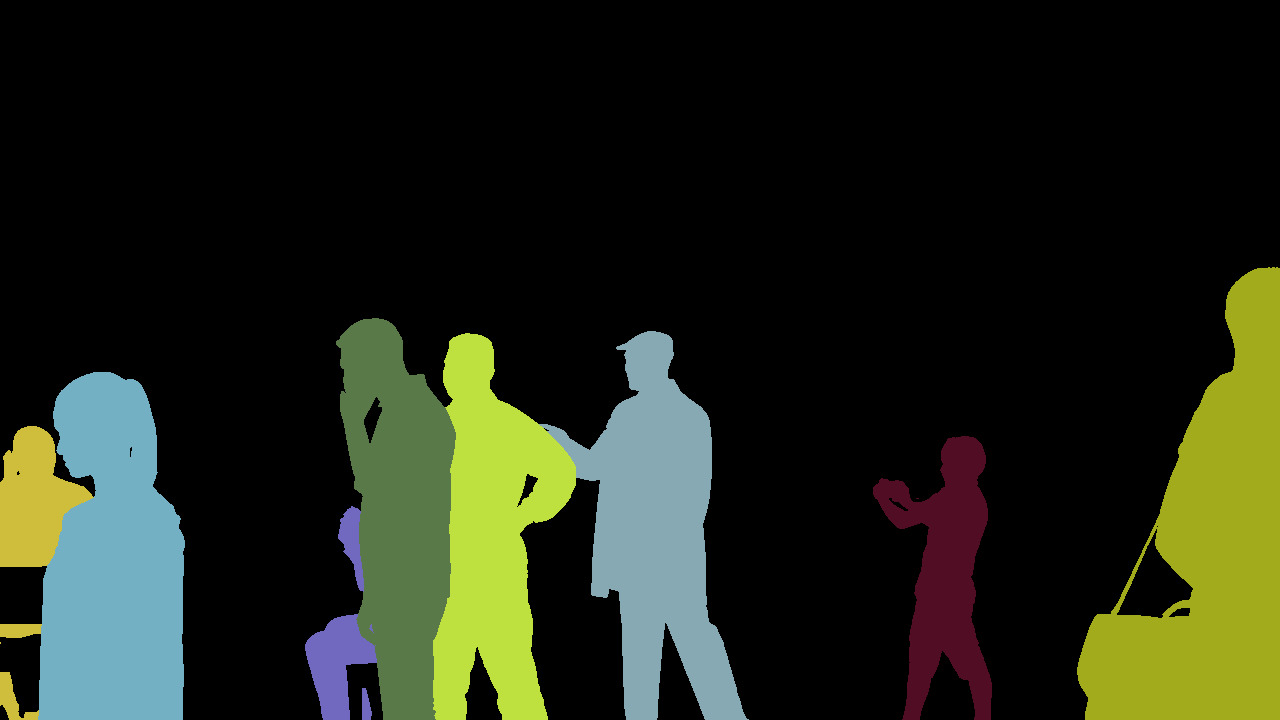}	\includegraphics[width=0.23\textwidth]{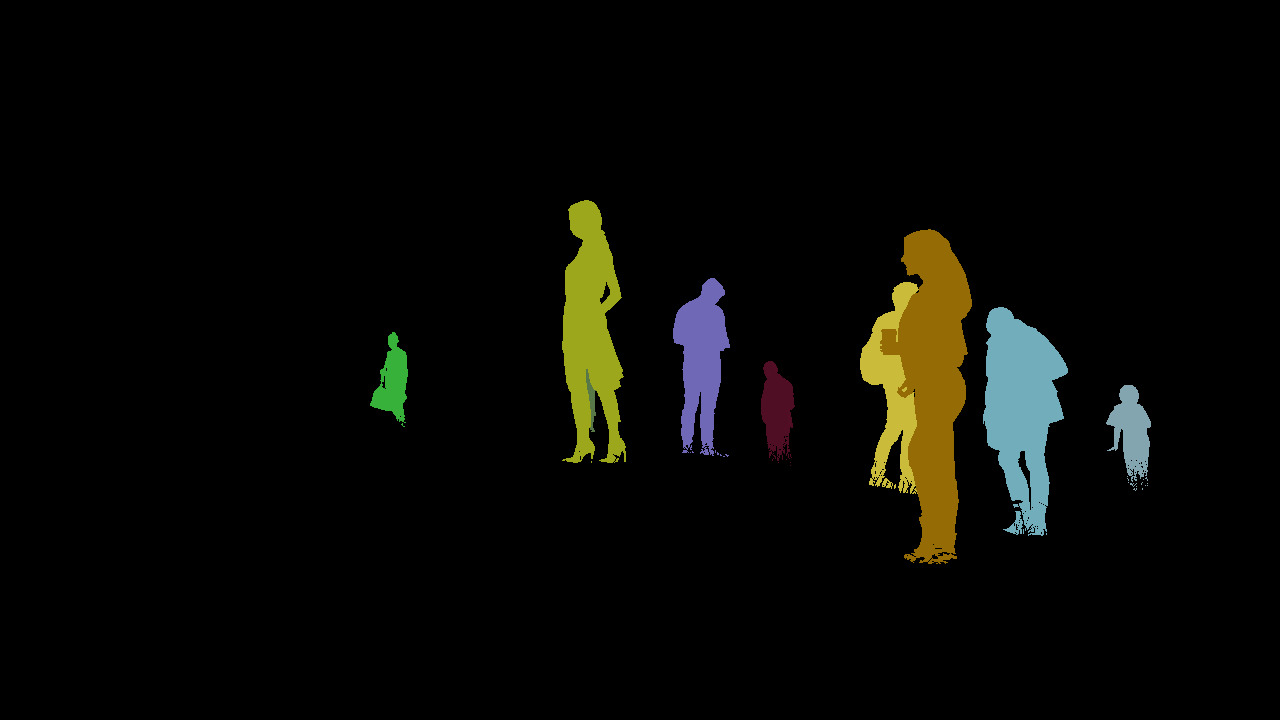} 
	\includegraphics[width=0.23\textwidth]{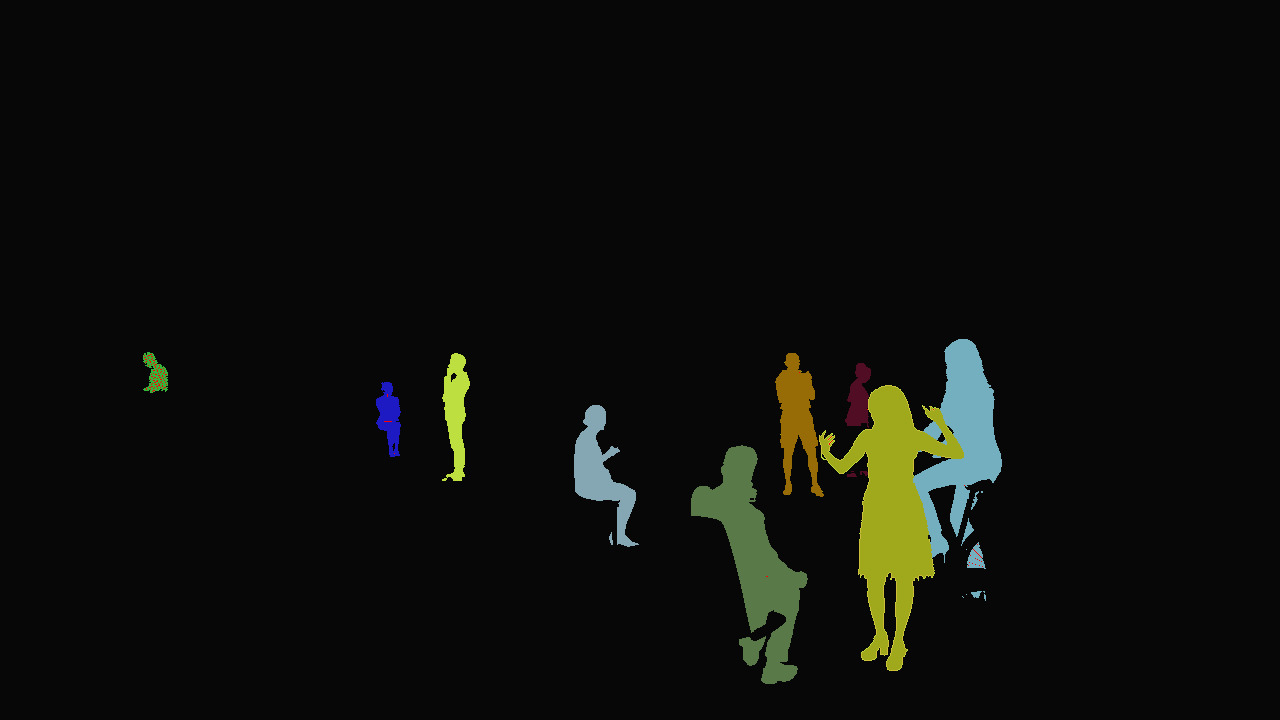}
	\includegraphics[width=0.23\textwidth]{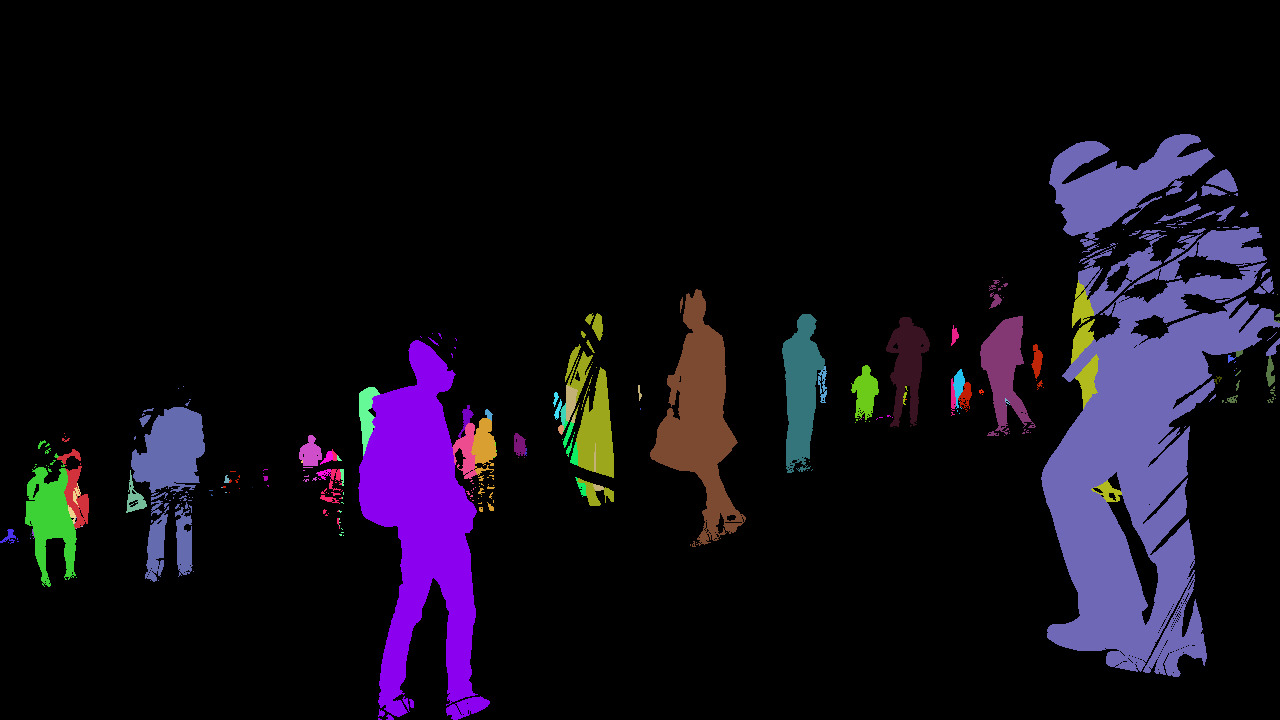}		}
	\end{minipage}
\vspace{-0.1in}
    \captionof{figure}{
    	{\textbf{\dbname}} dataset examples. Top row: images with different scenes.
		Middle: SMPL-X ground-truth bodies rendered in the scene.
		Bottom: per-person segmentation masks including environmental occlusion (see bottom right image).}
		\vspace{+4ex}
	\label{fig:teaser}
}
]
{
  \renewcommand{\thefootnote}%
    {\fnsymbol{footnote}}
  \footnotetext[1]{This work was done while DTH was at MPI-IS.}
}
\maketitle
\begin{abstract}

While the accuracy of 3D human pose estimation from images has steadily improved on benchmark datasets, the best methods still fail in many real-world scenarios.
This suggests that there is a domain gap between current datasets and common scenes containing people.
To obtain ground-truth 3D pose, current datasets limit the complexity of clothing, environmental conditions, number of subjects, and occlusion.
Moreover, current datasets evaluate sparse 3D joint locations corresponding to the major joints of the body, ignoring the hand pose and the face shape.
To evaluate the current state-of-the-art methods on more challenging images, and to drive the field to address new problems, we 
introduce~\emph{\dbname}, a synthetic dataset with high realism and highly accurate ground truth.
Here we use {\numScan} commercially-available, high-quality, textured human scans in diverse poses and natural clothing; this includes 257 scans of children.
We create reference 3D poses and body shapes by fitting the SMPL-X body model (with face and hands) to the 3D scans, taking into account clothing.
We create around 14K training and 3K test images by rendering between 5 and 15 people per image using either image-based lighting or rendered 3D environments, taking care to make the images physically plausible and photoreal. In total, {\dbname} consists of 173K individual person crops.
We evaluate existing state-of-the-art methods for 3D human pose estimation on this dataset.
and find that most methods perform poorly on images of children. Hence, we extend the SMPL-X model to better capture the shape of children.
Additionally, we fine-tune methods on {\dbname} and show improved performance on both {\dbname} and 3DPW, confirming the realism of the dataset.
We provide all the registered 3D reference training data, rendered images, and a web-based evaluation site at \url{https://agora.is.tue.mpg.de/}.

\end{abstract}
\section{Introduction}
The field of 3D human pose and shape (3DHPS) estimation from images has advanced rapidly with steadily decreasing errors on standard benchmarks \cite{Guler_2019_CVPR,iskakov2019learnable,kocabas2019vibe,kolotouros2019spin,Moon_2019_ICCV_3DMPPE,Qiu_2019_ICCV,sun2020centerhmr}.
Large training datasets and benchmarks, with ground truth, enable progress and quantitative evaluation. 
These are difficult to obtain in the case of 3DHPS.
Existing datasets have significant limitations and the rate of progress now suggests that these benchmarks are becoming saturated, making it difficult to evaluate how close the field is to fully robust and general solutions.
These datasets often have limited clothing, focus on single subjects, have limited occlusion, are captured in laboratory environments, or have a limited range of ages and ethnicities.
Additionally, accuracy is evaluated based on a small number of 3D joints, while the body is much more complex.
To drive advances in the field, we propose a novel dataset that includes challenging scenarios neglected by earlier datasets and a more challenging evaluation protocol.

\emph{\dbname} (Avatars in Geography Optimized for
Regression Analysis) is a new publicly available dataset that includes high-resolution (4K) images with ground truth 3D bodies.
{\dbname} goes beyond previous datasets in important ways.  
It includes accurate 3D body pose and shape of people in varied and complex clothing.
People with varied poses, ages and ethnicities appear in complex natural scenes with natural lighting.
Additionally, the dataset includes person-person occlusion, environmental occlusion, camera frame occlusion, crowds, children, face and hand pose, large field of view images and people appearing at a wide range of spatial scales. 
To the best of our knowledge, AGORA is the only dataset that provides all these features together with highly accurate 3D ground truth.
Figure \ref{fig:teaser} shows a few representative examples from the dataset. 

Since there is currently no technology to capture ground truth body shape and pose for real images of this complexity, 
we rely on synthetic data and a graphics rendering pipeline.
Specifically, we purchased  {\numScan} high-quality textured scans of people, which include  257 child scans from 3DPeople~\cite{3dpeople}, AXYZ ~\cite{axyz}, Human Alloy~\cite{humanalloy} and Renderpeople~\cite{renderedpeople}.
These scans provide a rich variety of ethnicity, age, pose, and clothing variation with realistic textures.
We also gathered a variety of scenes as HDRI panoramas and 3D environments.
We randomly sample 3D people and place them in scenes at random distances and orientations.
We then render them realistically using a game engine optimized for high-quality output~\cite{unreal}.

For every scan, we fit the SMPL-X body model \cite{SMPL-X:2019}, taking great care to accurately capture the correct body shape, pose, hand shape, and facial shape; see Fig.~\ref{fig:teaser} middle row.
To generate the AGORA ground truth (or reference data), we take an optimization-based approach that fits SMPL-X to  each scan. 
Specifically, we estimate both the pose and body shape under clothing, similar to~\cite{Balan:ECCV,zhang2017detailed}.
The estimated SMPL-X fits have an average error of 5mm, making them accurate enough to benchmark existing state-of-the-art (SOTA) methods.
For backward compatibility with SMPL, we also provide ground truth in the gender-neutral SMPL format\footnote{Given a SMPL-X mesh, we convert it to gender-neutral SMPL format by fitting the gender-neutral SMPL template to it. In this work, SMPL fits are always generated through this process unless otherwise stated.}~\cite{SMPL:2015} used by many current methods~\cite{kanazawa2018end,kolotouros2019spin,sun2020centerhmr}.

In addition to adults, the AGORA dataset contains images of children. It is probably the only dataset of children with reference 3D pose and shape.
Existing 3DHPS methods focus on adult bodies and perform poorly on images of children.
With AGORA, we evaluate this performance, but go further and extend the SMPL-X shape space to capture the variation in body shape beween infants and adults.  
Specifically, we introduce a shape dimension that interpolates between an adult SMPL-X body template and the infant SMIL template \cite{hesse2018learning}, which we convert to SMPL-X format.
This results in an extra shape parameter that can be optimized like any other SMPL-X shape parameter.

We make the SMPL-X fits available for all the {\numtrain} training and {\numval} validation images, enabling training with AGORA. 
We withhold the ground truth bodies from the {\numtest} test images and instead provide an evaluation server.
While we cannot provide the commercial scans, we provide a ``shopping list'' of the training and validation scans so that others can purchase them. 
Purchasing the scans extends the applications of AGORA to other problems such as 3D clothing modeling, neural avatars, and shape regression.
We also provide the test scripts for researchers to test their methods on the validation set.  

We use \dbname to evaluate SOTA 3DHPS methods with a novel protocol.
In addition to the common 3D joint-based error measures, we provide a vertex-to-vertex error, 
and evaluate body, hands and face pose and shape.
We also use 2D occlusion masks  (Fig.~\ref{fig:teaser} bottom row) to evaluate the performance of methods at varying levels of occlusion.
Since our images contain multiple people, methods may detect too few as well as too many people.
Consequently, we introduce an error measure that goes beyond the standard single-person measures and rewards methods for both 3DHPS and detection accuracy.
We observe higher errors for SOTA methods on AGORA than on other datasets, suggesting that AGORA is more challenging. 
We also show that our training set can be used to improve recent 3D pose estimation methods~\cite{kolotouros2019spin} not only on AGORA but also on 3DPW~\cite{vonMarcard20183dpw}.
This validates that the synthetic data is sufficiently real to be useful. 

In summary, we contribute a new, varied and challenging dataset to evaluate and improve the SOTA in 3D human pose and shape estimation and to push the field in new directions.
The dataset is synthetic but diverse and realistic.  We use new evaluation metrics and provide detailed analysis of limitations of current methods. We also introduce a new child model to generate better ground truth shape for children.
We provide the training and  validation set images with SMPL-X and SMPL ground truth and 2D masks. 
We also provide test images along with evaluation code and will maintain a web evaluation server: \url{https://agora.is.tue.mpg.de/}.

\section{Related Work}
\begin{table*}[t]
	\scriptsize
	\centering
		\begin{tabular}{l|r|c|l|l|r|r}
			\toprule \scriptsize
			Dataset & Sub. \# & Image & Complexity & Clothing & Body anno. & Ground truth format\\
			\midrule
			HumanEva~\cite{sigal2010humaneva}  	& 4  & lab & 1 subject, no occlusion & limited & B & 3D joint locations\\		
			\rowcolor{Gray}
			Human3.6M~\cite{h36m_pami} 			& 11  & lab & 1 subject, minor occlusion & limited & B & 3D joint locations\\		
			TotalCapture~\cite{trumbleBMVC2017}	& 5  & lab & 1 subject, no occlusion & mocap suit & B & 3D joint locations \\
			\rowcolor{Gray}		
			PanopticStudio~\cite{joo2019panoptic}& $\sim$100  & lab & multiple subjects \& furniture & varied & BFH& 3D joint locations\\	
			HUMBI~\cite{yu2018humbi} 			& 772 & lab & 1 subject, no occlusion & rich & BFH & meshes, SMPL\\		
			\rowcolor{Gray}
			3DPW~\cite{vonMarcard20183dpw} 	& 18   & natural & multiple subjects in the wild &  varied & B& SMPL \\		
			MuPoTS-3D~\cite{mehta20183dv} 		&  8 & natural & multiple subjects in the wild & varied & B & 3D joint locations\\	
			\rowcolor{Gray}
			MPI-INF-3DHP-Train~\cite{mono-3dhp2017} 	& 14   & both & 1 subject, minor occlusion &varied & B& 3D joint locations\\		
			3DOH50K~\cite{zhang2020object} 	&  	n/a  & lab & 1 subject, object occlusion &limited & B & SMPL \\		
			\rowcolor{Gray}
			EFT~\cite{joo2020exemplar} 	&  $>$ 1000 & natural & multiple subjects, in the wild &varied & B & SMPL \\	
			STRAPS~\cite{sengupta2020synthetic}		& 62  & natural & 1 subject, in the wild &limited & B  & SMPL \\
			\rowcolor{Gray}
			SMPLy~\cite{Leroy2020SMPLyB3}		&  742 & natural & multiple subjects, in the wild, frequent occlusion & rich & B & SMPL \\	

			\midrule
			MuCo-3DHP~\cite{mehta20183dv} 		& 8  & composite$^\dagger$ & multiple subjects in the lab & limited & B & 3D joint locations \\	
			\rowcolor{Gray}
			MPI-INF-3DHP-Test~\cite{mono-3dhp2017} 	& 14   & composite$^\dagger$ & 1 subject, minor occlusion & varied & B & 3D joint locations\\	
			SURREAL~\cite{varol17_surreal} 		& 145  & composite$^\dagger$ & 1 subject, no occlusion & texture$^\mathparagraph$ & B& SMPL\\
			\rowcolor{Gray}	
			3DPeople~\cite{Pumarola_2019_ICCV} 	& 80  & composite$^\dagger$ & 1 subject, no occlusion & synthetic$^\ddagger$  & B & 3D joint locations\\
			\hline
			{\dbname (ours)}	& $>$350  & realistic$^{\dagger\dagger}$ & multiple subjects in the wild, frequent occlusion & rich & BFH & SMPL-X, SMPL, masks\\
		
			\bottomrule
		\end{tabular}
\vspace{-0.1in}
	\caption{ Comparison of datasets that provide images and 3D human pose annotations. Body annotation type B, F, and H correspond to body, face, and hands respectively. $^\dagger$: 2D foreground layers pasted on background images. $^{\dagger\dagger}$: 3D models positioned in 3D with panoramic background or full 3D scenes.
		$^\mathparagraph$: unclothed human body with clothing texture. $^\ddagger$: clothed human body with texture.
		}
	\label{tab:dataset-comparison}
\end{table*}
Many datasets have been proposed for 3DHPS estimation, but each has limitations as summarized in Table~\ref{tab:dataset-comparison}. 
While there are many 2D datasets, we focus on those with 3D ground truth of one form or another.

\textbf{Datasets with real images.}
Unlike 2D annotation, 3D body poses are difficult for humans to annotate since the task is ambiguous and requires metric accuracy. 
Consequently, existing benchmarks rely on  multiple synchronized cameras. 
For example, HumanEva~\cite{sigal2010humaneva} , Human3.6M~\cite{h36m_pami}, and TotalCapture~\cite{trumbleBMVC2017} synchronize video cameras with motion capture (mocap) systems that provide ground truth through optical markers. 
While providing accurate 3D pose, the image complexity is limited: lack of background variation in lab scenarios, only one subject in each image, no scene occlusions, and little clothing variety due to the attachment of markers, which, unfortunately are also visible in the images.
These methods typically evaluate accuracy based on 3D joint locations.  
Note that, while the 3D joints are commonly treated as ``ground truth'', they are not directly observed, but rather are inferred by the mocap system based on an approximate skeletal body structure.

Alternatively, several methods use marker-less motion capture, \eg~MuPoTS-3D~\cite{mehta20183dv}, PanopticStudio~\cite{joo2019panoptic}, MPI-INF-3DHP-Test~\cite{mono-3dhp2017}, and HUMBI~\cite{yu2018humbi}.
Such methods are typically less accurate than marker-based systems, 
but they avoid intrusive markers, allow more varied clothing, and sometimes are used in more realistic scenes \eg~outdoors.
IMU sensors provide another way to measure 3D poses, which is less intrusive than mocap markers but also less accurate due to yaw drift.
Von Marcard~\etal~\cite{vonMarcard20183dpw} explicitly account for this by combining IMU data with monocular video, enabling in-the-wild capture.
We consider these datasets as reference data rather than ``ground truth''  because the accuracy of the method is evaluated in a separate process (\eg~using mocap data) and not on the image data in the benchmark.
In contrast, for {\dbname} we report how close the SMPL-X meshes are to these reference scans, directly indicating the fidelity of our pseudo ground truth.

All the above are limited in the complexity of the clothing, occlusions, scene variety, ethnicity, etc. 
Of the above only PanopticStudio~\cite{joo2019panoptic} and HUMBI~\cite{yu2018humbi} consider the face and hands together with bodies. 		

\textbf{Synthetic datasets.}
Computer graphics has the potential to synthesize large-scale image datasets, where ground truth is generated by animating parametric 3D human models such as SMPL~\cite{SMPL:2015}, MakeHuman~\cite{makehuman}, or Mixamo~\cite{mixamo}.
The main challenge for such methods lies in creating data that is sufficiently realistic in terms of body shape, ethnicity, motion, cloth deformation, texture, and interaction with environments.
In several datasets, images are created by compositing 3D people on image backgrounds.
MHOF~\cite{multihumanflow}, LTSH~\cite{hoffmann2019learning}, 3DPeople~\cite{Pumarola_2019_ICCV}, and SURREAL~\cite{varol17_surreal} render 3D people on the background image, while MPI-INF-3DHP-Train~\cite{mono-3dhp2017} and MuCo-3DHP~\cite{mehta20183dv} paste a  segmented real human foreground on top of the background.
Such composition does not faithfully reflect the local statistics of pixel intensity in real images and does not support methods that learn  how humans interact with scenes.
Most similar to us is SimPose \cite{zhu2020simpose}, which poses 17 rigged commercial scans \cite{renderedpeople} and SURREAL data rendered in a 3D scene. The 3D scenes are simplistic, the scans lack diversity, there is no evaluation site, and the dataset is not public.

A recent promising direction synthesizes realistic looking people in images \cite{zanfir2020human,Zhang:arXiv:19}.
Zanfir~\etal~\cite{zanfir2020human} use a learned human synthesis method to insert generated people in images such that they make sense relative to the scene geometry and lighting.  While they can condition the generated person on pose and shape, the resulting images contain artifacts that are common to generative models, making the results unsuitable as ground truth.

\textbf{Other human-related datasets.}
There are many other datasets of real humans in images that do not contain 3D ground truth.
For example, 
OCHuman~\cite{pose2seg2019} focuses on occlusion in real single-view images and provides 2D joint landmarks and human segmentation masks. Early multiview sequences \eg~Adobe data~\cite{vlasic2008articulated}, MVIC~\cite{LiuPAMI2013}, and MARCOnI~\cite{EEJTP15} also consider 2D landmarks and silhouettes as evaluation measures. 
The hunger for large training corpora for deep learning motivates self-supervised strategies~\cite{kanazawa2018end,kocabas2019vibe,tung2017nips} that leverage 2D landmark annotations in LSP-Extended~\cite{johnson2011learning}, COCO~\cite{lin2014microsoft}, and MPII~\cite{andriluka14benchmark}.
Several recent datasets, \eg~EFT~\cite{joo2020exemplar}, STRAPS~\cite{sengupta2020synthetic} and 3DOH50K~\cite{zhang2020object}, are generated by fitting a body model to the images, while others fit to videos of complex scenes with ``frozen'' people \cite{li2019learning} using structure from motion \cite{Leroy2020SMPLyB3} or multi-view matching \cite{shen2020multi}.
Methods like EFT and SMPLy \cite{Leroy2020SMPLyB3} provide image variety, which is good for robustness, but with unknown accuracy in body shape and pose.

In summary, no single dataset can address all needs of the community.  
{\dbname} provides realistic textures, complex body shapes and clothing, complex varied scenes and lighting, high-resolution (4K) imagery, varied occlusion, all with high-quality 3D ground truth.   
This new benchmark reveals limitations of current approaches while providing novel, high-quality training data for multiple applications.
\section{Method: Obtaining reference data}\label{sec-method}
To construct {\dbname}, we purchased high-quality textured 3D scans from 3DPeople~\cite{3dpeople}, AXYZ \cite{axyz}, Human Alloy~\cite{humanalloy} and Renderpeople~\cite{renderedpeople}.
We selected {\numScan} scans for inclusion in the dataset spanning more than {\numSub} unique subjects.
A scan $\scan$ comprises a set of 3D points $S\subset \mathbb{R}^{3}$ and their connectivity $F_{\scan}$, $\scan=\{S , F_{\scan}\}$. 
To each scan $\scan$ we fit a parametric SMPL-X body model $\model = \{\modelpoints, F_{\model} \}$, whose vertex locations $\modelpoints(\theta,\beta,\psi) \subset \mathbb{R}^3$ are controlled by parameters for pose $\theta$, shape $\beta$, and facial expression $\psi$~\cite{SMPL-X:2019}. 
$\theta$ consists of body pose $\theta_b$ and hand pose $\theta_{h}$.
Hand pose $\theta_{h}$ is a function $\theta_{h}(Z_{h})$ of a PCA latent vector $Z_{h} \in \mathbb{R}^{6}$. 

Fitting a SMPL-X mesh $\model$ to a scan $\scan$ amounts to solving for the optimal parameters $(\theta,\beta,\psi)$ such that $\model$ resembles $\scan$.
The fitting process 
takes into account that SMPL-X explains the body in minimal clothing while the scans are typically clothed. 
In this process we exploit the fact that a person may appear in multiple scans and their shape parameter, $\beta$, should be the same across scans.

We first initialize the parameters by an approach that extends the single-view SMPLify-X fitting \cite{SMPL-X:2019} to multi-view images rendered using $C$ pre-defined virtual cameras.
The initial mesh, $\modelpoints$, obtained by multi-view SMPLify-X fitting is only approximately aligned with $\scanpoints$.
While sparse 2D landmarks constrain the 3D pose, they provide little information about body shape.
To refine the shape and pose, we  fit SMPL-X to the 3D scan surface.  
However, this is challenging because SMPL-X cannot model things like hair and clothing that are present in the scans. To address this, we use the idea of fitting body shape under clothing \cite{Balan:ECCV,zhang2017detailed}.

Similar to~\cite{zhang2017detailed}, we define energy terms $\skinterm$ and $\clothterm$ for skin and clothing, respectively.
Both aim to bring the model surface close to the scan, whereas $\clothterm$ additionally penalizes body vertices being outside the clothing. 
In other words, our objective function tries to move the model as close as possible to the scan near the visible skin while discouraging the clothing vertices from penetrating the model. We label skin and cloth vertices on scan using Graphonomy \cite{gong2019graphonomy}.
We keep the 2D landmark data term $E_J^c$ that penalizes differences between projected and observed keypoints from multi-view fitting,
as they provide information complementary to $\skinterm$ and $\clothterm$. See Sup.~Mat.~for more details.

We fit each model $M_i(\beta_{i}, \theta_{i}, \psi_{i})$ to the corresponding scan  $S_i$  in parallel, for scans, $i$, of the same identity. We optimize jointly for $\theta_{i}$, $\psi_{i}$ and  $\beta_{i}$ while minimizing the shape (inter-beta) distance $E_\text{ib}$ between scans of the same identity.
The objective function is:
\begin{equation}
\begin{split}
&E(\beta_{1},\dots, \beta_{N},\theta_{1},\dots, \theta_{N}, \psi_{1}, \dots,\psi_{N}) = \\
& \sum_{i=1}^{N} \left(  \lambda_J \sum_{c=1}^{C} E_J^{c,i} + \lambda_\text{s}\skinterm^i + \lambda_\text{c}\clothterm^i + E_{\text{reg}}^i \right) + \lambda_{\text{ib}} E_{\text{ib}}, \\
& E_\text{ib} =\sum_{i=1}^{N} \sum_{j=i+1 }^{N} \left\Vert \beta_{i} - \beta_{j}\right\Vert_2^2, \\
& E_\text{reg} = \lambda_{\theta_b} 	E_{\theta_b} (\theta_b) + \lambda_{\theta_h} 		E_{\theta_h} (\theta_h)+ \lambda_{\beta}    		E_{\beta}(\beta)	+ 	\lambda_{\mathcal{E}}    		E_{\mathcal{E}}(\psi), \nonumber
\end{split}
\end{equation}
where $E_\text{reg}$ contains $L_2$ priors used to constrain the body shape, pose and expression, as defined in \cite{SMPL-X:2019}.  
Different weights denoted by $\lambda$ are used for each term.

This approach exploits semantic information (2D landmarks, skin/clothing segmentation) as well as geometry (3D shapes) 
to obtain accurate fits as demonstrated in Sec.~\ref{subsec-ex-fittting}.

\subsection{Fitting child scans}\label{kid-scan-fit}
\begin{figure}[t]
	\centerline{\includegraphics[trim=0 0 0 3.4in, clip=true,width=\columnwidth]{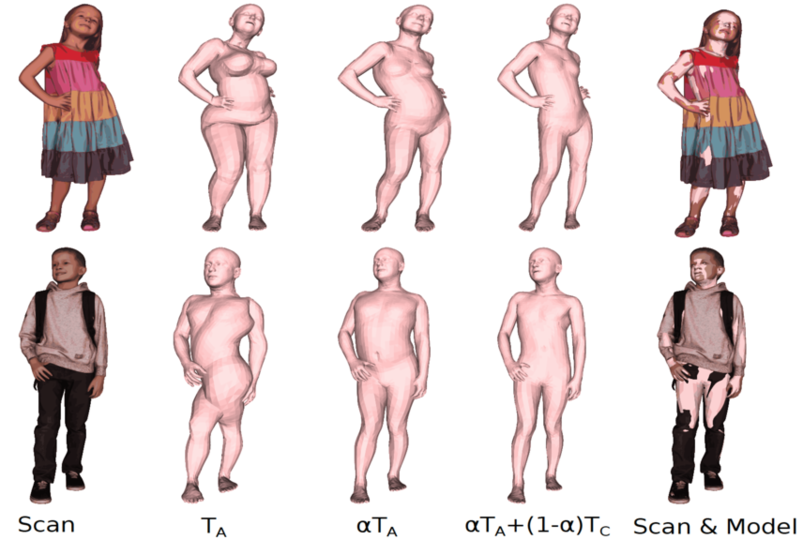}}
	\vspace{-0.1in}
	\caption{Fitting child scans using an adult template ($T_A$), a scaled adult template ($\alpha T_A$),  and our proposed approach, which interpolates between adult and infant templates ($\alpha T_A + (1-\alpha) T_C$).}
	\label{fig:child-scan-fitting}
\end{figure}

AGORA also contains 257 child scans. Fitting SMPL-X directly to these scans results in distorted fits as shown in the 2nd column of {Fig.~\ref{fig:child-scan-fitting}}, because SMPL-X cannot represent children. 
Naively scaling the adult SMPL-X template, i.e.~$\alpha T_A$, by optimizing for a global scale parameter $\alpha$ is better but still unnatural since children have different proportions than adults. 
To solve this problem, we take the mean infant body template from SMIL~\cite{hesse2018learning}  and convert it to SMPL-X topology, $T_C$.
 We find that interpolating between the adult SMPL-X template $T_A$ and the SMIL infant template $T_C$ approximately captures the shape of children. See Sup.~Mat. 
Note that, while not perfect, we use the adult shape space for children and only vary the template shape.
Incorporating children then involves only a minor change to the fitting process.
In addition to optimizing the shape parameters, $\beta$,  we also optimize for a weight, $\alpha\in[0,1]$, the linearly interpolates the templates.
This produces more accurate body shapes for children, as shown in Fig.~\ref{fig:child-scan-fitting}.

\section{AGORA Dataset}
{\dbname} consists of {\numScan} scans spanning more than {\numSub} unique subjects, all paired with SMPL-X fits
(we also supply SMPL fits for backward compatibility).
While generally robust, the fitting approach fails sometimes for hands and faces. 
This typically happens when the hands are grasping objects. 
Since we want high-quality ground truth, we manually curate the results of the automatic process and create two different sets:
(1) those with well aligned body, face and hands ({\numBFH}, BFH); and (2) those only with well aligned bodies ({\numBody}, B). 
When evaluating algorithms for 3DHPS estimation, only body joints and vertices are considered for B scans, while body, hand and face joints and vertices are evaluated for BFH scans.

We sample {\numtestscans} scans to create the test set; for these, the SMPL-X fits are withheld from public release. 
We make sure there are no overlapping scans between these {\numtestscans} scans and the rest, and no selected scans have been included to train previous work~\cite{saito2019pifu}. 
Our test set spans {\numtestsub} subjects.
We create approximately {\numtest} images from the test scans, including many challenging scenarios, with the goal of making them photorealistic.
From the remaining scans, we sample {\numtrainscans} scans as a training set and {\numvalscans} scans as a validation set and create {\numtrain} training and {\numval} validation images, whose ground-truth SMPL-X parameters and 2D segmentation masks are included in the released dataset. 
Since the ground truth 3D scans are commercially available, it is possible for people to cheat by using test scans for training.
We have also built in several countermeasures to detect cheating that we do not describe in this paper.

AGORA images are rendered using perspective cameras with focal lengths 18mm, 28mm and 50mm.  
All images are rendered using Unreal Engine~\cite{unreal} on a single Windows 10 PC with NVIDIA RTX 2080 graphics hardware.
Renderings are generated either with image-based lighting using freely available HDR backgrounds~\cite{hdrhaven} or with free and commercial 3D environments obtained from the Unreal Marketplace.
The image-based lighting scenes used hardware-accelerated ray-tracing for accurate ground plane shadows.
For scans in seated poses, we insert random chairs at the appropriate height 
so that the scans appear naturally supported.
See Sup.~Mat.~for details of the rendering process.

\subsection{Fitting Accuracy}\label{subsec-ex-fittting}
To evaluate the accuracy of results on \dbname, and to know whether
improvements are significant, we first need to know the accuracy of
our ground truth\footnote{Note that with traditional mocap ground truth, only the accuracy of
the markers is known -- the accuracy of the 3D joints is actually unknown.}.
We define accuracy relative to the high-quality 3D scans in the
following ways:

\textbf{1.~Skin error}.
For the visible skin vertices on the scan, we compute the Euclidean
distance of the nearest point on the triangle of the reconstructed
model $M$. 
An accurate model fit should fit closely to the skin. 
We report the weighted mean distance as our final error value where the weight is the probability of the scan vertex belonging to skin calculated using Graphonomy \cite{gong2019graphonomy}. 

\textbf{2.~Penetrating clothing error}.
The SMPL-X fits are supposed to be fully inside the clothing.
We report two values for clothing vertices on the scans: (1) the
percentage of them that penetrate the body model. 
(2) for those penetrating vertices, we calculate their distance to the closest point on the model surface and compute the weighted avg.~error.

We consider only the scans without any large objects for the error calculation and  
report an average skin error of approximately 4.73mm. 
Only 16\% of cloth vertices are inside the body with an average
distance of 4.63mm.
An error of approximately 5mm is significantly below any industry standards for the
measurement of live humans and is less than the soft-tissue motion of mocap
markers on the body \cite{CAMOMILLA201714}.
Thus, we believe that the SMPL-X fits provide valid pseudo ground truth.

\begin{savenotes}
\begin{table*}[t]
	\scriptsize
	\centering
	\begin{tabular}{l|l|cccc|cccc|cc|cc|c}
	\toprule
	
	&\multirow{2}{*}{Method} & \multicolumn{4}{c|}{MPJPE $\downarrow$} & \multicolumn{4}{c|}{MVE $\downarrow$} & \multicolumn{2}{c|}{NMJE $\downarrow$} & 
	\multicolumn{2}{c|}{NMVE $\downarrow$} &  \multirow{2}{*}{F1 score$\uparrow$} \\ 
	\cline{3-14}
& & \multicolumn{1}{c}{B} & \multicolumn{1}{c}{LH/RH} & \multicolumn{1}{c}{F} & \multicolumn{1}{c|}{FB} & \multicolumn{1}{c}{B} & \multicolumn{1}{c}{LH/RH} & \multicolumn{1}{c}{F} & \multicolumn{1}{c|}{FB} &  \multicolumn{1}{c}{B} & \multicolumn{1}{c|}{FB} &  \multicolumn{1}{c}{B} & \multicolumn{1}{c|}{FB} & \multicolumn{1}{c}{}   \\
\midrule
 \multirow{4}{*}{\rotatebox[origin=c]{90}{~SMPL}} 
&HMR~\cite{kanazawa2018end}         &  180.5 &          N/A &   N/A &    N/A &  173.6 &          N/A &   N/A &    N/A &  226.0 &    N/A &  217.0 &    N/A & 0.80 \\
&CenterHMR~\cite{sun2020centerhmr}   &  168.1 &          N/A &   N/A &    N/A &  161.4 &          N/A &   N/A &    N/A &  242.3 &    N/A &  233.9 &    N/A &       0.69 \\
&EFT~\cite{joo2020exemplar}       &  165.4 &          N/A &   N/A &    N/A &  159.0 &          N/A &   N/A &    N/A &  203.6 &    N/A &  196.3 &    N/A &       \textbf{0.81} \\
&SPIN~\cite{kolotouros2019spin}      &  175.1 &          N/A &   N/A &    N/A &  168.7 &          N/A &   N/A &    N/A &  223.1 &    N/A &  216.3 &    N/A &       0.78 \\

	\cline{2-15}
&SPIN-ft (ours)    &  \textbf{153.4} &          N/A &   N/A &    N/A &  \textbf{148.9} &          N/A &   N/A &    N/A &  \textbf{199.2} &    N/A &  \textbf{193.4} &    N/A &       0.77 \\

\midrule
\multirow{3}{*}{\rotatebox[origin=c]{90}{~SMPL-X}} 
&SMPLify-X~\cite{SMPL-X:2019}  &  182.1 &    \textbf{46.5/49.6} &  \textbf{52.9} &  231.8 &  187.0 &    \textbf{48.3/51.4} &  \textbf{48.9} &  236.5 &  256.5 &  326.5 &  263.3 &  333.1 &      0.71 \\

&ExPose~\cite{choutas2020monocular}     &  \textbf{150.4} &    72.5/68.8 &  55.2 &  \textbf{215.9} &  \textbf{151.5} &    74.9/71.3 &  51.1 &  \textbf{217.3} &  \textbf{183.4} &  \textbf{263.3} &  \textbf{184.8} &  \textbf{265.0} &       \textbf{0.82} \\
&Frankmocap~\cite{rong2020frankmocap} &  165.2 &    52.3/53.1 &    N/A &    N/A &  168.3 &    54.7/55.7 &  N/A &  N/A &  204.0 &  N/A &  207.8 & N/A &  0.81 \\
\bottomrule
\end{tabular}
\vspace{-0.07in}
	\caption{Comparison of SOTA 3DHPS methods on the AGORA testset. SPIN-ft is SPIN after finetuning on the AGORA training set described in Sec.~\ref{finetuning}. SMPL-based methods are evaluated on B+BFH and SMPL-X-based methods are evaluated on the BFH subset of AGORA. Error  metrics are described in Sec.~\ref{datset-metric}. All numbers are in millimeters.}
	\label{tab:baselines-evaluation}
\end{table*}
\end{savenotes}
\subsection{Evaluation metrics}\label{datset-metric}
A common practice in evaluating 3DHPS methods is applying Procrustes alignment~\cite{gower1975generalized} before computing the error. Doing so eliminates discrepancies in scale, translation and rotation, measuring only the error in poses (PA-MPJPE) and shapes (PA-MVE/V2V).
This convention is largely due to the fact that existing HPS datasets, \eg~\cite{joo2020exemplar,vonMarcard20183dpw}, contain only pose and shape annotations, and HPS methods estimate the body relative to the camera.
In contrast, \dbname provides \emph{complete} 3D pseudo ground truth: body parameters of each person and their spatial arrangement in the 3D scene, enabling a more comprehensive error measure.

Consequently, we do not apply Procrustes alignment but only align at the pelvis, \ie~MPJPE and MVE/V2V, because estimating absolute depth is ambiguous.
Furthermore, since \dbname has 5-15 people per image, methods may not detect every person leading to misses, \ie~false negatives. 
Due to occlusions, methods may also detect bodies where there are actually no people, \ie~false positives.
Accuracy on AGORA means high detection performance and low error for every correct detection; consequently, we must penalize false negatives and false positives.
Thus, we normalize the MPJPE and MVE/V2V error by the standard detection metric, F1 score (the harmonic mean of recall and precision), and refer to this as {\em Normalized Mean Joint Error (NMJE)} and  {\em Normalized Mean Vertex Error (NMVE)}. 
F1 score punishes both misses and false alarms so NMJE/NMVE increase the reported error for methods that make either type of mistake in detection.
As a result, to reduce the overall NMJE/NMVE, the method needs to miss no one, 
detect no spurious bodies, and estimate accurate poses and shapes for each correct detection, 
making NMJE/NMVE more challenging and comprehensive than other metrics.

We evaluate 3DHPS methods along different dimensions and also provide a combined score. For SMPL-based methods, we just evaluate on body joints and vertices using both B and BFH scans.
For SMPL-X-based methods, we evaluate separately on the body, hands and face and also provide a weighted sum of the three as a full body (FB) error. SMPL-X-based methods are evaluated only for BFH scans.

\textbf{B-MPJPE} is evaluated on 24 body joints of SMPL and 22 body joints of SMPL-X after aligning the pelvis. \textbf{LH-MPJPE, RH-MPJPE} are evaluated on 15 hands joints on the left and right hands, respectively, after aligning the wrist joint. \textbf{F-MPJPE} is evaluated on 51 facial landmarks after aligning the neck joint. \textbf{FB-MPJPE} is a weighted sum of the above 4 errors. Since the number of hand joints and face landmarks outweigh the number of body joints, we define the FB error as $\text{FB} = \text{B}+(\text{LH+RH+F})/3$.

While 3D joint error evaluates pose, it does not provide evaluation of shape, for which we have ground truth. To encourage research on body shape estimation, we also evaluate the methods on vertices. 
We segment the body, left hand, right hand and face vertices using SMPL~\cite{SMPL:2015}, MANO~\cite{romero2017embodied} and FLAME~\cite{li2017learning} vertex indices of the SMPL-X template and calculate \textbf{B-MVE}, \textbf{LH-MVE}, \textbf{RH-MVE} and \textbf{F-MVE} respectively. \textbf{FB-MVE} uses the same weighted combination as joint error, FB-MPJPE.
We also calculate \textbf{B-NMJE}, \textbf{B-NMVE}, \textbf{FB-NMJE}, \textbf{FB-NMVE} and penalize the methods for missed detections and false positives.

\subsection{Evaluation protocol}\label{eva-protocol}
\label{subsec-ex-protocol}

When a method estimates a body, the matching ground truth body in AGORA is not known.
Therefore, to match the predicted person with the ground truth, we project the estimated 3D keypoints to the image plane and find the closest ground-truth subject in terms of 2D joint error.
If there is no match found for a particular ground truth body,
we count it as a miss (see Sup.~Mat.~for details).
Similarly, if a detection does not match any ground truth, we count it as
false positive. 
For the correctly matched predictions, we calculate all the errors as described in Sec.~\ref{datset-metric}.

\section{Experiments}

\begin{table}[]
    \centering
    \resizebox{0.35\textwidth}{!}{
\begin{tabular}{l|l|cccc}
\toprule
&\multirow{2}{*}{Method}  & \multicolumn{1}{c}{MPJPE (mm) $\downarrow$} & \multicolumn{1}{c}{MVE (mm) $\downarrow$} \\
	\cline{3-4}
& & \multicolumn{1}{c}{B} & \multicolumn{1}{c}{B}\\
\midrule
\multirow{5}{*}{\rotatebox[origin=c]{90}{~SMPL}}
&HMR~\cite{kanazawa2018end}        &  219.4 &  209.3 \\
&CenterHMR~\cite{sun2020centerhmr}  &  207.4 &  198.5\\
&EFT~\cite{joo2020exemplar}        &  202.7 &  193.5 \\
&SPIN~\cite{kolotouros2019spin}       &  203.7 &  193.2  \\
	\cline{2-4}
&SPIN-ft    &  \textbf{191.7} &  \textbf{186.7}  \\
\midrule
\multirow{3}{*}{\rotatebox[origin=c]{90}{~SMPL-X}} &SMPLify-X~\cite{SMPL-X:2019}  &  208.3 &  213.3  \\
&ExPose~\cite{choutas2020monocular}     &  \textbf{176.6} &  \textbf{174.0}  \\
&Frankmocap~\cite{rong2020frankmocap} &  203.7 &  204.2  \\
\bottomrule
\end{tabular}}
\vspace{-0.07in}
    \caption{Performance of SOTA methods on ``AGORA kids." }
    \label{tab:kidshape-baselines-evaluation}
\end{table}

We evaluate existing methods on AGORA to determine whether the dataset provides new insights about the current SOTA.
We also evaluate whether the AGORA training set can help improve the accuracy of SOTA methods by using it to fine-tune SPIN~\cite{kolotouros2019spin}.

\begin{figure}[t]
	\centerline{\includegraphics[width=\columnwidth]{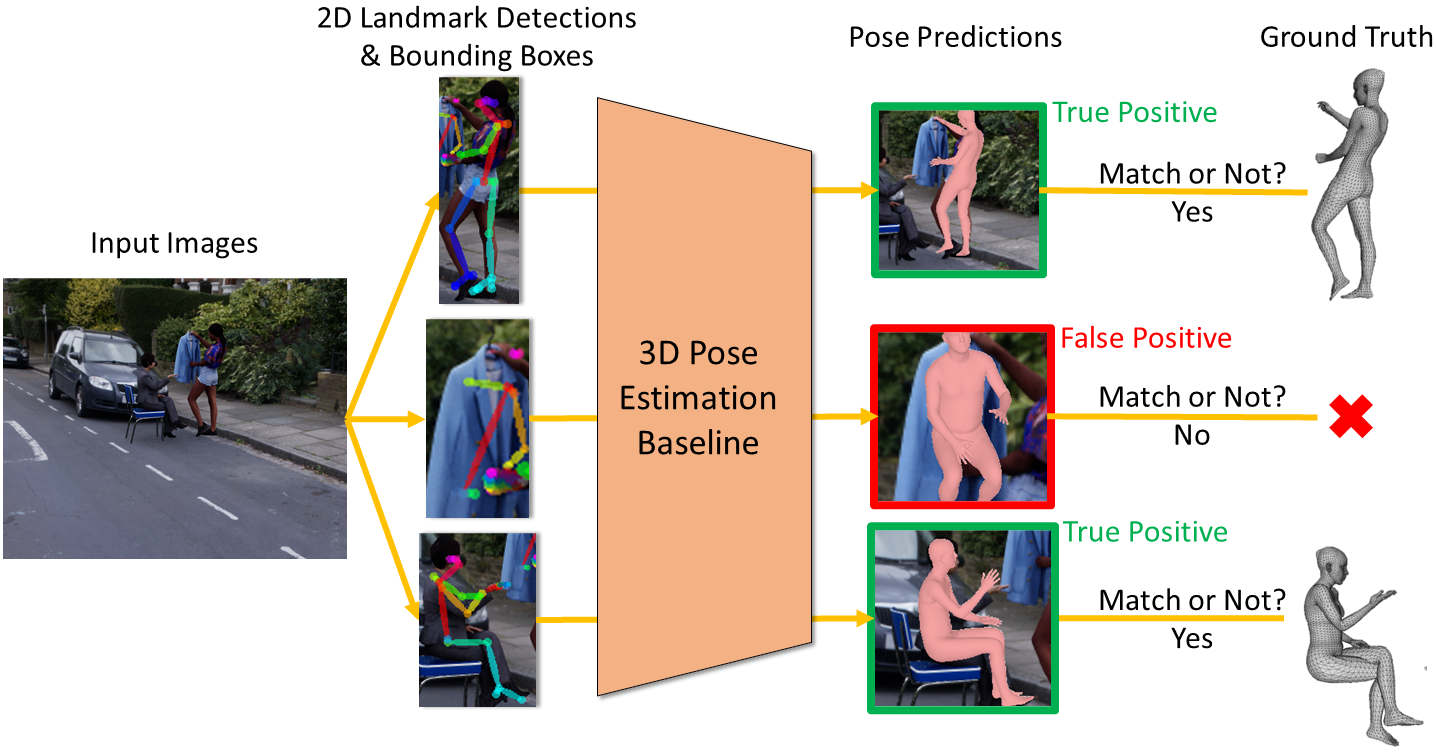}}
	\vspace{-0.1in}
	\caption{Baseline Evaluation. Given a test image, we detect keypoints with \cite{cao2019openpose} to obtain bounding boxes centered at each detected person, followed by network inference to reconstruct a human mesh for each cropped image. We identify true positives (to compute pose error), false negatives (misses) and false positives by matching predictions and ground truth. See Sec.~\ref{subsec-ex-protocol} for details.}
	\label{fig:Evaluation-Protocol}
\end{figure}

\subsection{Baseline Evaluation.}\label{baselines}
\begin{figure*}[t]
	\centerline{\includegraphics[width=0.33\textwidth]{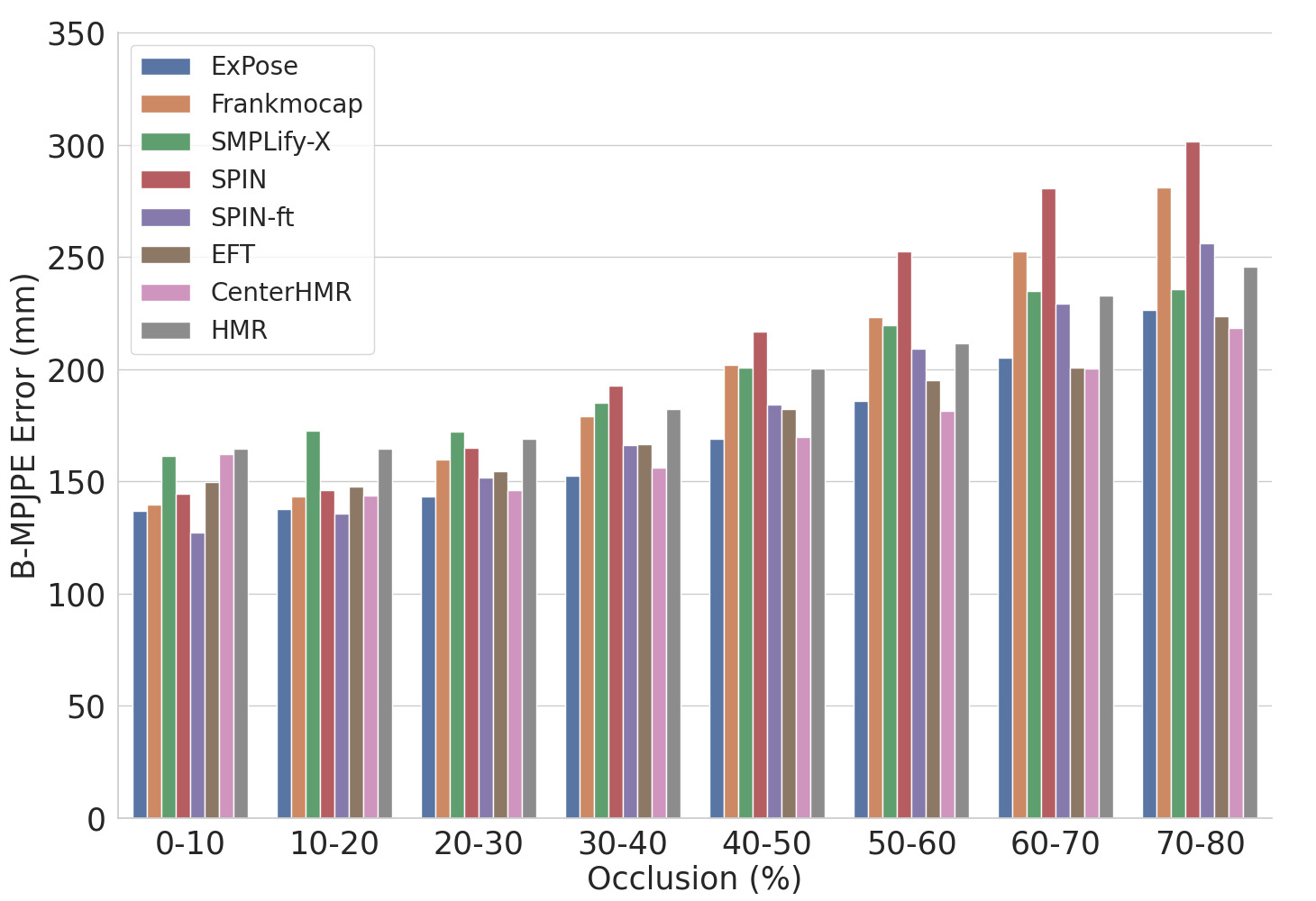}
		{\includegraphics[width=0.33\textwidth]{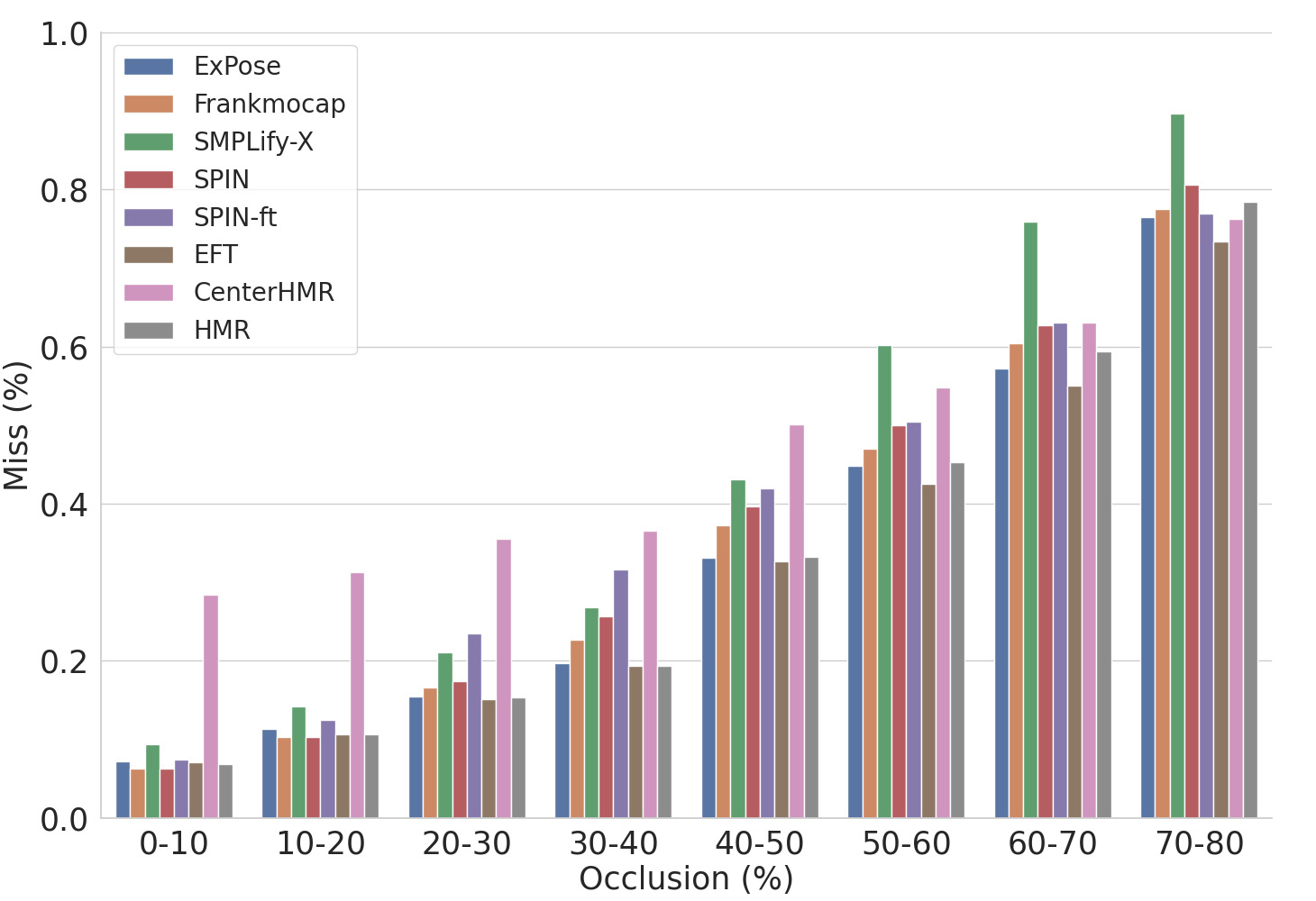}}	
		{\includegraphics[width=0.33\textwidth]{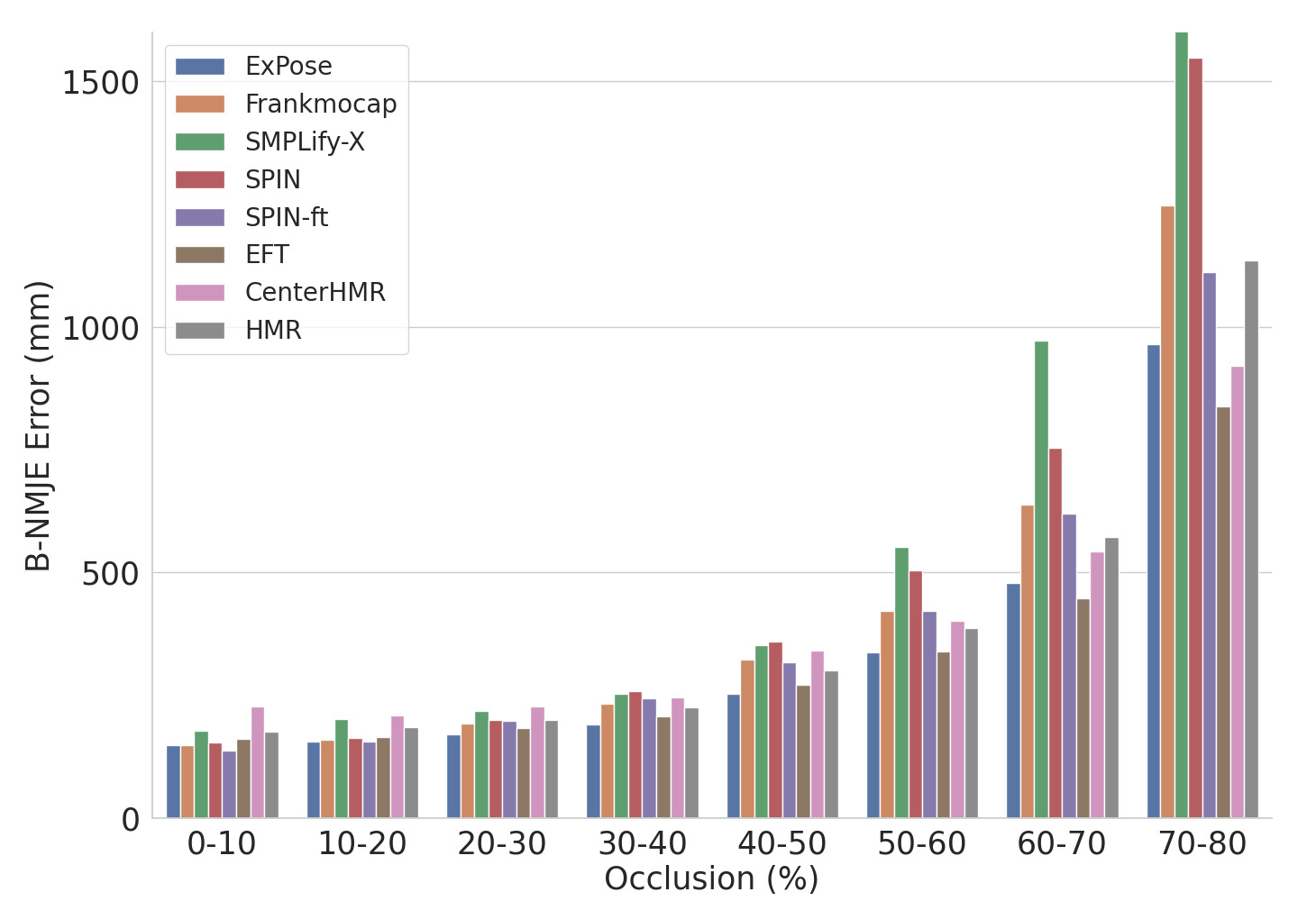}}}
\vspace{-0.1in}
	\caption{SOTA evaluation: B-MPJPE for the correct predictions (left), percentage of misses (center) and B-recall-NMJE for the correct predictions (right). Evaluated on BFH subset of AGORA for 22 SMPL-X and 24 SMPL joints.}
	\label{fig:Baseline-occlusion}
\end{figure*}

The evaluation protocol for existing methods is  shown in Fig.~\ref{fig:Evaluation-Protocol}. 
Most current methods assume that the input image is tightly cropped around the person~\cite{choutas2020monocular,joo2020exemplar,kanazawa2018end,kolotouros2019spin,rong2020frankmocap} or require 2D keypoint detections~\cite{SMPL-X:2019}.
Therefore, to fairly test prior methods on AGORA with the same input, we use OpenPose~\cite{cao2019openpose} to detect people and their respective
keypoints and construct tight bounding boxes based on these
detections. 
For single-stage approaches \eg~\cite{sun2020centerhmr} we directly use the entire image as input without any cropping.

Table~\ref{tab:baselines-evaluation} reports results for multiple baselines on the AGORA testset using the evaluation metrics described in Sec.~\ref{datset-metric}. To compare our new metrics with metrics used in earlier work, we also report MPJPE and MVE without penalizing for missed detections and false positives. See Fig.~7 in Sup.~Mat.~for qualitative results.
While SPIN fine-tuned on the AGORA training set (Sec.~\ref{finetuning}) outperforms other SMPL-based SOTA methods by a large margin in terms of MPJPE and MVE error, its error increases under our new NMJE and NMVE metrics because of misses and false positives. This shows that MPJPE alone is not enough to evaluate performance on multi-person images.
We hope AGORA will drive research on multi-person pose estimation.

We notice that among SMPL-X based methods, ExPose~\cite{choutas2020monocular} performs best for the body  while the optimization-based method SMPLify-X~\cite{SMPL-X:2019} beats regression based methods in hand and face estimation.
These errors are further analysed w.r.t.~different parameters like occlusion, child shape, distance to the center of the image and orientation (Sup.~Mat.~for orientation). 
\begin{table}[t]
	\resizebox{\columnwidth}{!}{
		\begin{tabular}{l|cc|cc|c}
			\toprule
			\multicolumn{1}{c|}{Models} & 
			\multicolumn{2}{c|}{3DPW (14) }&         
			\multicolumn{2}{c|}{3DPW (24)}&
			\multicolumn{1}{c}{AGORA (24)}  \\
			\midrule
			& \multicolumn{1}{l}{\footnotesize MPJPE} & \multicolumn{1}{l|}{\footnotesize PA-MPJPE} & 
			\multicolumn{1}{l}{\footnotesize MPJPE} & \multicolumn{1}{l|}{\footnotesize PA-MPJPE} & 
			\multicolumn{1}{c}{\footnotesize MPJPE}  \\
			\midrule
			SPIN-pt~\cite{kolotouros2019spin} &96.9 &59.3 &95.5&65.5&175.1 \\
			SPIN-ft-EFT 
			\cite{joo2020exemplar} & 97.4 & 59.7 & 95.3& 66.1 & 173.7 \\
			SPIN-ft (ours) & \textbf{85.7} & \textbf{55.3} & \textbf{83.7} & \textbf{61.8} & \textbf{153.4} \\
			\bottomrule						\end{tabular}
		}
		\vspace{-0.07in}
    \caption{Pretrained SPIN vs.~SPIN finetuned with AGORA and EFT([MPII+LSPet+COCO]). Parens.: (\#\/joints).}
	\label{tab:finetune-comparison}
\end{table}

\textbf{Occlusion.}
Using the ground-truth segmentation masks, Fig.~\ref{fig:Baseline-occlusion} plots the error of SOTA methods vs.~the percentage of occlusion.
Since this is analyzed on ground-truth bodies in which false positives are not included, we normalize the MPJPE by recall (correctly detected and matched bodies divided by total number of bodies), denoted as recall-NMJE and we also plot it for different ranges of occlusion.
As expected, the MPJPE for correct detections increases with increasing occlusion and the percentage of misses also increases as shown in the left and middle plots in Fig.~\ref{fig:Baseline-occlusion}. 
We observe that CenterHMR performs well for high occlusion but suffers from many misses, particularly with small people. See Fig.~7 in Sup.~Mat.
This shows that bottom-up methods that work on the full image are good in dealing with images of multiple people but need to improve their detection accuracy. FrankMocap and SPIN are highly sensitive to occlusion, leading to large errors as occlusion increases.
We also notice that fine-tuning with AGORA  improves the performance of SPIN for high occlusion cases.

\begin{figure}[t]
\centerline{	\includegraphics[width=0.4\textwidth]{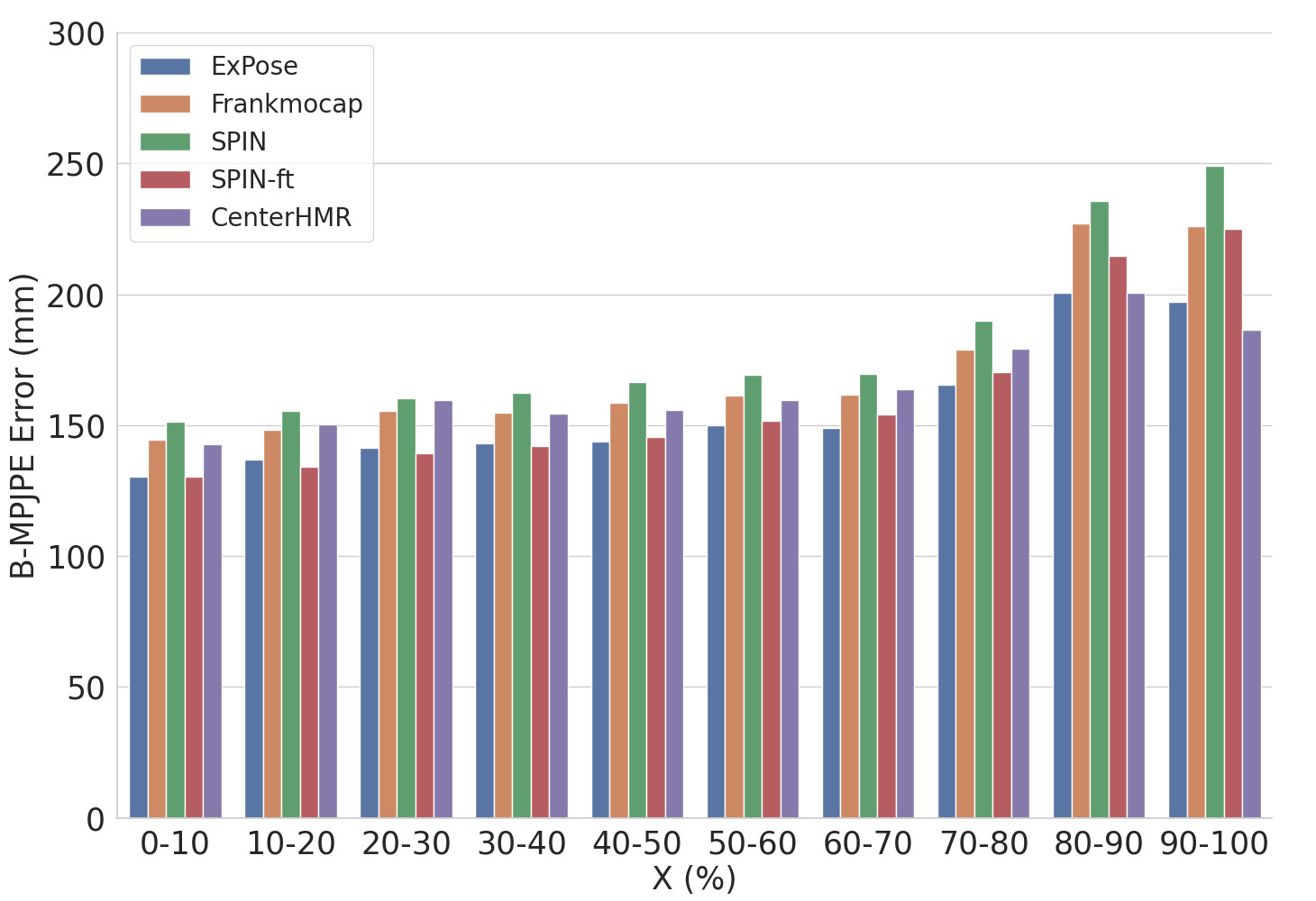}}
\vspace{-0.1in}
	\caption{Horizontal distance from center of the image: B-MPJPE for selected SOTA baselines. Evaluated on the BFH subset of AGORA for 22 SMPL-X and 24 SMPL joints.}
	\label{fig:Baseline-distance}
\end{figure}

\textbf{Distance from center.}
Most methods rely on a weak perspective camera assumption, which breaks when people occur off-center in images, as also pointed out by \cite{yu2020pcls}.
The large field-of-view images in AGORA facilitate the analysis of this error. We plot the B-MPJPE error of the selected SOTA methods vs.~horizontal distance from the center of the image in Fig.~\ref{fig:Baseline-distance}. 
We find that error consistently increases for all methods as the distance from the center increases.
This effect is less significant for CenterHMR~\cite{sun2020centerhmr}, the only method that works on full images instead of crops.

\textbf{Child Shape.}
AGORA contains child scans with  pseudo ground truth shape generated as described in Sec.~\ref{kid-scan-fit}. 
We calculate body joint and vertex error for the SOTA on a ``kids'' subset of AGORA. 
We find that performance is  significantly worse for predicting child shape compared to adult shape as shown in Table~\ref{tab:kidshape-baselines-evaluation}.
We hope that this, together with the child shape representation we provide, will encourage work in this direction.

\subsection{Baseline Improvement.}\label{finetuning}
To evaluate the efficacy of the AGORA training set, we fine-tune
a pretrained SPIN model (SPIN-pt) using only crops from AGORA training images and refer to the fine-tuned model as SPIN-ft. 
We chose SPIN for the fine-tuning experiment as it uses HMR as a backbone, which is a base for many 3DHPS methods, e.g.~EFT \cite{joo2020exemplar}, VIBE~\cite{kocabas2019vibe} and FrankMocap~\cite{romero2017embodied}.
For fair comparison, we use the same hyperparameters and loss functions as SPIN. However, unlike SPIN, we do not use SMPLify in loop, replacing that supervision with AGORA ground truth. 

While AGORA images are rendered using perspective cameras, SPIN assumes a weak perspective camera, which is unable to capture perspective warping, especially when people occur off-center in images (see Sec.~\ref{baselines}). This makes the global orientation of the ground-truth 3D joints and vertices in AGORA inconsistent with SPIN predictions. During SPIN fine-tuning, we therefore set global orientation to zero before calculating all 3D losses, such that information about global orientation comes only from the 2D keypoint loss. 
We note that this is required only due to the weak perspective camera assumption in SPIN and we hope AGORA will encourage research  with more realistic perspective camera models. 

Since SPIN reports Procrustes aligned MPJPE (PA-MPJPE) on 3DPW, we compare SPIN-pt and SPIN-ft on both MPJPE and PA-MPJPE. 
We also compare the AGORA training set with the EFT dataset, [MPII+LSPet+COCO]\textsubscript{EFT}, 
We call fine-tuning on EFT, SPIN-ft-EFT.
We evaluate SPIN-pt, SPIN-ft-EFT and SPIN-ft on the 3DPW testset with known association and on AGORA testset without known association. Training with AGORA leads to significant improvement in performance on both datasets, with MPJPE improving by $\sim$12\% for 3DPW and $\sim$13\% for AGORA; see Table~\ref{tab:finetune-comparison}. 
Higher MPJPE on AGORA compared to 3DPW also shows that AGORA is more challenging than 3DPW.
Note that we calculate the error using 14 joints to compare with original SPIN-pt results.

\section{Conclusions and Future Work}
We have presented AGORA, a new dataset that goes beyond current datasets to include challenging cases of environmental occlusion, person-person occlusion, scale variation, children, crowds, etc.
AGORA is challenging and reveals limitations of existing methods.
Despite being synthetic, fine-tuning on AGORA improves performance of a SOTA method on the natural 3DPW dataset. 
We introduce a new metric to include misses and false positives and facilitate analysis of the SOTA methods on images with multiple people. We also introduce a simple child body model and provide better 3D ground truth for images with children. 
Future work should include adding images of varied camera height, indoor scenes, multi-view images, larger crowds, animals, and movement. 

{\small
\noindent\textbf{Acknowledgements.} We thank Galina Henz, Taylor McConnell and Tsvetelina Alexiadis for the help in data labeling. 
\noindent\textbf{Disclosure.} MJB has received research gift funds from Adobe, Intel, Nvidia, Facebook, and Amazon. While MJB is a part-time employee of Amazon, his research was performed solely at, and funded solely by, Max Planck. MJB has financial interests in Amazon, Datagen Technologies, and Meshcapade GmbH.
}

{\small
\bibliographystyle{ieee_fullname}
\balance
\bibliography{agorabib}
}

\newpage
\appendix
{\noindent\Large\textbf{Supplementary Material}}
\newline
\setcounter{page}{1}
\begin{figure*}[t]
	\centerline{\includegraphics[width=0.85\textwidth]{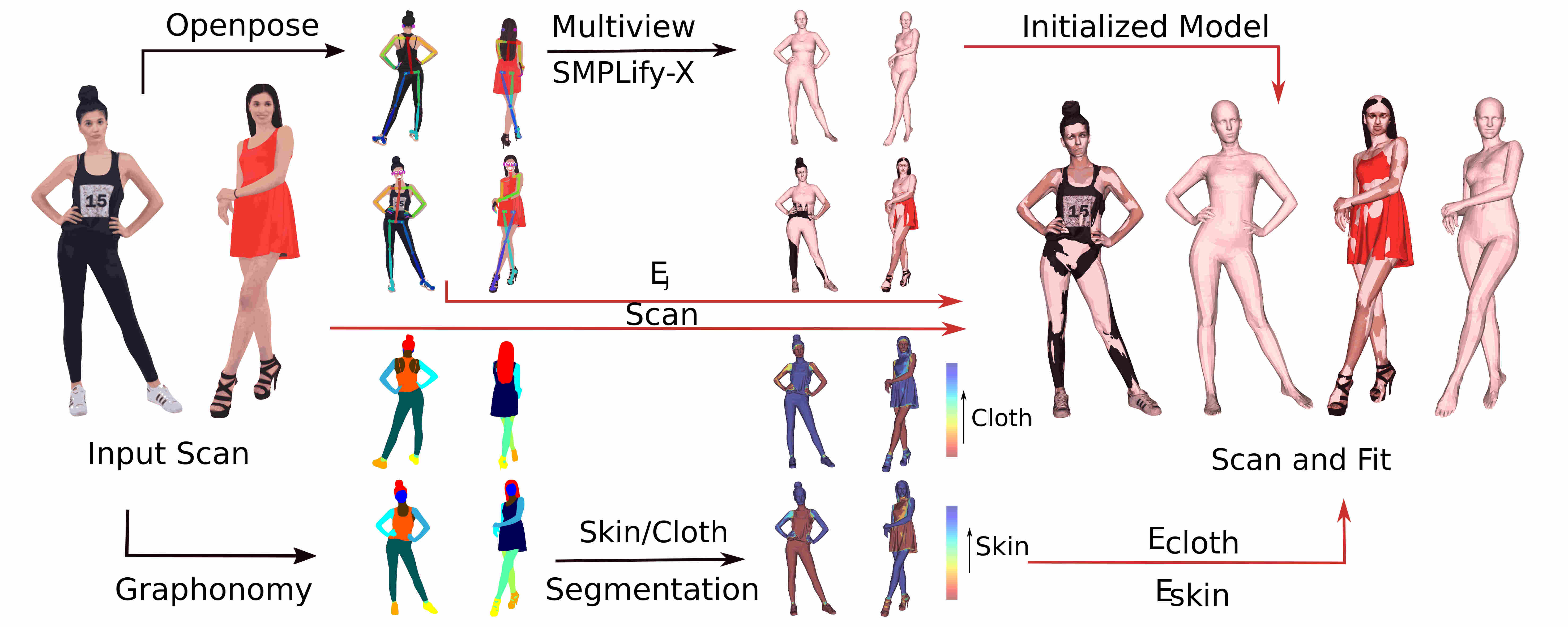}}
	\vspace{-0.5em}
	\caption{\textbf{SMPL-X fitting to scans.} Keypoints and body part segmentation across multiple rendered views are generated using OpenPose and Graphonomy. Multiview SMPLify-X initializes the model in proper pose. Shape is further refined using 2D+3D refinement (see text).}
	\vspace{-1em}
	\label{fig:modelonly-fitting}
\end{figure*}
\section{Method}

Here we further explain the method section of the main paper in detail.
We initialize the parameters of the SMPL-X~\cite{SMPL-X:2019} model with a multi-view fitting approach, followed by a refinement step that fits SMPL-X to the 3D scan as show in Fig.~\ref{fig:modelonly-fitting}.

\subsection{Multi-view Initialization.}\label{subsec-multiview}
The scans contain arbitrary poses, varied clothing, and people holding objects. 
This makes the automatic fitting of SMPL-X a challenge without a good initialization.
We first center $\scanpoints$ and render images of it from $C$ pre-defined virtual cameras. 
2D landmarks are detected in each rendered image with~\cite{cao2019openpose} and we initialize the parameters with an approach that extends the single-view SMPLify-X fitting \cite{SMPL-X:2019} to incorporate landmarks in multiview images. 

SMPLify-X~\cite{SMPL-X:2019} takes one color image as input and optimizes the pose $\theta$, shape $\beta$ and facial expression $\psi$ of SMPL-X to match the observed 2D landmarks by minimizing the following objective:
\begin{eqnarray}
E(\beta,\theta,\psi) & = & E_J + E_{\text{reg}}, \\ \label{eq:objective-smplifyx}
E_{\text{reg}} & = & \lambda_{\theta_b} 		E_{\theta_b}  + \lambda_{\theta_h} 		E_{\theta_h} + 
\lambda_{\alpha}		 	E_{\alpha}			+  \nonumber\\
& &	\lambda_{\beta}    		E_{\beta}	+ 	\lambda_{\mathcal{E}}    		E_{\mathcal{E}}			+	\lambda_{\mathcal{C}}    		E_{\mathcal{C}},
\end{eqnarray}
where $E_J$ is the data term that penalizes differences between projected and observed landmarks, and $E_{\text{reg}}$ includes several regularization terms: $E_{\alpha}(\theta_b)$ penalizes strong bending of elbows and knees, while $E_{C}$ prevents mesh-intersections. 
$E_{\theta_b}(\theta_b)$, $E_{\theta_h}(\theta_h)$, $E_{\beta}(\beta)$, and $E_{\mathcal{E}}(\psi)$ are $L_2$ priors on the body pose, hand pose, body shape and facial expressions. $\lambda$'s denote  weights for each respective term. 
We adapt Eq.~\ref{eq:objective-smplifyx} to take multi-view data with known camera parameters for each camera $c$: $E_{\text{mv}} = E_{\text{reg}} + \sum_{c=1}^{C} E_J^c$. 
Unlike in~\cite{SMPL-X:2019} where one needs to estimate camera translation first, here the intrinsics and extrinsics are given.

\subsection{2D+3D Refinement}

To get the skin and cloth vertices from scan, we use segmentation masks provided by Renderpeople to group scan points into skin, clothing (including shoes), and the rest (hair and objects). Since we do not have segmentation masks from other vendors, we generate them using Graphonomy \cite{gong2019graphonomy}. 
Graphonomy provides human parts segmentation given an image with labels for cloth as well as body parts.

\noindent\textbf{Skin-Cloth Segmentation.} 
For each rendered multi-view image of a scan, Graphonomy outputs segmentation masks for different body parts and types of clothing. 
We group these  into 3 labels: skin, cloth and other. 
Given the known cameras, we project visible vertices into the images and give them the corresponding label.
Aggregating  labels across all views gives a us a probability of each vertex being skin, clothing, or other.
For Renderpeople scans, the probability is either 1 or 0 as we have segmentation masks.
Similar to~\cite{zhang2017detailed}, we define energy terms $\skinterm$ and $\clothterm$ for skin and clothing scan vertices, respectively. 

Here we explain in detail, the two optimization terms, $\skinterm$ and $\clothterm$ used in 2D+3D refinement. Please refer to the main paper for the full equation.

\noindent\textbf{Skin term}.
For each scan vertex $s \in S$ we find the point on the closest model triangle. We minimize the point-to-surface distance between them weighted by probability $p(s) \in P_\text{skin}$. Here $P_\text{skin}$ is the probability  of the vertex $s$ belonging to skin calculated using Graphonomy~\cite{gong2019graphonomy}.
\begin{equation}
\begin{split}
&E_\text{skin}(\beta,\theta, \psi) = \sum_{s \in S}\rho \left( \sqrt{p(s)} \cdot \text{dist}(s, M(\beta, \theta, \psi))\right),
\label{eq:objective-skin_term}
\end{split}{}
\end{equation}
where $\text{dist}(\cdot)$ represents the distance of the closest point on the model $M(\beta, \theta, \psi)$ surface from the scan vertex $s$. $\rho(\cdot)$ is Geman-McClure robust error function~\cite{geman1987statistical}  that prevents outliers from contributing too much in the energy.
Of course, we start with the initialization from Sec.~\ref{subsec-multiview}.

\noindent\textbf{Clothing term}.
The goal of $\clothterm$ is to prevent clothing scan points from penetrating inside the model while keeping the model close to the scan, so that the body does not shrink. 

Each scan point is further classified into two categories: points penetrating the body model $S_P$ and points outside the body model $S_O$. We get the probability of each scan vertex being cloth $P_\text{cloth}$ from Graphonomy~\cite{gong2019graphonomy}, $p(s) \in P_\text{cloth}$.
For $s \in S_P$ we penalize the distance with weight $\lambda$, while for $s \in S_O$  we use again Geman-McClure function to accommodate loose clothes like skirts, saris, bath robes, etc. 
Specifically:
\begin{eqnarray}
\lefteqn{\clothterm(\beta,\theta; \psi)  =  }\nonumber\\
&& \sum_{s \in S_O} \rho \left(\sqrt{p(s)} \cdot \dist(s, M(\beta, \theta, \psi))\right)\nonumber\\
& & + \lambda \sum_{ s \in S_P} p(s) \cdot \dist^{2}\left(s, M(\beta, \theta, \psi)\right),
\label{eq:objective-cloth_term}
\end{eqnarray}
where we do not optimize facial expression, $\psi$, because it is not covered by clothing and where
$\dist(\cdot)$ and $\rho$ are the same as in Eq.~\ref{eq:objective-skin_term}. 

\begin{figure}[t]
	\centerline{\includegraphics[width=\columnwidth]{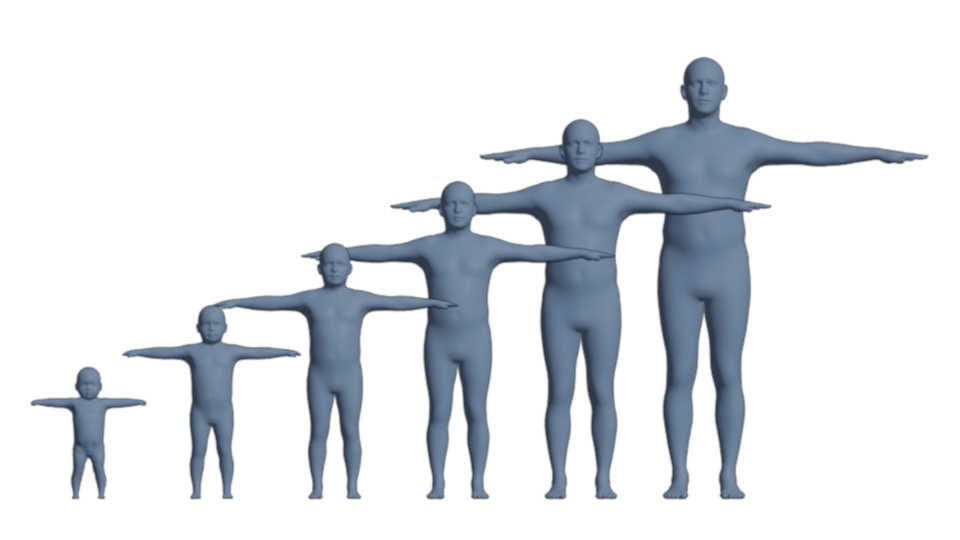}}
	\vspace{-1em}
	\caption{SMIL-X and adult male SMPL-X template interpolation.}
	\label{fig:interpolation}
\end{figure}

Since the initialization in Sec.~\ref{subsec-multiview} is already close, the classification of $S_P$ and $S_O$ can be approximated as follows.
Each vertex $s \in S$ has a point $m$ on the nearest triangle of the model with a corresponding normal $n$. 
We define a displacement vector $d = m - s$ and consider it $S_P$ if the inner product of $d$ and $n$ is greater than $0$, otherwise $S_O$.
\begin{figure}[t]
    \centering
    \includegraphics[trim=20 0 50 40,clip,width=0.235\textwidth]{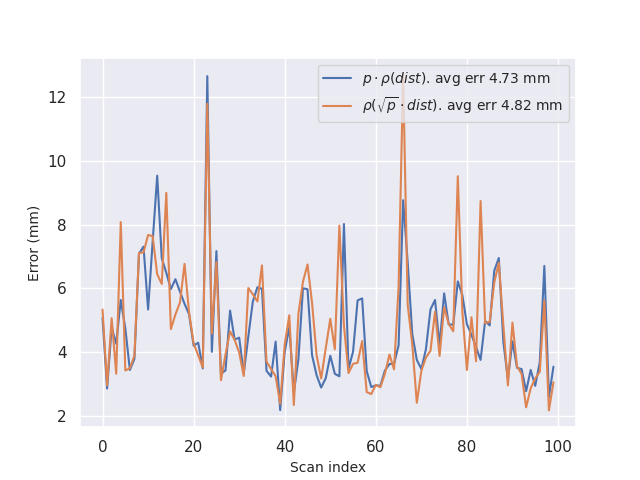}	
    \includegraphics[trim=20 0 50 40,clip,width=0.235\textwidth]{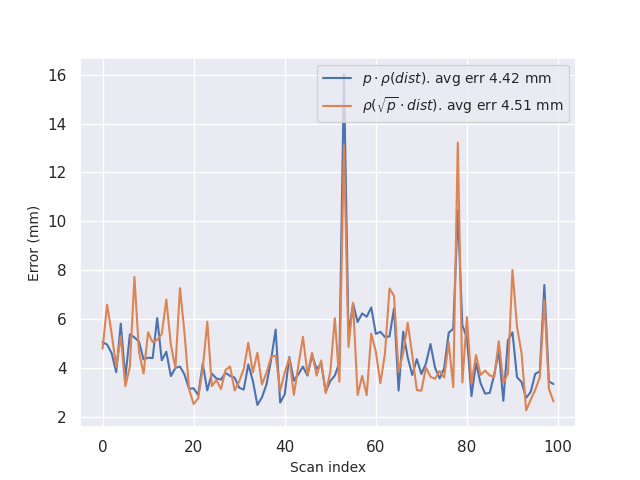}		
    \caption{Fitting errors made by different formulations of $E_\text{skin}$ and $E_\text{cloth}$.}
    \label{fig:feature-analysis}
\end{figure}

\begin{figure*}
    \centering
    \includegraphics[width=0.32\textwidth]{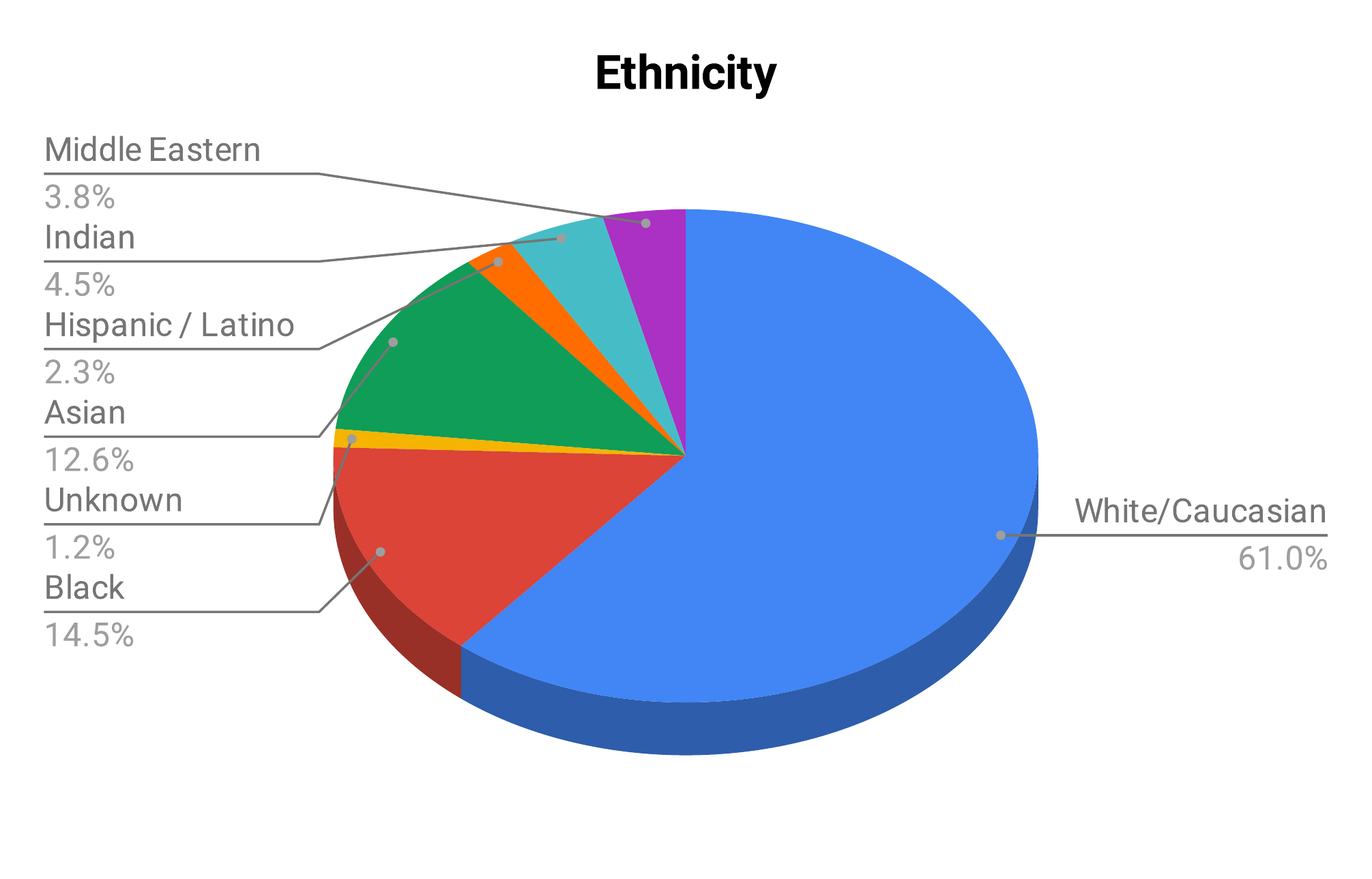}	
    \includegraphics[width=0.32\textwidth]{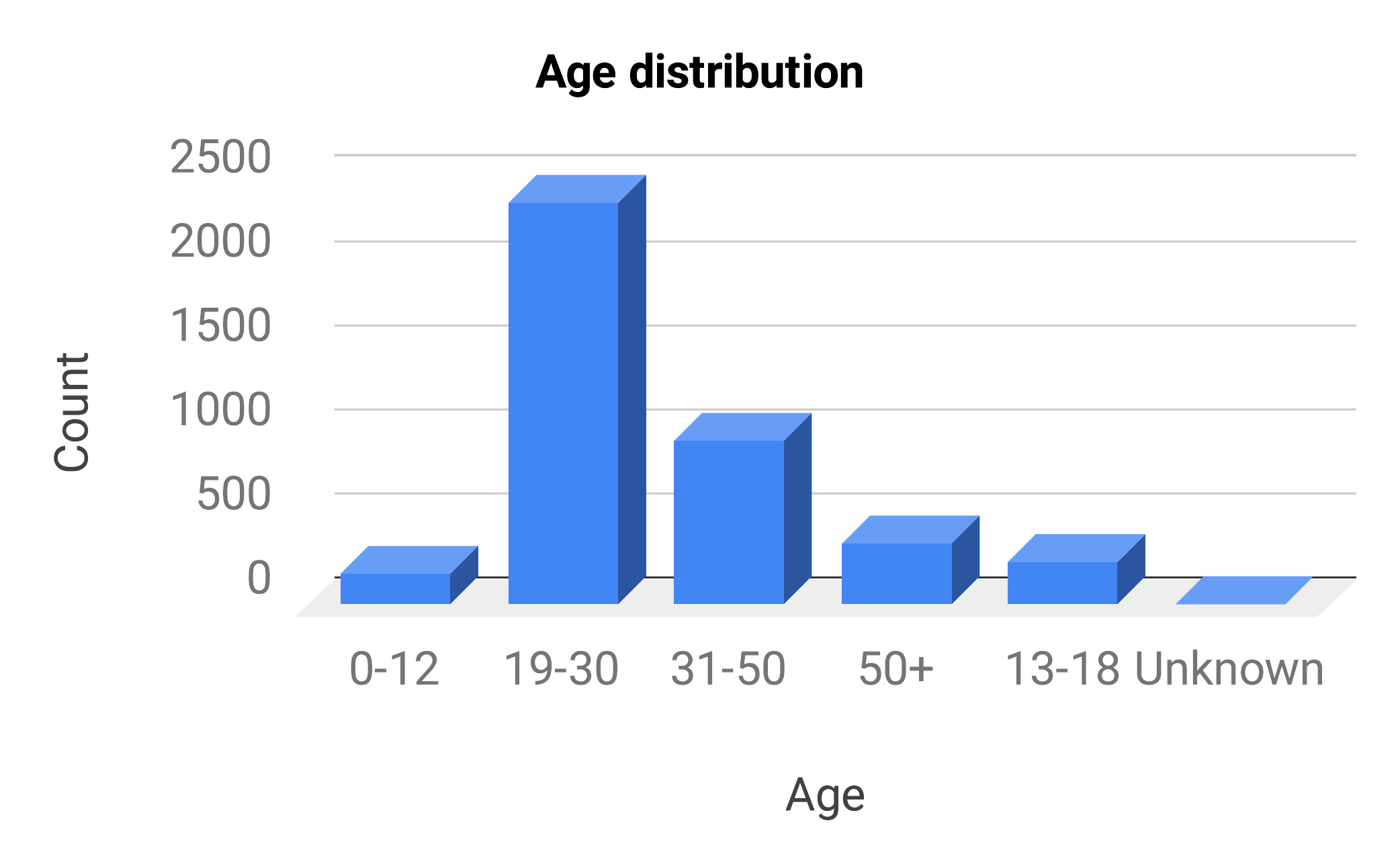}		
   \includegraphics[width=0.32\textwidth]{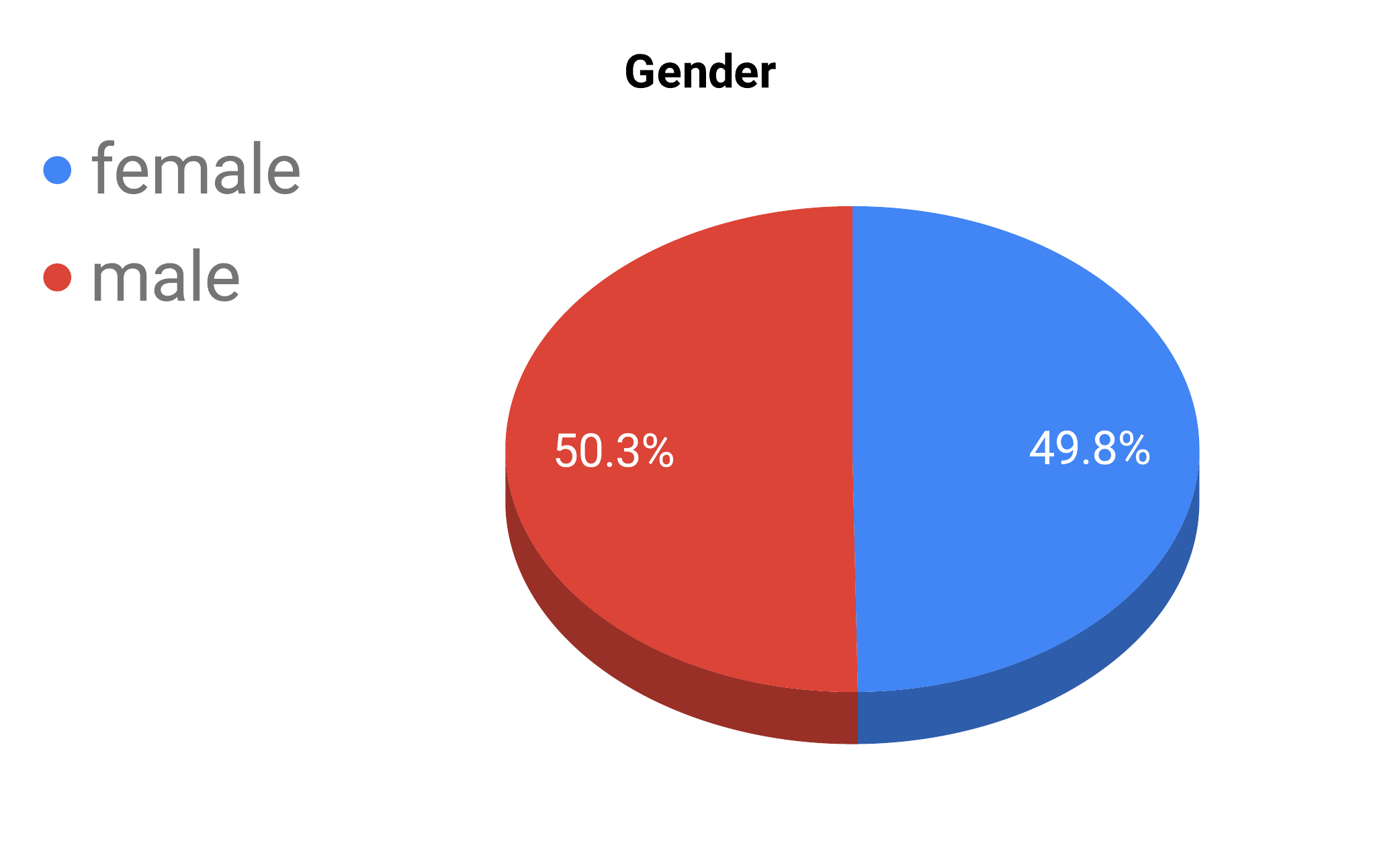}
    \caption{Breakdown of AGORA dataset in ethnicity, age, and gender.}
    \label{fig:statistics}
\end{figure*}

\noindent\textbf{Discussion}.
We compare the formulation of Eq.~\ref{eq:objective-skin_term} and \ref{eq:objective-cloth_term} with another design choice, 
where the skin-cloth probability $p(s)$ is multiplied outside the Geman-McClure robust function without square root, \ie~$p \cdot \rho(\text{dist})$.
For scans that have vendor-provided skin-cloth segmentation masks, the probability $p(s)$ is either 1 or 0, so this formulation is equivalent to Eq.~\ref{eq:objective-skin_term} and \ref{eq:objective-cloth_term}.
For rest of scans, we fit 100 examples with this formulation and report in average 4.73mm skin error and 4.42mm cloth error following the definition in Sec.~\ref{subsec-ex-fittting} of the main paper.
The corresponding errors using Eq.~\ref{eq:objective-skin_term} and \ref{eq:objective-cloth_term} are 4.82mm and 4.51mm respectively. 
Fig.~\ref{fig:feature-analysis} further shows the error comparison for each scan.
We see that while Eq.~\ref{eq:objective-skin_term} and \ref{eq:objective-cloth_term} yield higher errors, the difference are negligible.

\subsection{Child scans fitting}
As described in Sec.~\ref{kid-scan-fit} of the main paper, we fit 257 children scans by using a template that is an interpolation of adult SMPL-X template and SMIL infant template~\cite{hesse2018learning}. Fig.~\ref{fig:interpolation} shows how varying interpolation coefficient gives us approximate template from different age group.

	\section{AGORA}
\begin{figure}
	\centerline{\includegraphics[width=0.5\textwidth]{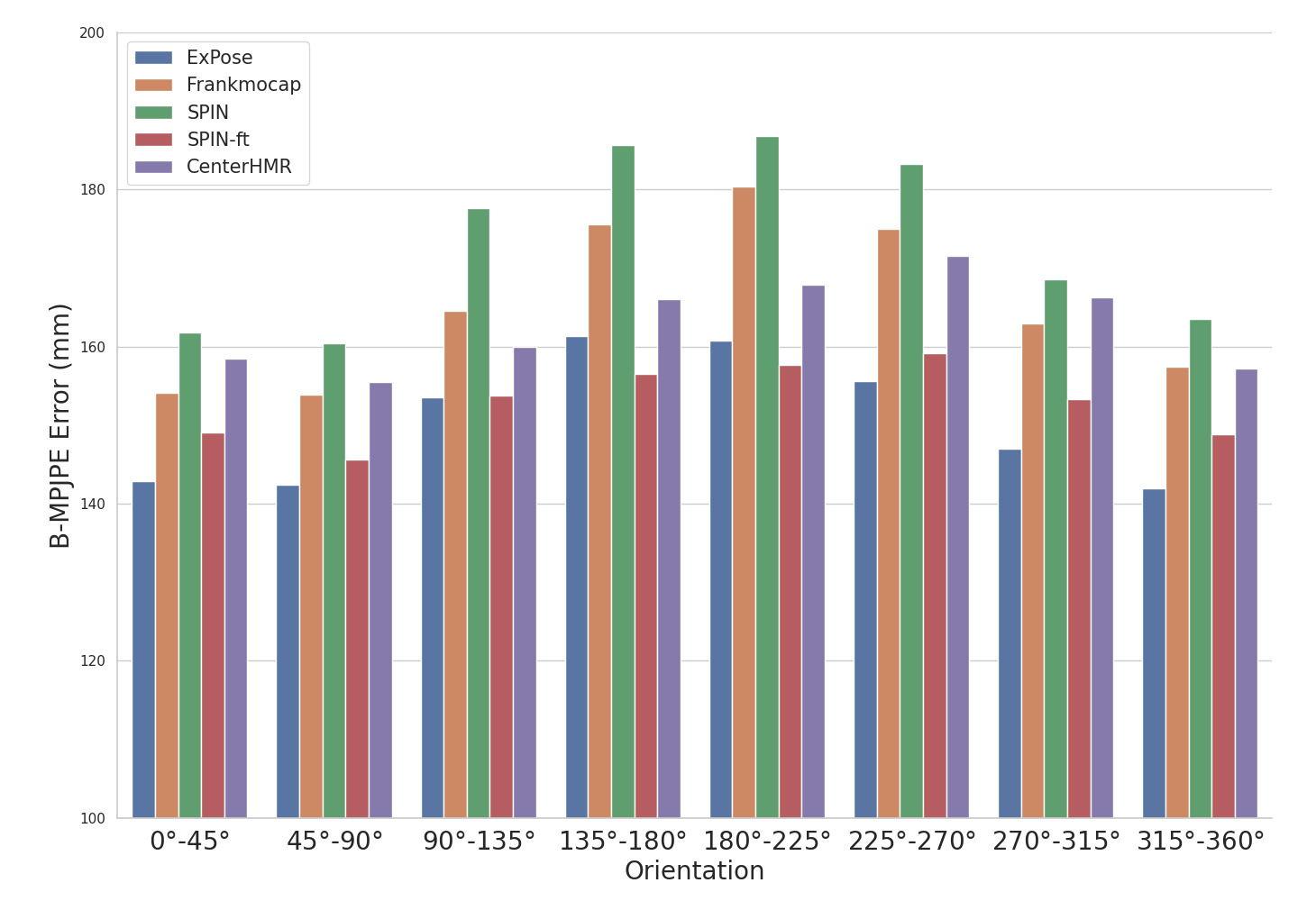}}
	\vspace{-1em}
	\caption{Error in varied orientations relative to the camera. $0^\circ$ corresponds to facing the camera. Evaluated on BFH subset of AGORA for 22 SMPL-X and 24 SMPL joints.}
	\vspace{-1em}
	\label{fig:Baseline-orientation}
\end{figure}
\begin{figure*}
	\centerline{\includegraphics[width=0.33\textwidth]{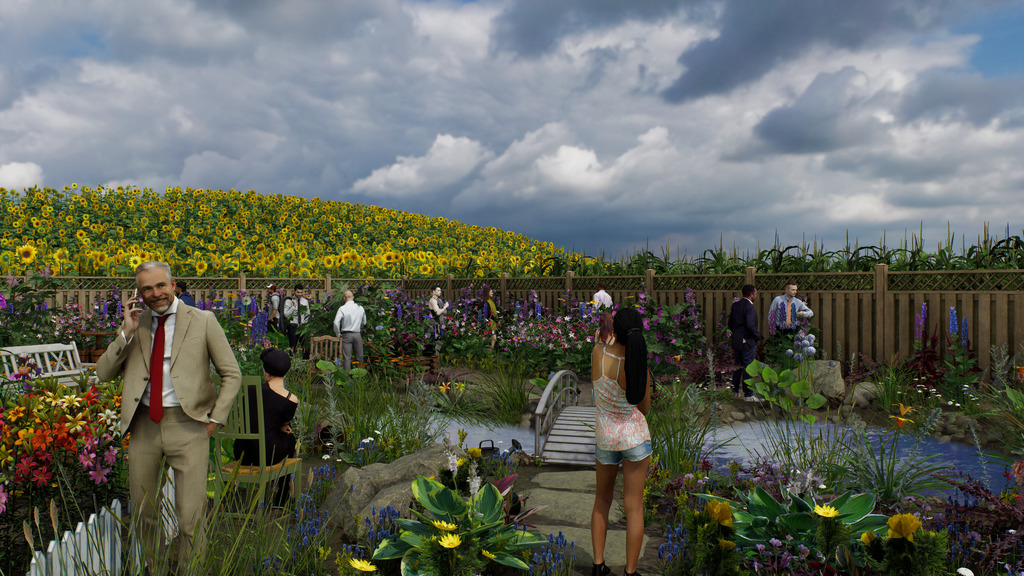}
		\includegraphics[width=0.33\textwidth]{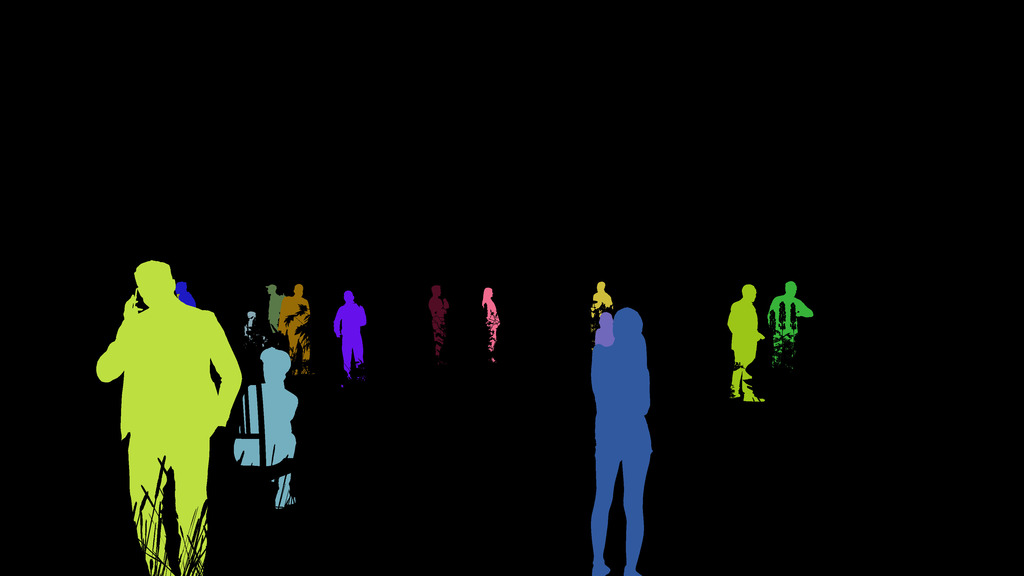}
		\includegraphics[width=0.33\textwidth]{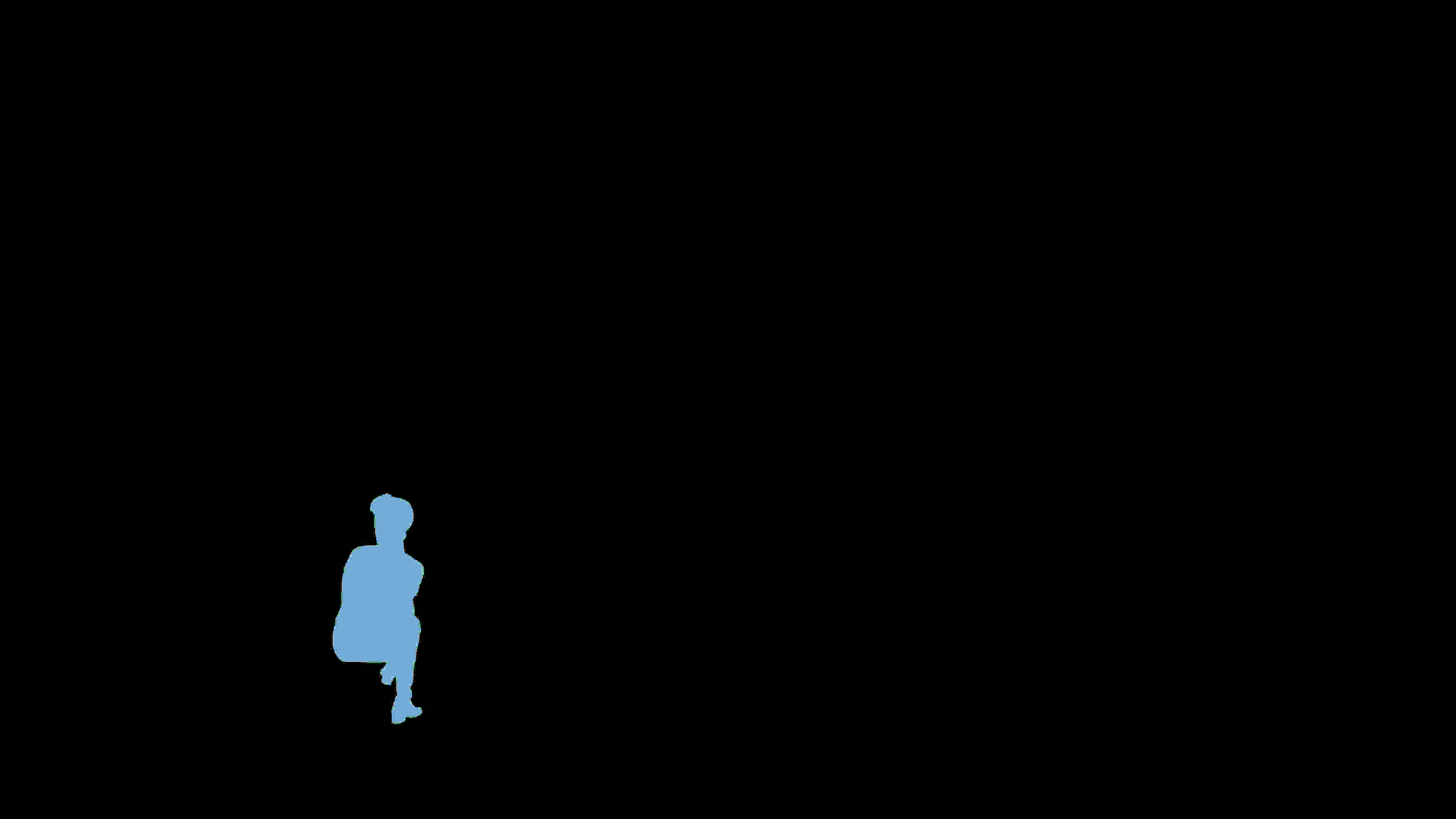}
	}
	
	\centerline{\includegraphics[width=0.33\textwidth]{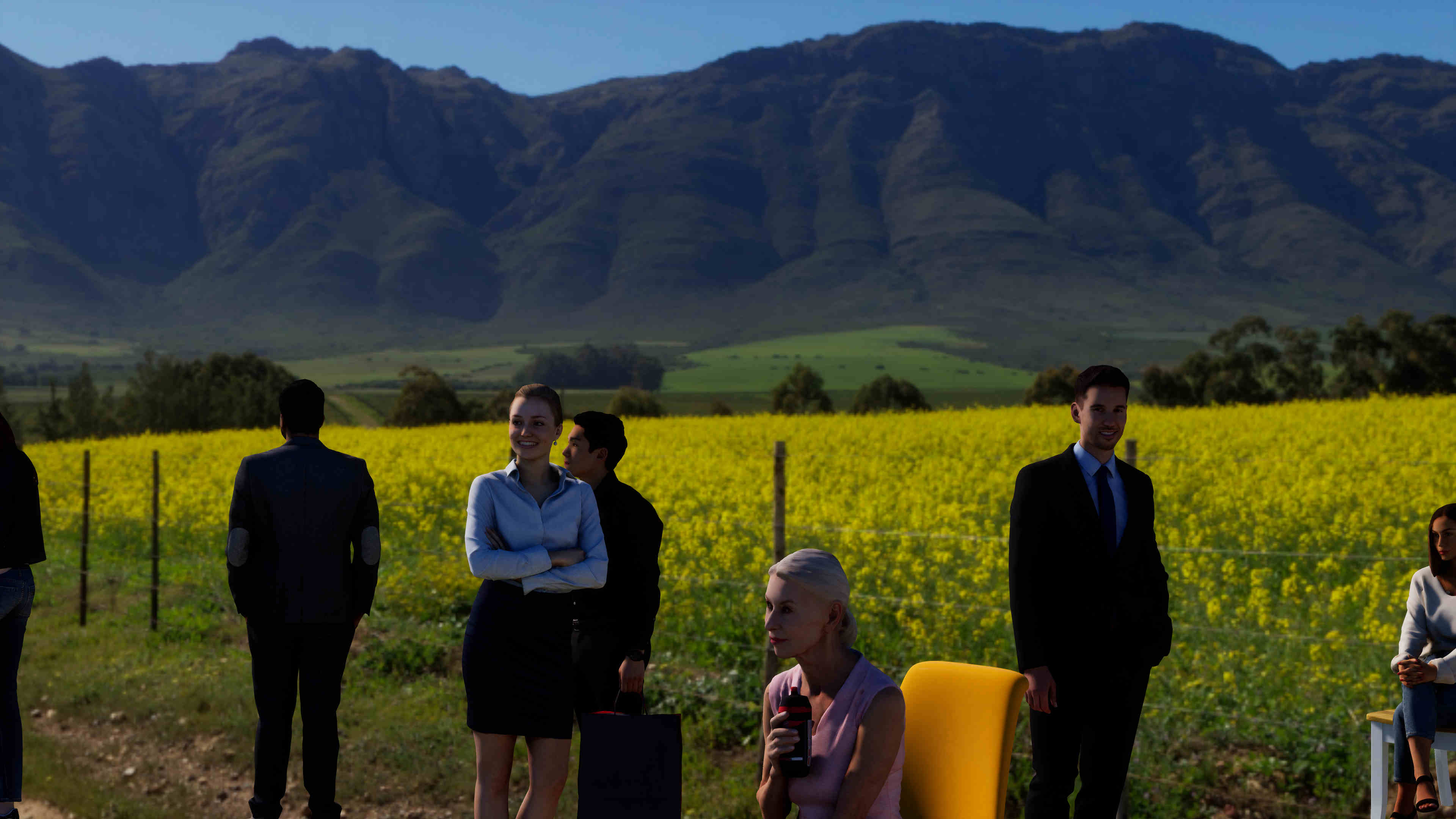}
		\includegraphics[width=0.33\textwidth]{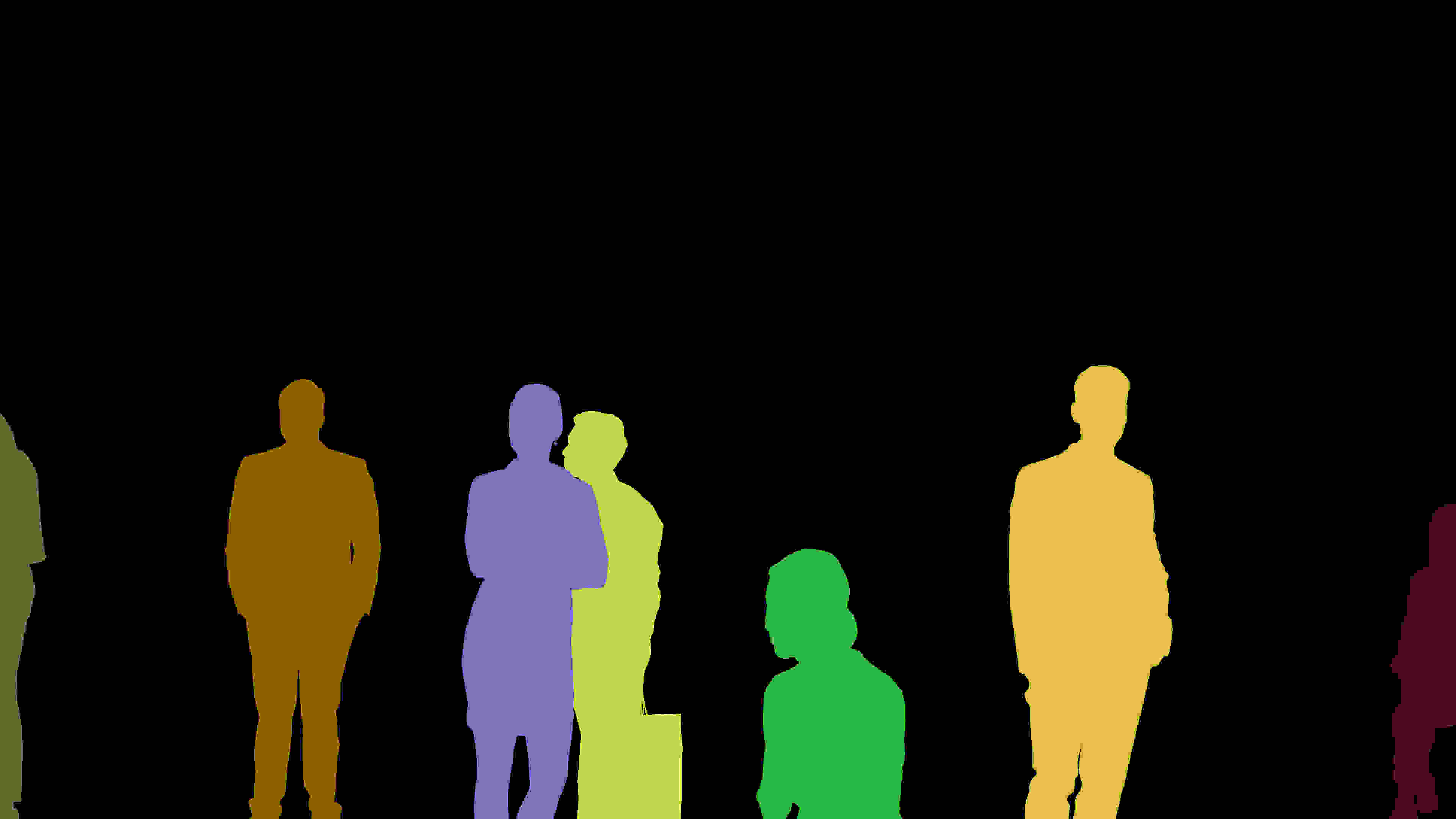}
		\includegraphics[width=0.33\textwidth]{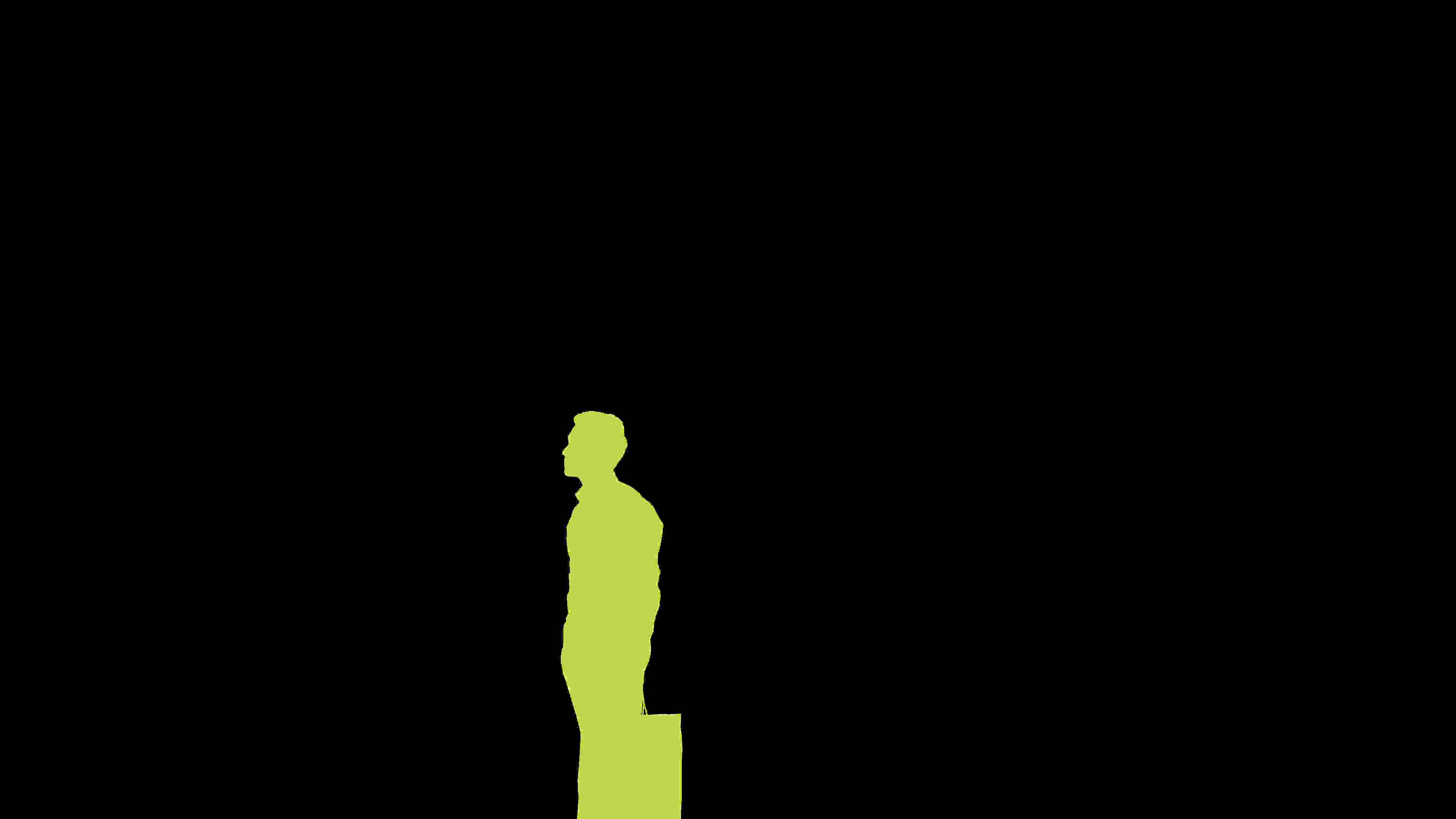}
	}
	\vspace{-0.5em}
    \caption{Person segmentation masks. Left: color images. Middle: full masks. Right: individual masks rendered with no occlusions.}
    \vspace{-1em}
    \label{fig:indi-mask}
\end{figure*}

\begin{figure*}
	\centerline{\includegraphics[width=0.32\textwidth]{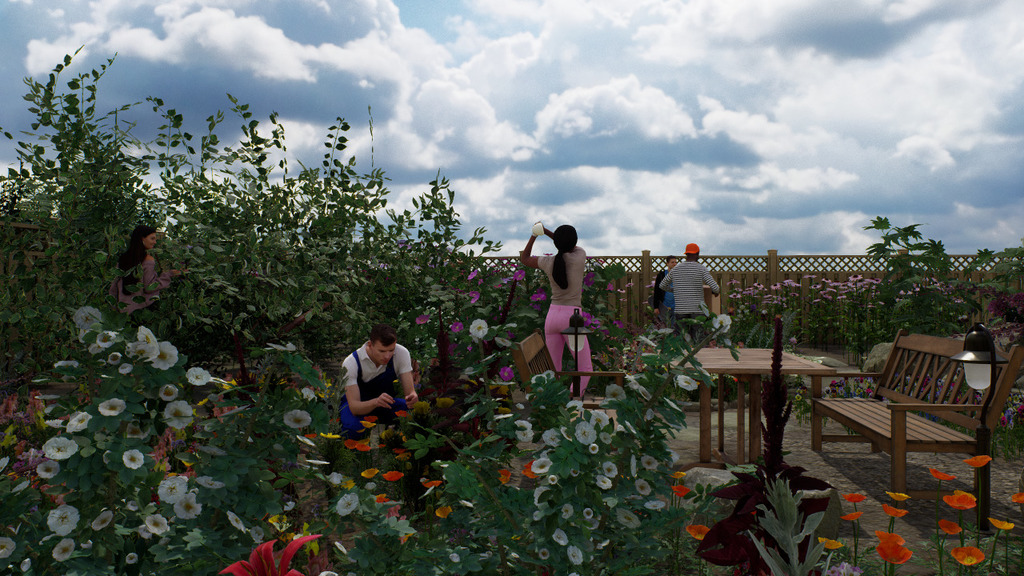}		
		\includegraphics[width=0.32\textwidth]{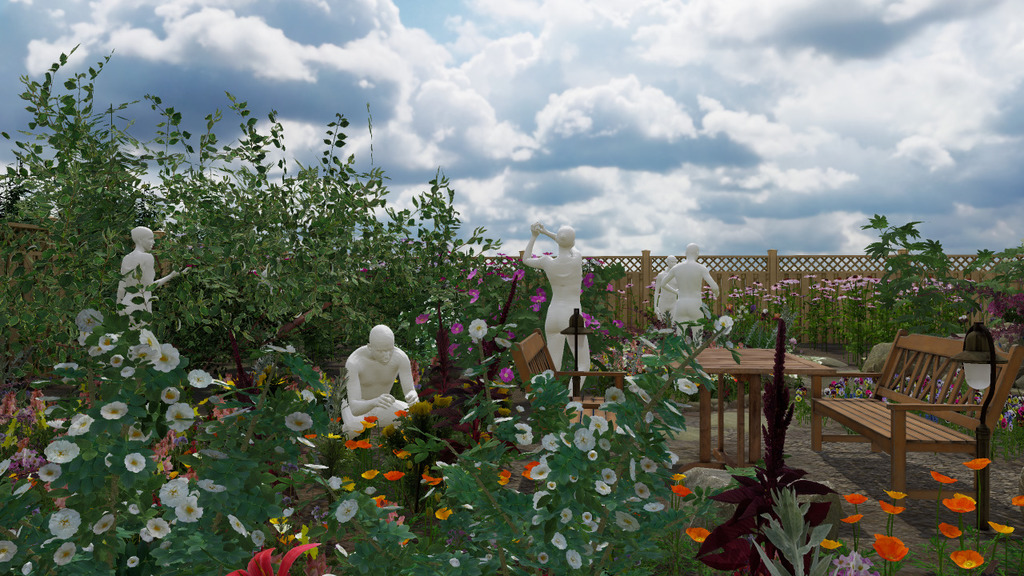}		
		\includegraphics[width=0.32\textwidth]{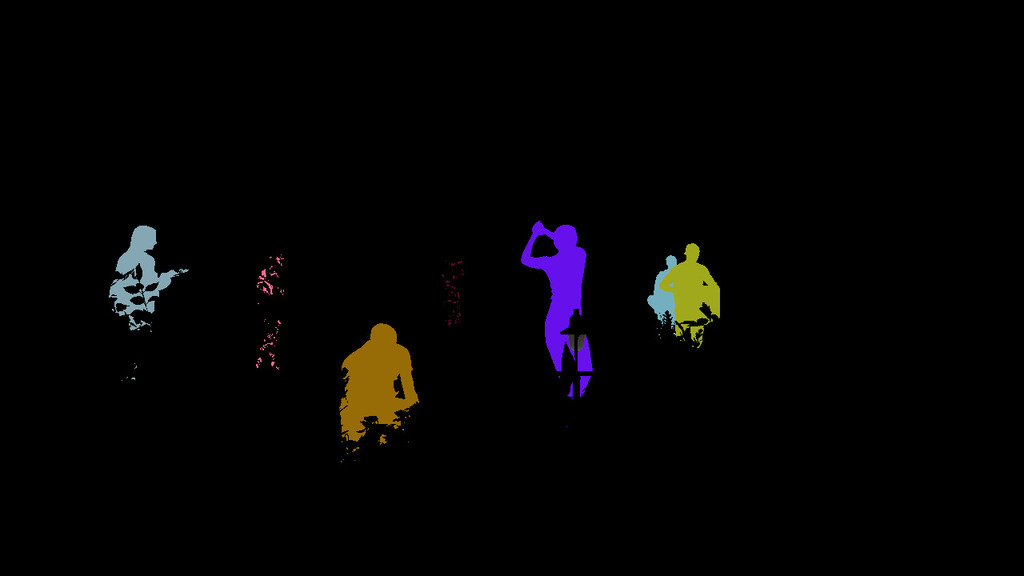}		
	}
	\centerline{\includegraphics[width=0.32\textwidth]{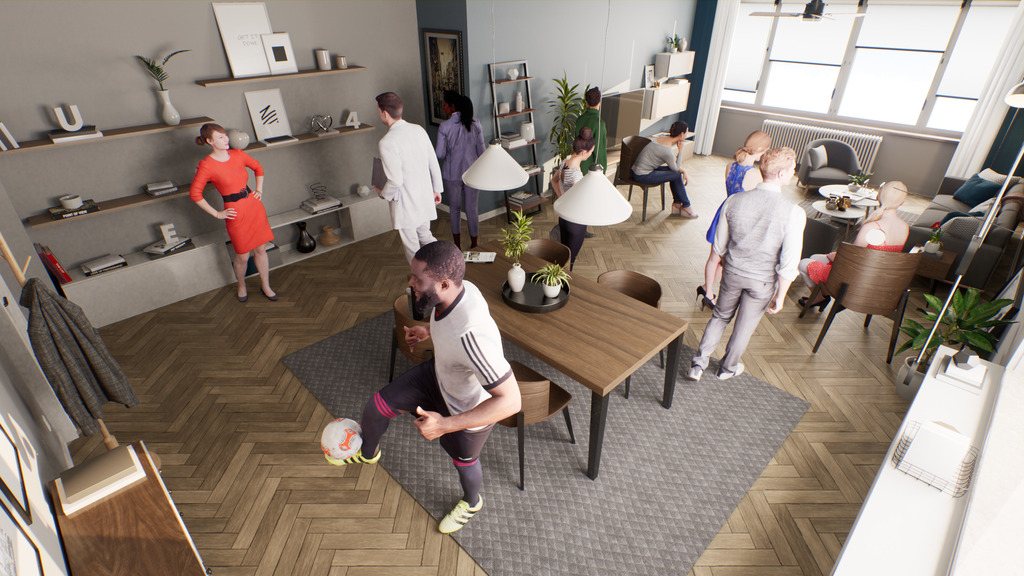}		
		\includegraphics[width=0.32\textwidth]{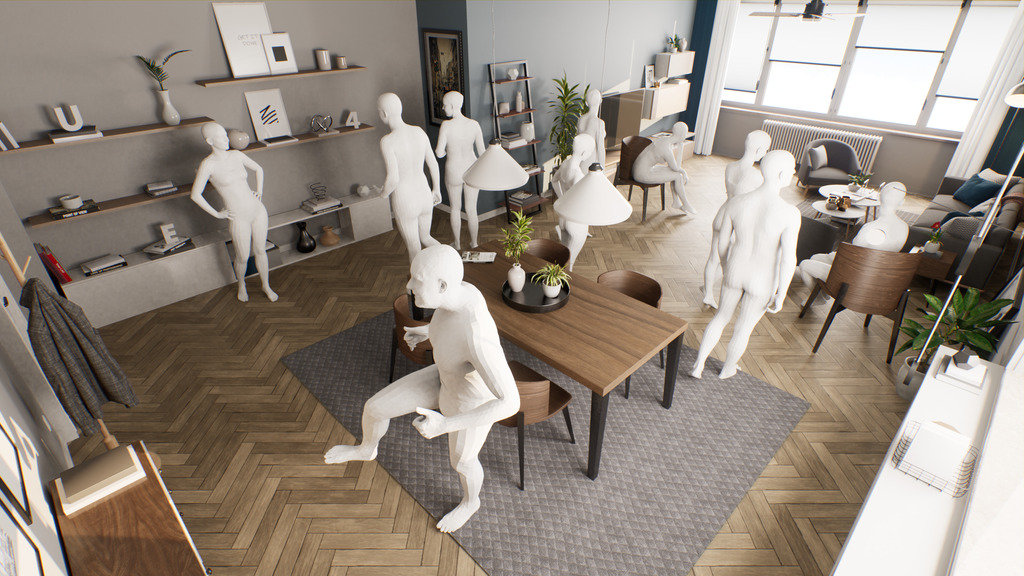}		
		\includegraphics[width=0.32\textwidth]{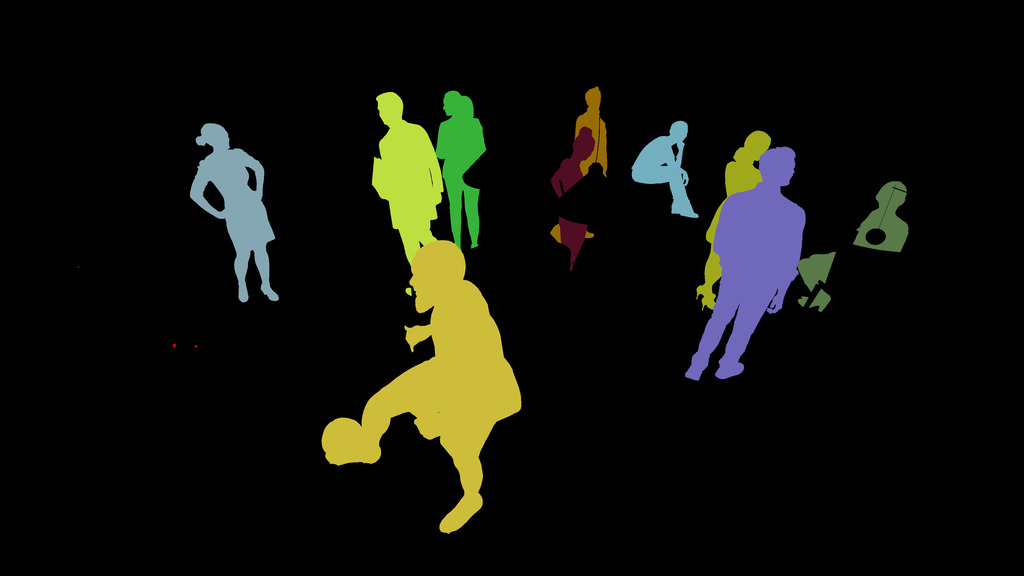}		
	}
	\centerline{\includegraphics[width=0.32\textwidth]{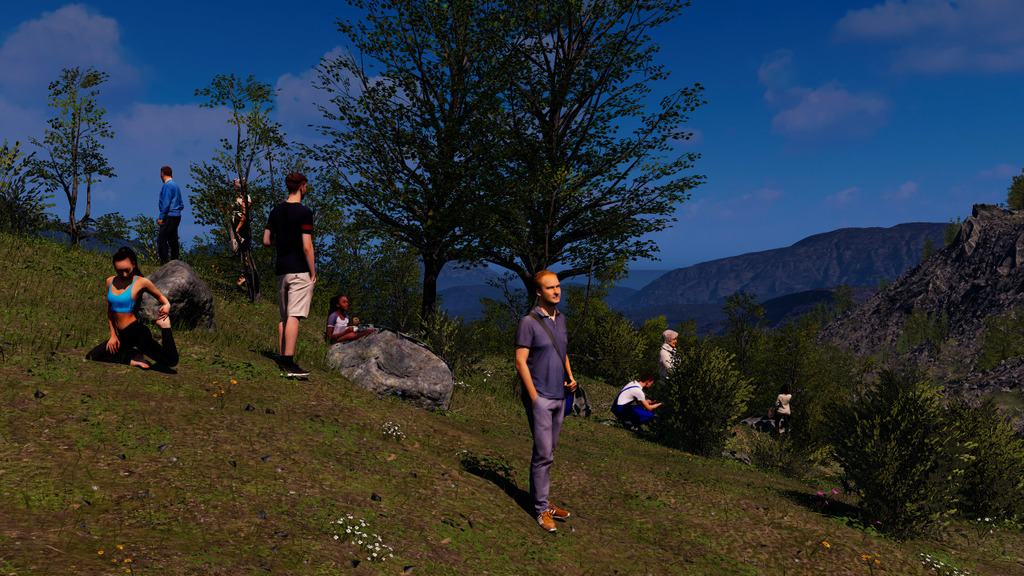}		
		\includegraphics[width=0.32\textwidth]{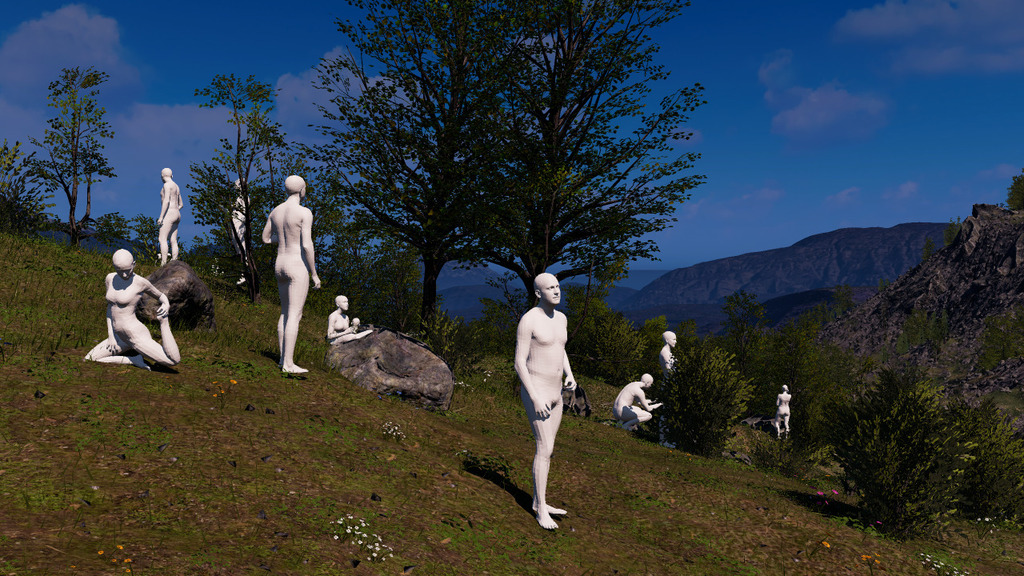}		
		\includegraphics[width=0.32\textwidth]{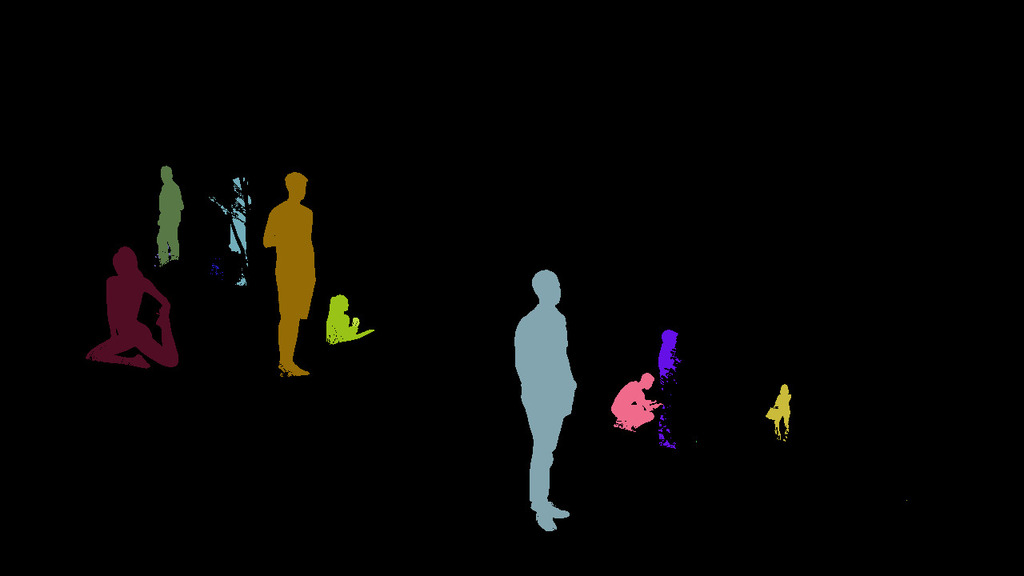}		
	}
	\centerline{\includegraphics[width=0.32\textwidth]{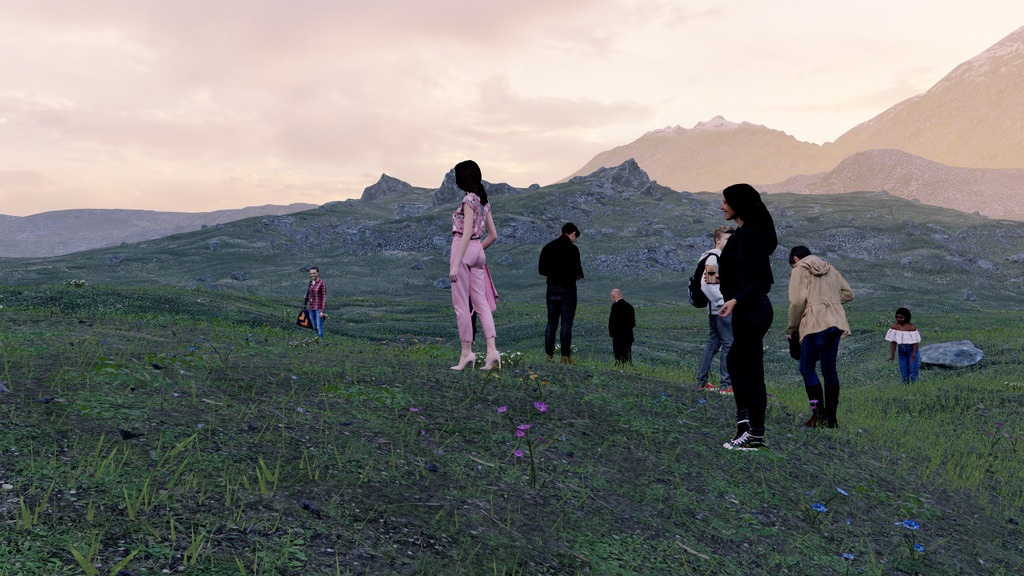}		
		\includegraphics[width=0.32\textwidth]{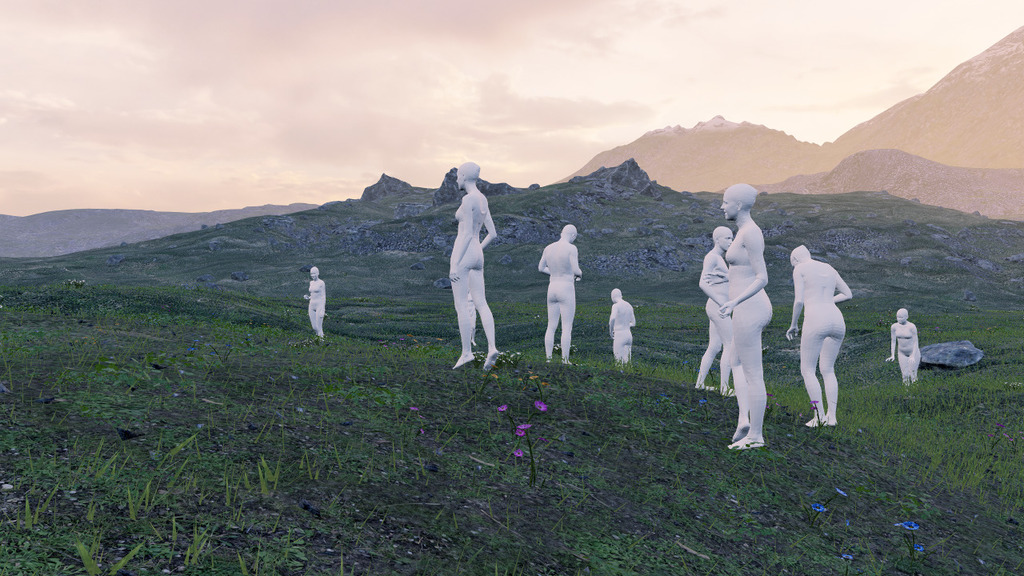}		
		\includegraphics[width=0.32\textwidth]{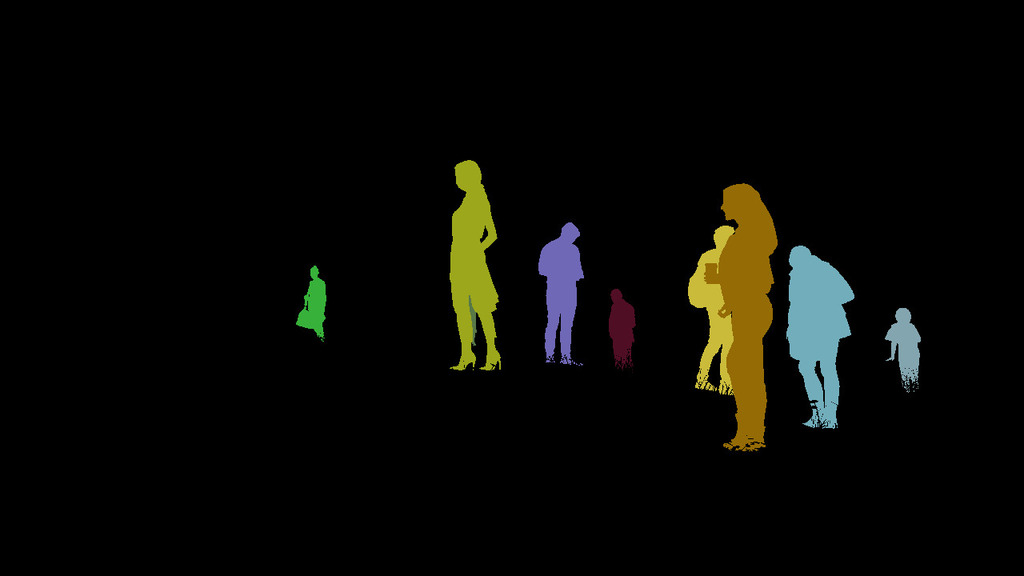}		
	}
	\centerline{\includegraphics[width=0.32\textwidth]{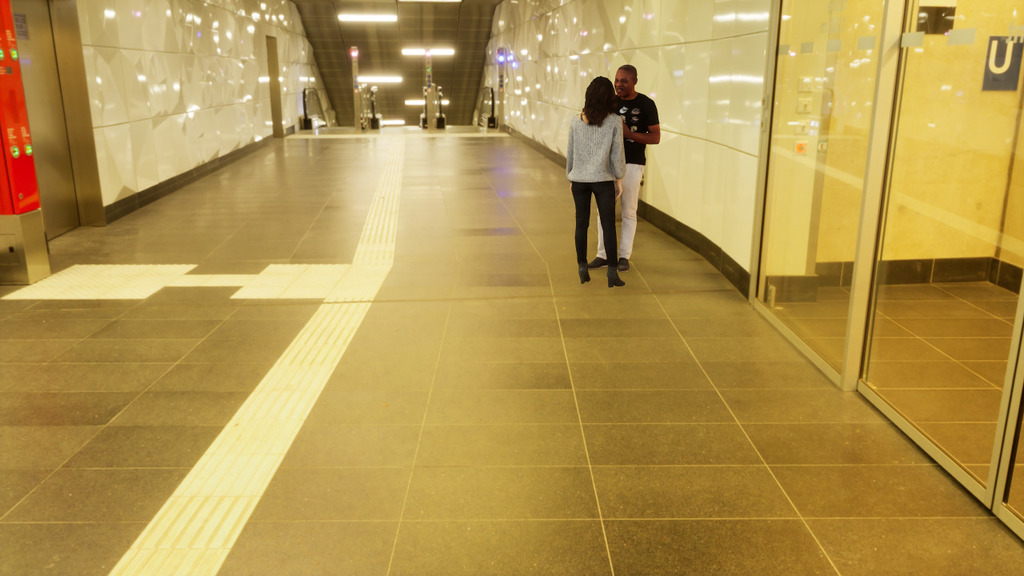}		
		\includegraphics[width=0.32\textwidth]{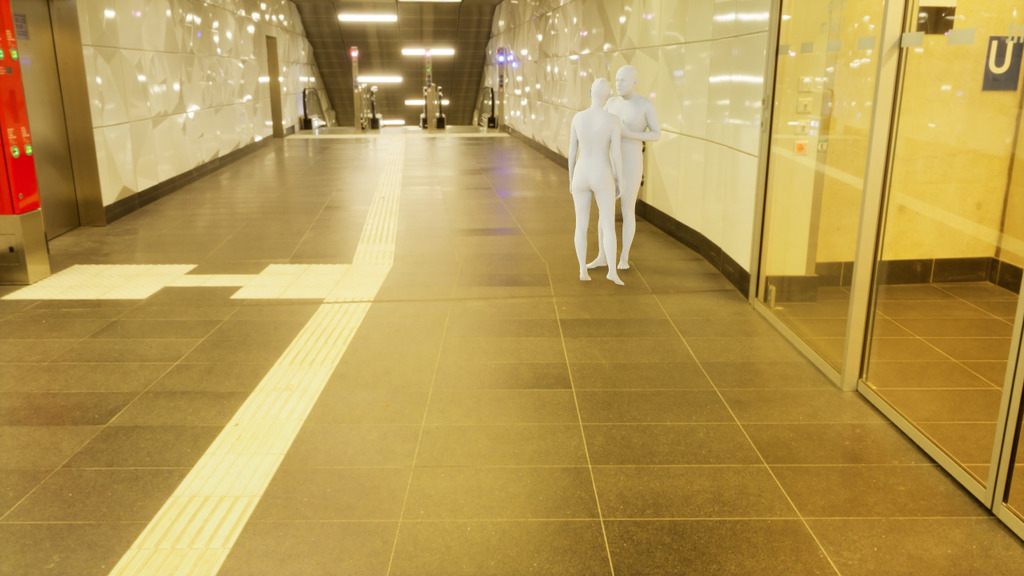}		
		\includegraphics[width=0.32\textwidth]{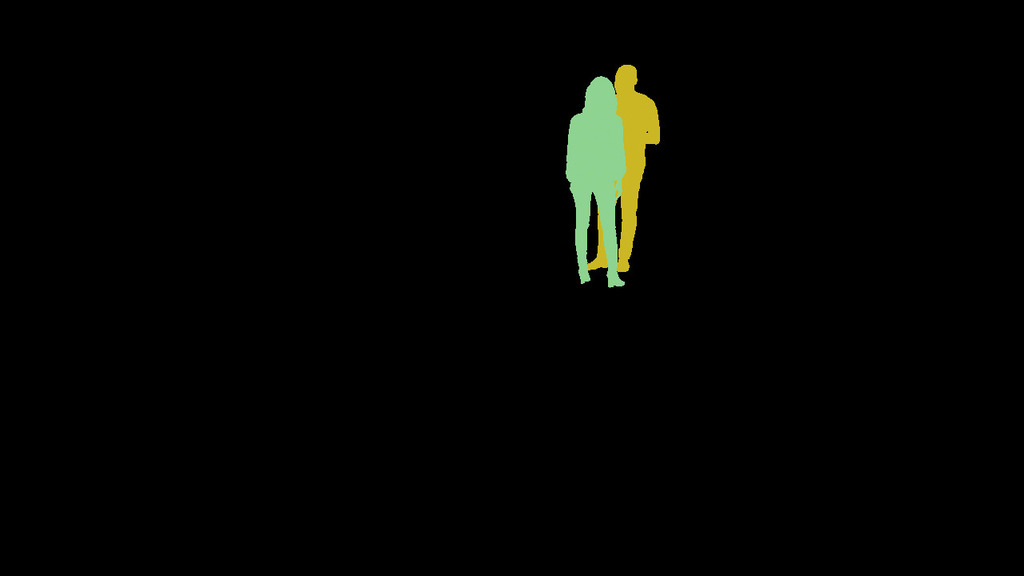}		
	}

	\caption{Examples images from the {\dbname} dataset.
}
	\label{fig:teaser}
\end{figure*}

\noindent\textbf{AGORA Statistics. }
We provide the dataset distribution across various attributes i.e.~age, ethnicity and gender for AGORA ground truth scans in Fig.~\ref{fig:statistics}. 
AGORA has evenly distributed gender and a varied range of age and ethnicity.
To create this distribution, we use gender, age and ethnicity information provided by Renderpeople~\cite{renderedpeople}. For other vendors since no age and ethnicity information was given we label them with the help of Amazon Mechanical Turks~\cite{mturks}. We recruit 5 different subjects with $>$ 5000 HITs approved and an approval rate $>$97\%. 
We ask them to classify the rendered image of the scan into predefined categories of ethnicity and age (including Unknown). If there is a majority vote, we label the scan with the respective category. If there is a tie, we resolve it by ourselves by selecting the best estimate or by selecting Unknown. For others, we marked them Unknown. We label all the scans with gender ourselves.

\noindent\textbf{AGORA Dataset. }
Fig.~\ref{fig:teaser} provides more examples of our dataset. 
From left to right we show the RGB image, the RGB image with ground truth SMPL-X fits and segmentation masks. 
The 3D scenes (row 1-4) lead to challenging environmental occlusion, as can be seen in the segmentation masks on the right.
Row 5 shows a example from the easy-split experiment described in Sec.~\ref{easy-split}.
We also render individual subject masks as though there is no occlusion. 
See Fig.~\ref{fig:indi-mask} for examples. We use these masks to determine how much a person is occluded.

	\section{Additional Analysis on Baseline Experiment}

Here we provide details for the Evaluation Protocol as well as additional analysis of the baselines.

\subsection{Evaluation Protocol.}

In the following we describe our evaluation protocol in detail.
A less detailed description with visualization can be found in the main paper in Section \ref{eva-protocol}.
We evaluate the following methods: ExPose~\cite{choutas2020monocular}, FrankMocap~\cite{rong2020frankmocap}, SMPLify-X \cite{SMPL-X:2019}, HMR \cite{kanazawa2018end}, SPIN \cite{kolotouros2019spin}, EFT \cite{joo2020exemplar}, and CenterHMR ~\cite{sun2020centerhmr}. which collectively provide a good picture of the current SOTA.
Our protocol can be split into four parts: 1) Detection, 2) 3D pose and shape estimation, 3) matching of predictions to ground truth and 4) computing errors. In the following we will explain each component.

\noindent\textbf{Detection. }
All the methods we evaluate require the person to first be localized.  
To obtain person detections we run OpenPose \cite{cao2019openpose} on the input image. 
OpenPose requires a large number of settings. To select these settings we draw inspiration from the maximum accuracy setting as reported on the OpenPose GitHub page\footnote{\url{https://github.com/CMU-Perceptual-Computing-Lab/openpose}}.
However, these settings have huge memory requirements on the GPU.
Thus, we modify the settings such that OpenPose can run on a common GPU with 12 GB of memory. 
We end up with the following settings: We scale the larger side of the input images to 272 pixels while keeping the aspect ratio fixed. We use two scale processing for the body keypoints with a scale gap of 0.25. We run the face detection network with default settings.
Note that only SMPLify-X makes use of hand and face detections, however, we use the same settings for all the evaluated methods to reduce influence of the keypoint detection on the results.
For CenterHMR~\cite{sun2020centerhmr} we directly use the entire image as input without any cropping.

\noindent\textbf{3D pose and shape estimation. }
For each OpenPose detection we run all the methods.
All methods in our experiment except~\cite{sun2020centerhmr} require either 2D keypoints or tight crops around the detected person as input. 
For ExPose~\cite{choutas2020monocular}, FrankMocap~\cite{rong2020frankmocap},HMR \cite{kanazawa2018end} and SPIN \cite{kolotouros2019spin} we use their demo code to generate crops from OpenPose detections. SMPLify-X \cite{SMPL-X:2019} operates directly on OpenPose keypoints and no  pre-processing is needed. EFT considers both the image feature of the crop and keypoints. For CenterHMR~\cite{sun2020centerhmr} we directly use the entire image as input for 3D pose and shape estimation.

\noindent\textbf{Matching predictions to ground truth.}
Let $M$ be the set of predicted meshes and $N$ be the set of ground truth humans.
To match the predictions of the methods to ground truth, we project the 3D keypoints of the estimated SMPL-body to the image plane. To this end, the camera parameters as assumed or estimated by the method are used. Similarly, we project the ground truth 3D keypoint to the image plane using the ground truth camera parameters. 
We compute the 2D joint error for all combinations of $m \in M$ and $n\in N$ and match them based on minimal 2D keypoint error. 

\noindent\textbf{False positives and false negatives.}
The predictions are often noisy leading to \emph{false positives} and \emph{false negatives}, as shown in Fig.~3 of the main paper.
It is crucial to detect false positives to avoid that a false positive is incorrectly matched to ground truth, resulting in large errors, which might distort the results.
To detect false positives we construct 2D axis-aligned bounding boxes (AABB) for each prediction and ground truth based on the 2D keypoints.
Before matching a prediction with ground truth we compute the intersection over union (IoU) for this pair. If $\text{IoU} < \tau$ and no other possible match exists for a given prediction it is considered a false positive. We choose our threshold $\tau=0.1$, such that predictions which have a large distance in 2D are not considered to match, but small enough to ensure that differences in the scale of prediction and ground truth do not lead to erroneous classification as a false positive.
Finally, each unmatched ground truth body is considered as a false negative.

\noindent\textbf{Computing errors. }
We compute the errors using different metrics, as described in Section \ref{datset-metric} in the main paper.

\begin{figure*}[t!]

	\centerline{\includegraphics[width=0.33\textwidth]{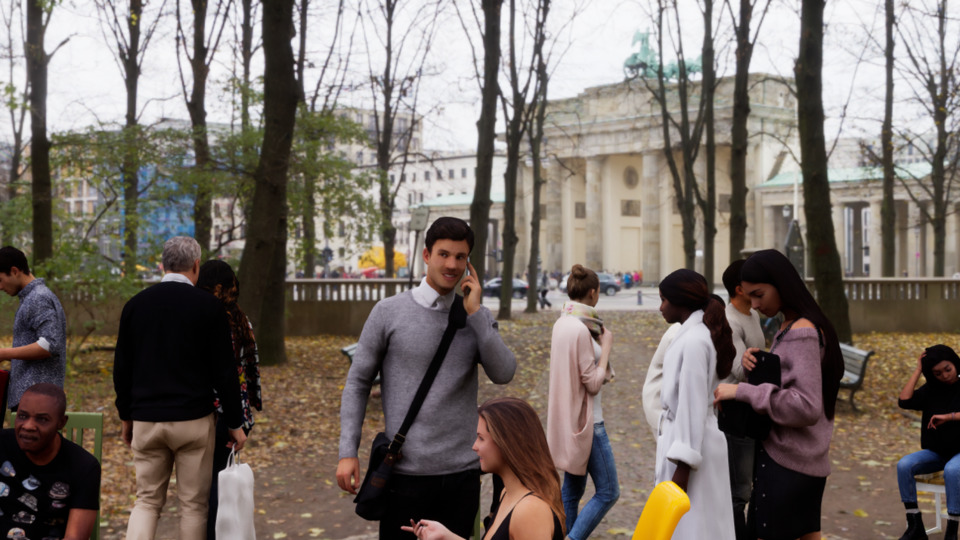}
		\includegraphics[width=0.33\textwidth]{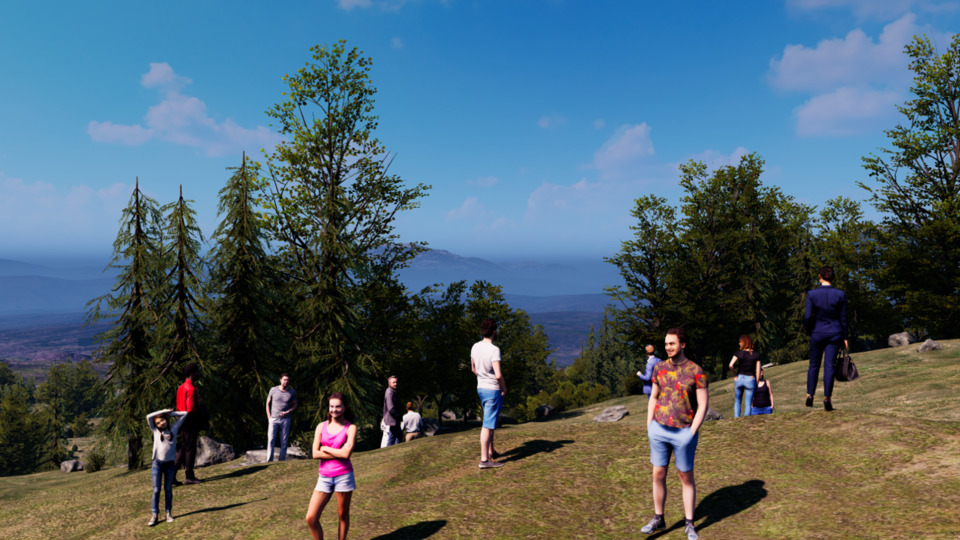}
		\includegraphics[width=0.33\textwidth]{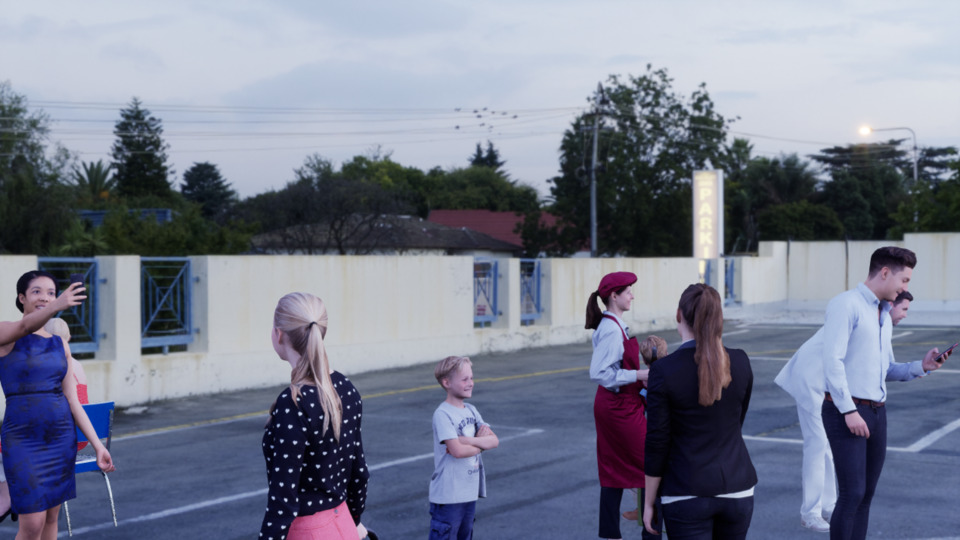}
	}
	\centerline{\includegraphics[width=0.33\textwidth]{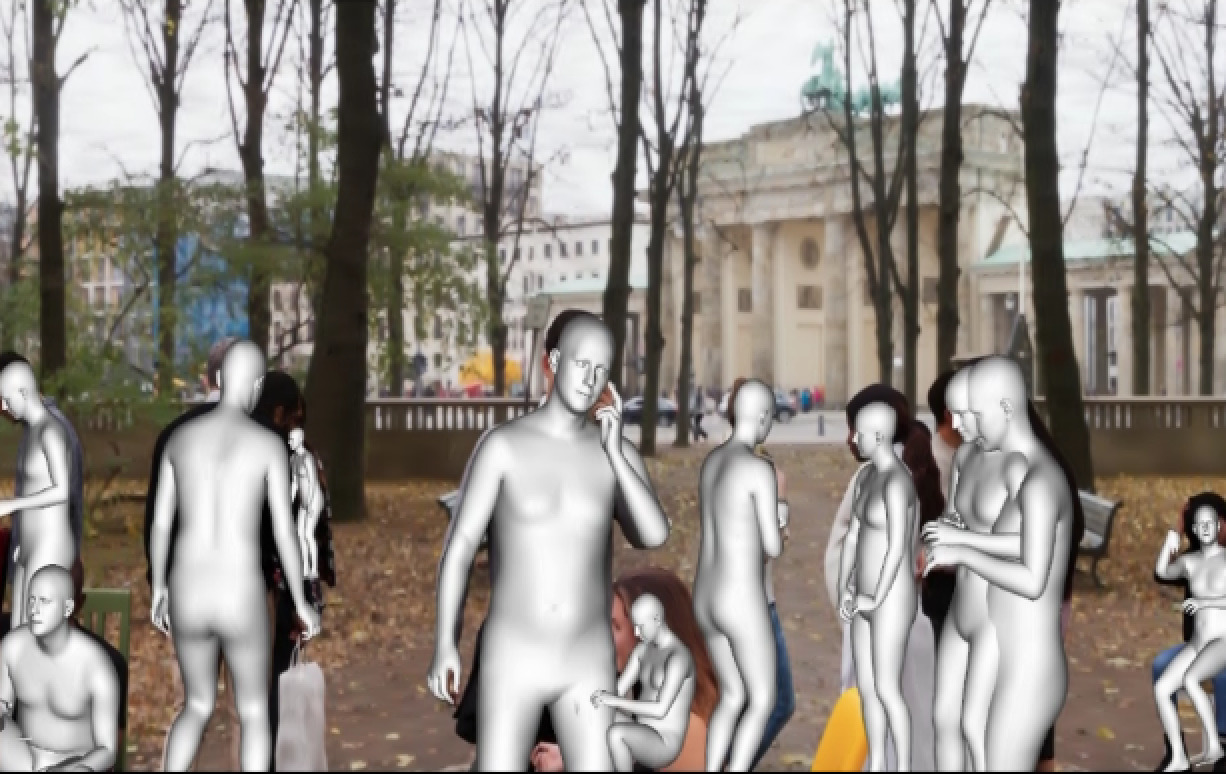}
		\includegraphics[width=0.33\textwidth]{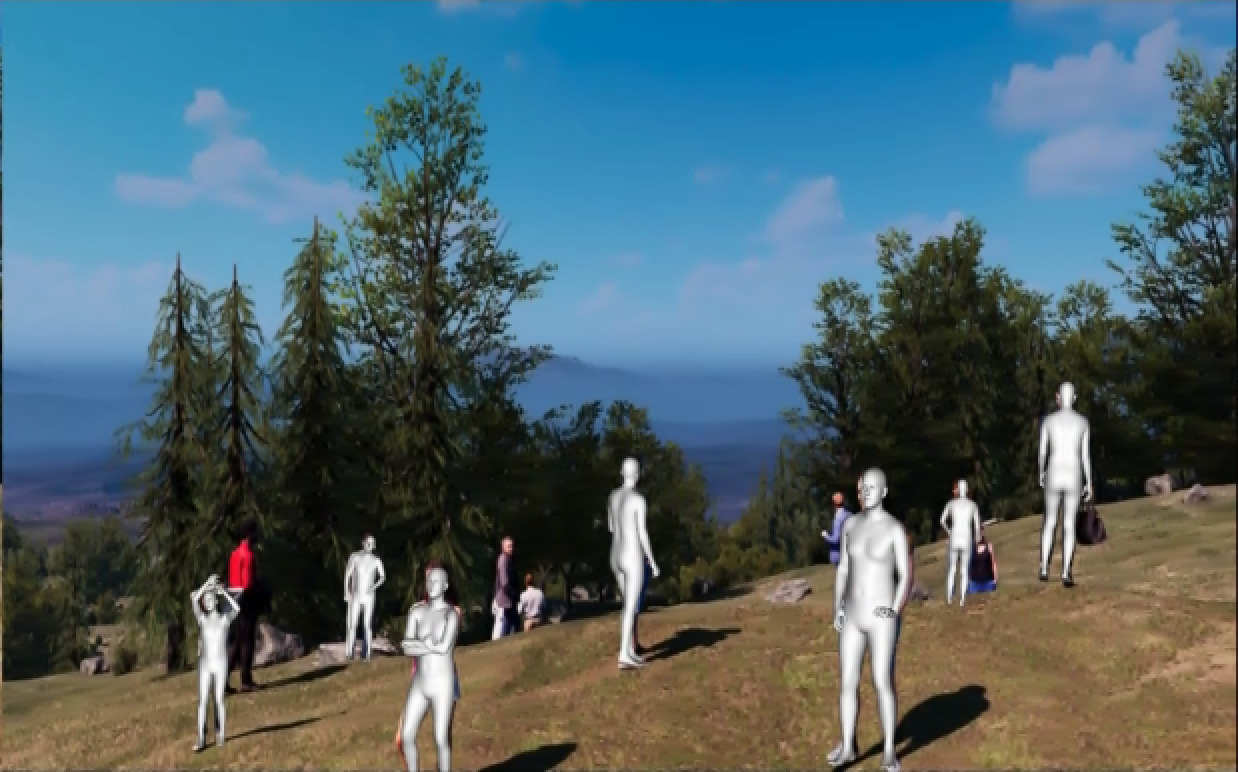}
		\includegraphics[width=0.33\textwidth]{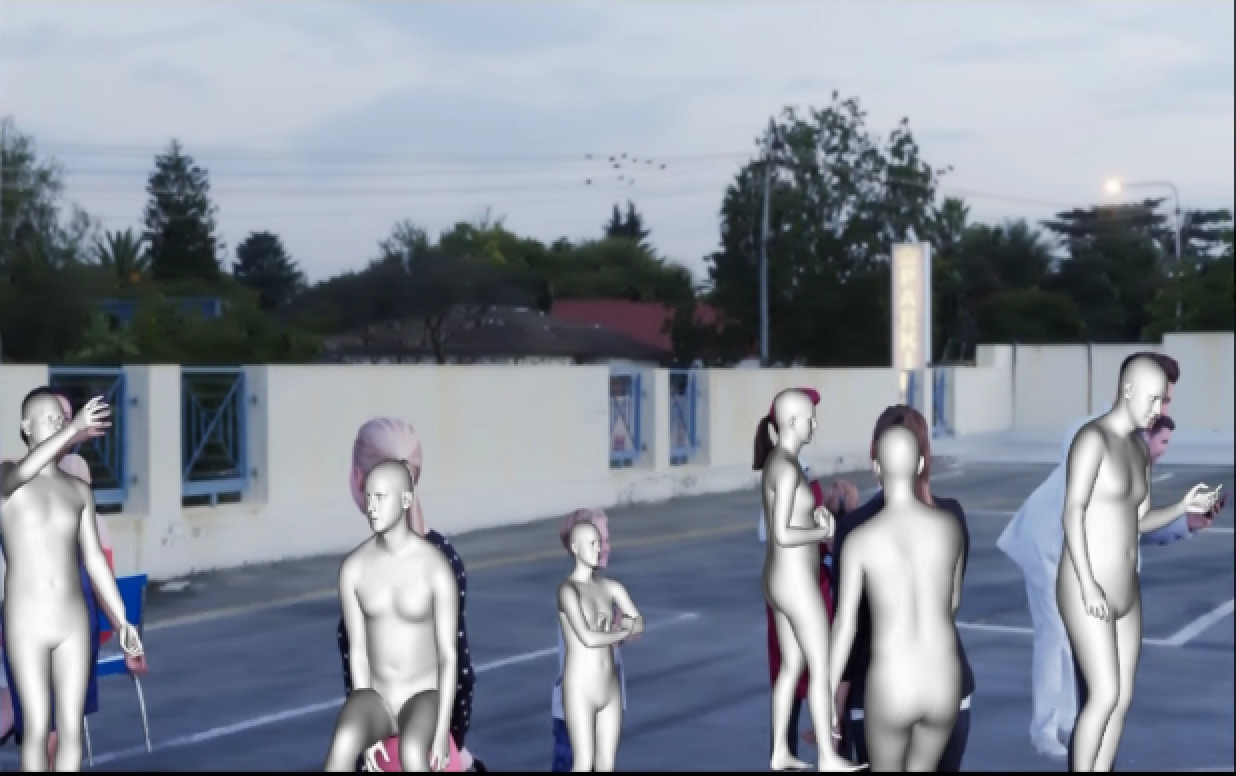}
	}
	\centerline{\includegraphics[width=0.33\textwidth]{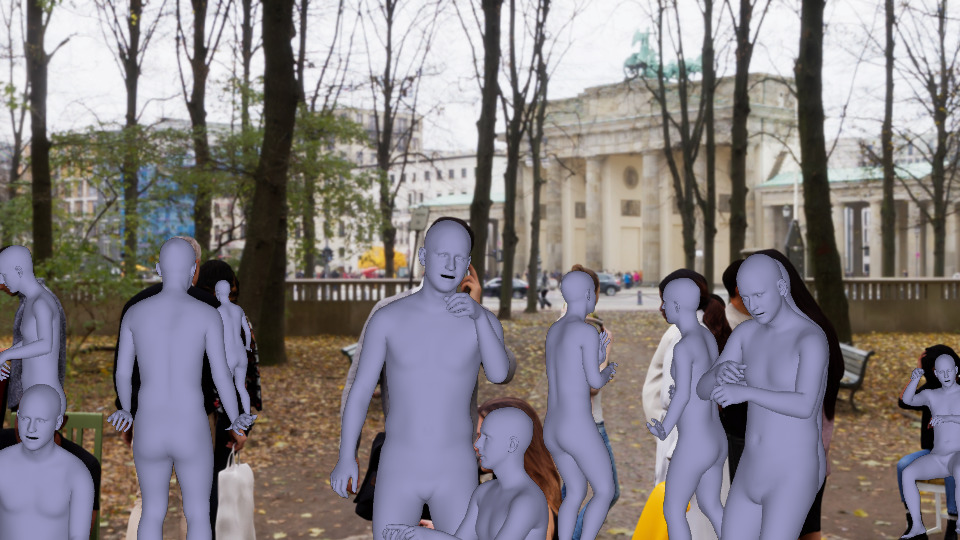}
		\includegraphics[width=0.33\textwidth]{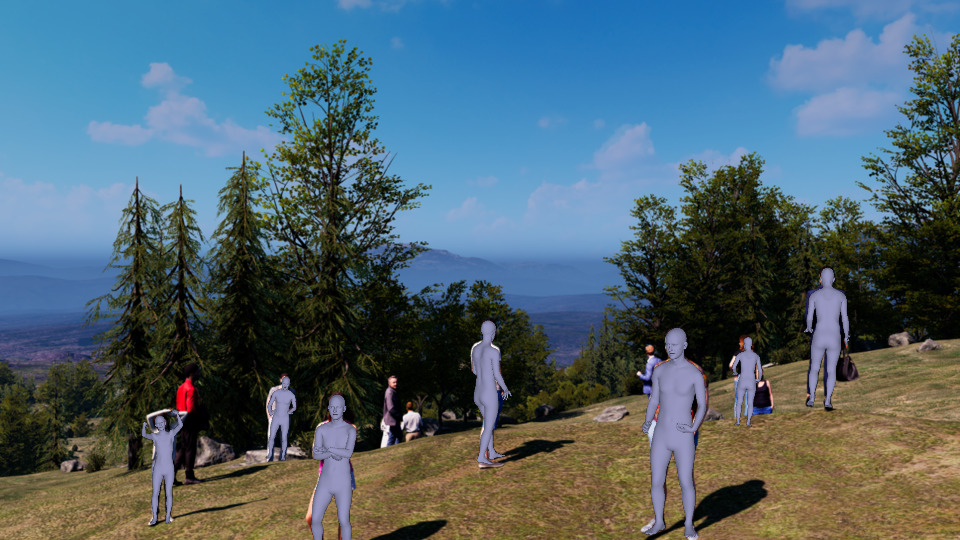}
		\includegraphics[width=0.33\textwidth]{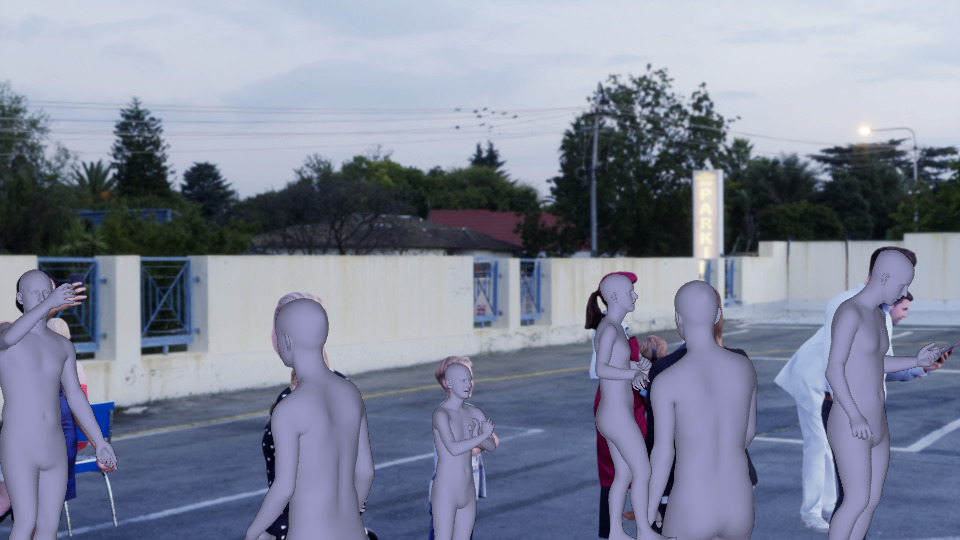}
	}
	\centerline{\includegraphics[width=0.33\textwidth]{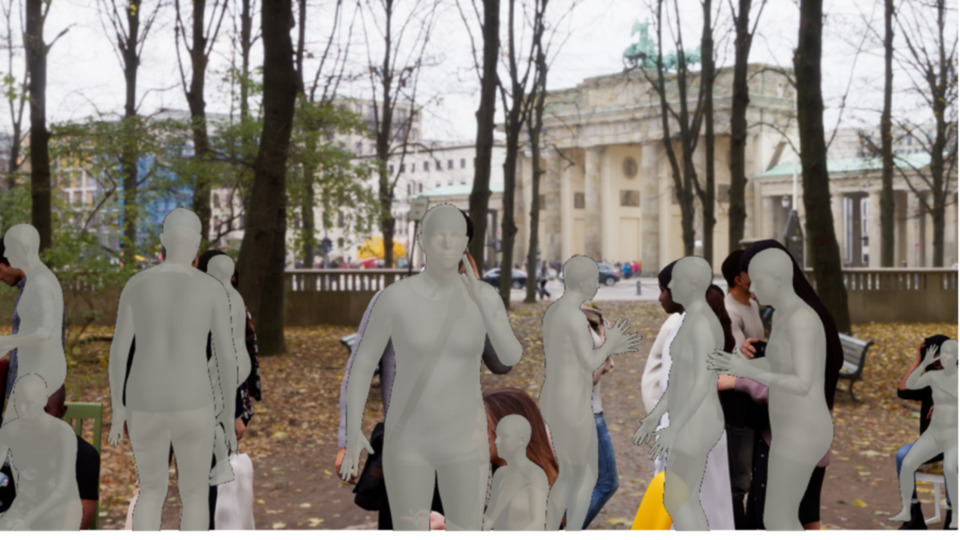}
		\includegraphics[width=0.33\textwidth]{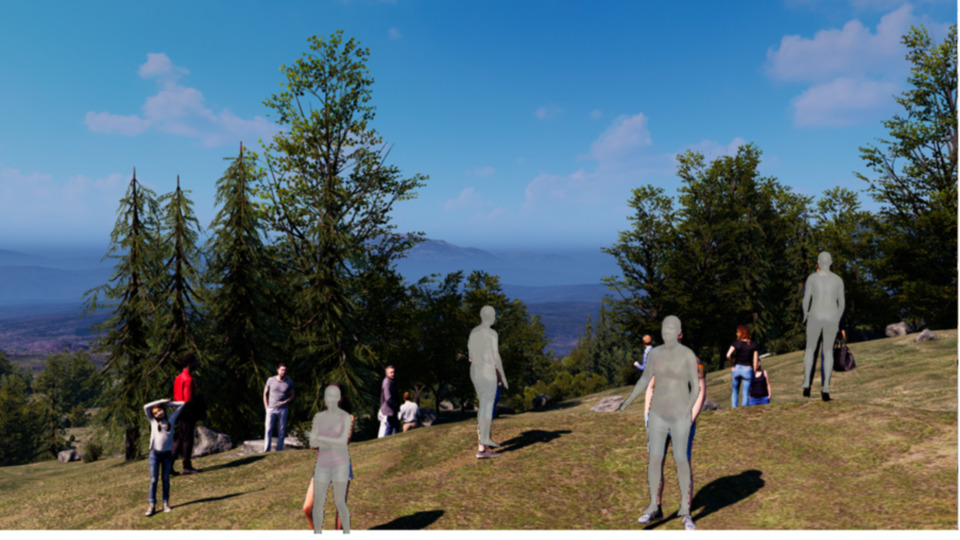}
		\includegraphics[width=0.33\textwidth]{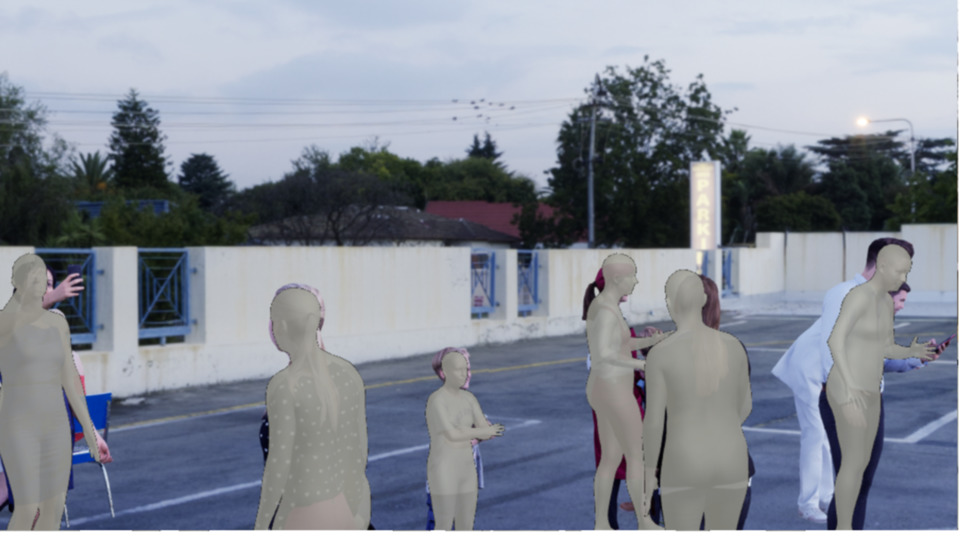}
	}
	\centerline{\includegraphics[width=0.33\textwidth]{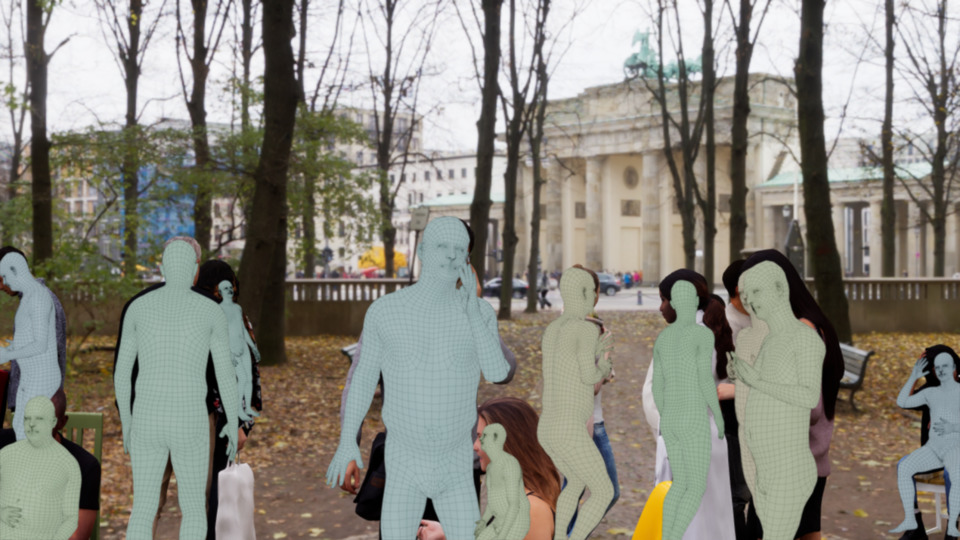}
		\includegraphics[width=0.33\textwidth]{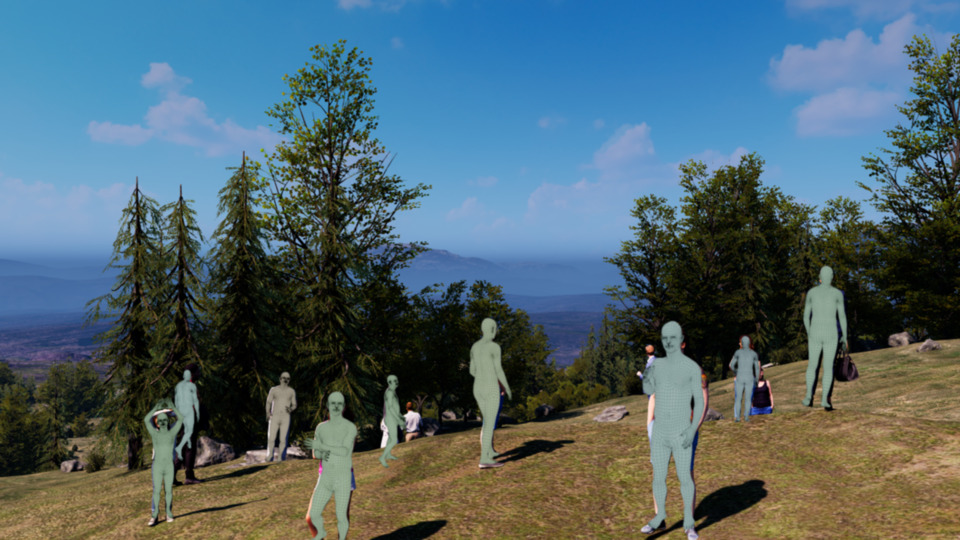}
		\includegraphics[width=0.33\textwidth]{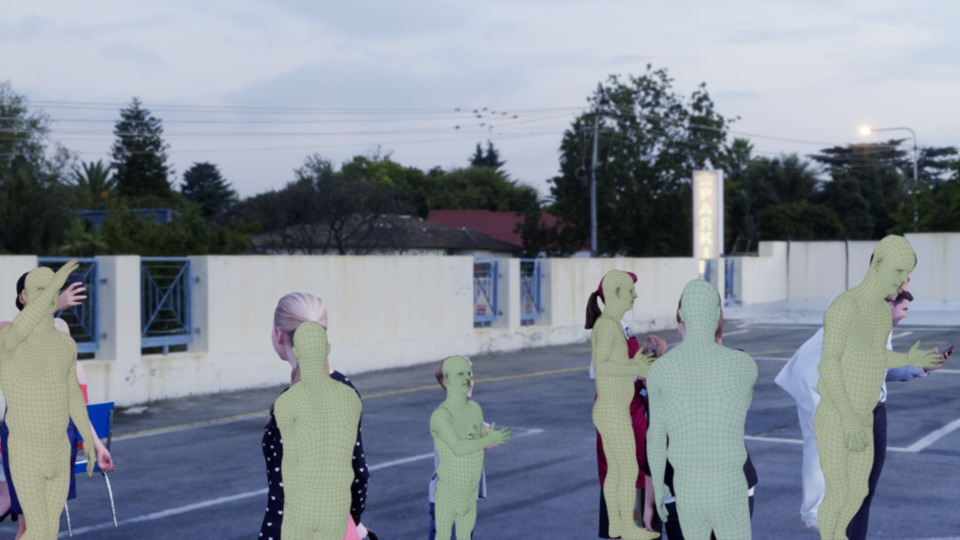}
	}
	\centerline{\includegraphics[width=0.33\textwidth]{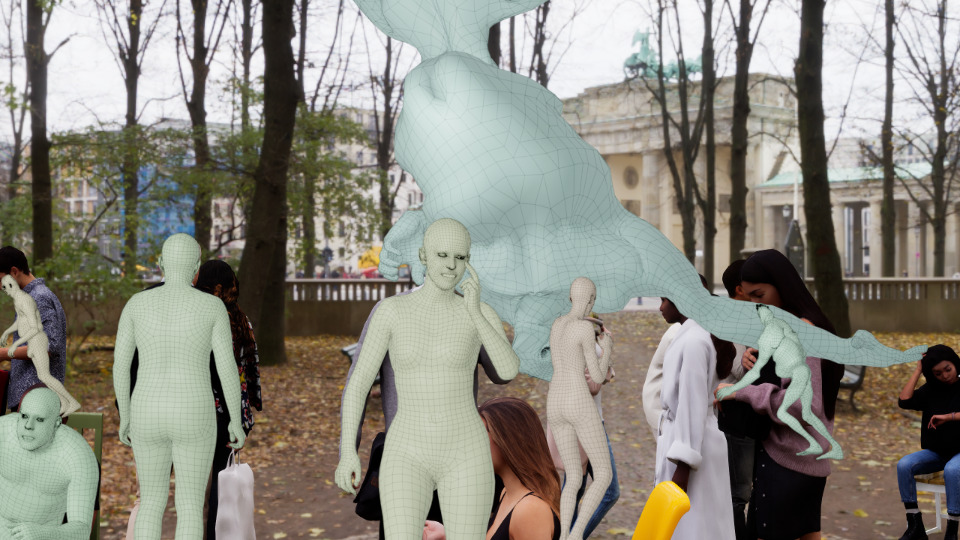}
		\includegraphics[width=0.33\textwidth]{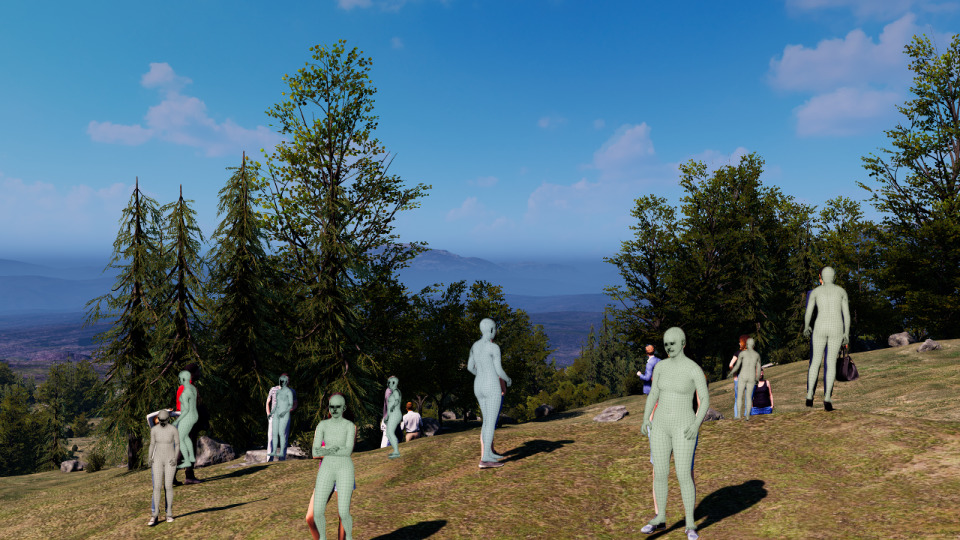}
		\includegraphics[width=0.33\textwidth]{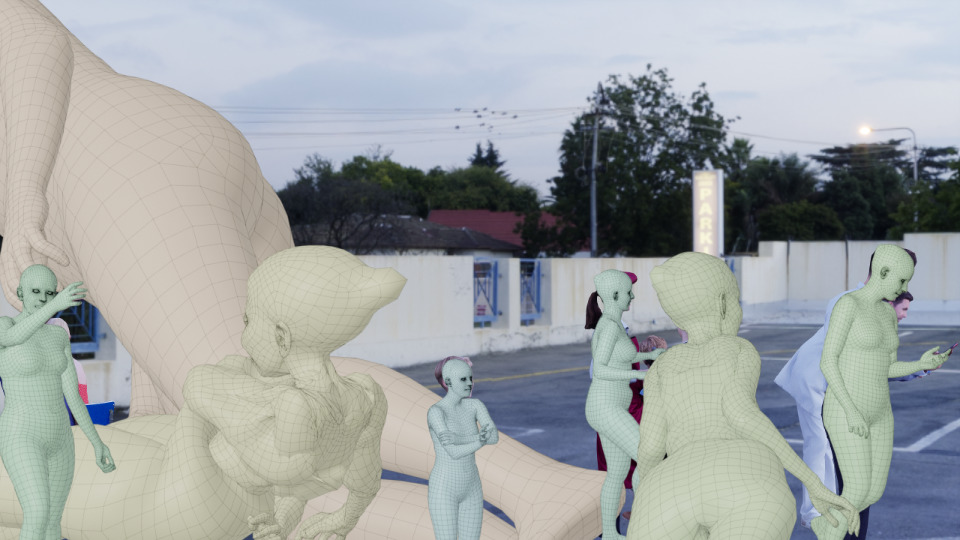}
	}
	\centerline{\includegraphics[width=0.33\textwidth]{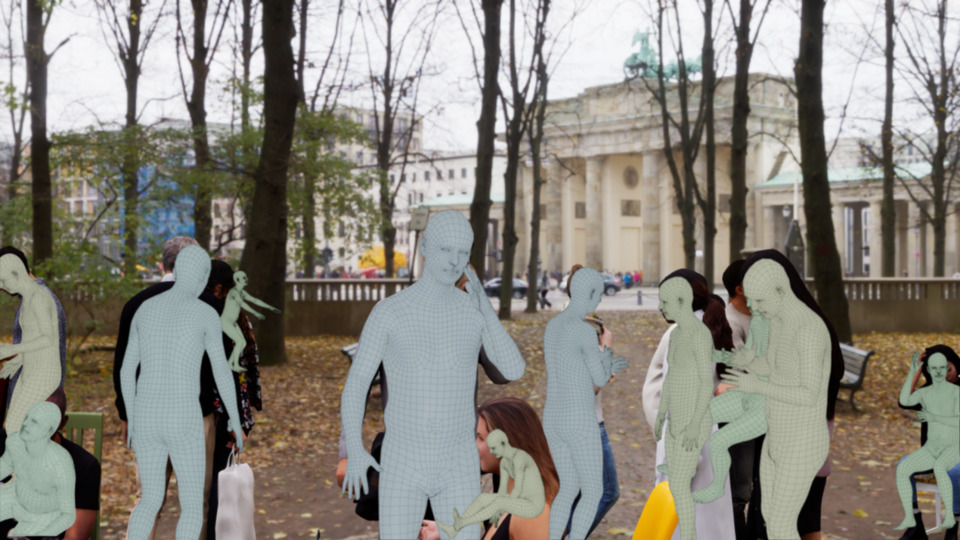}
		\includegraphics[width=0.33\textwidth]{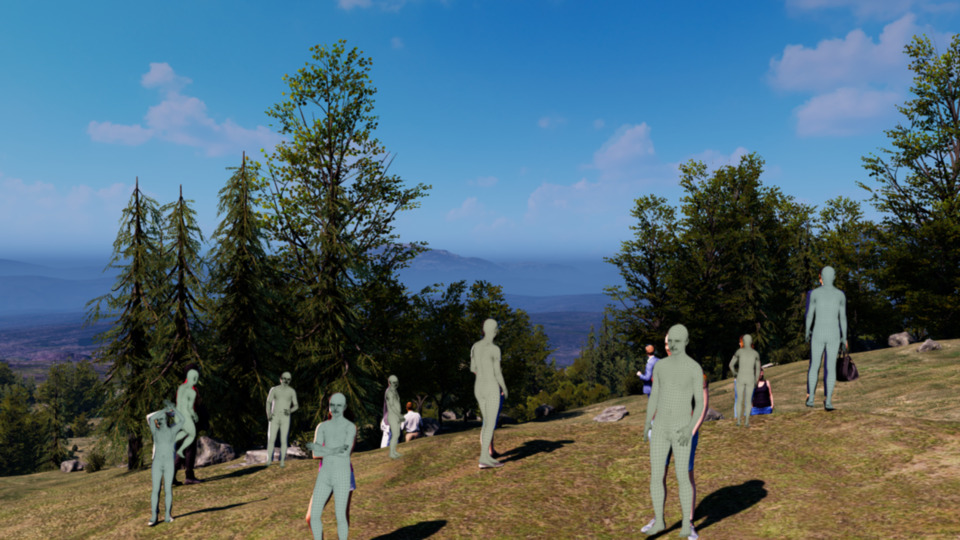}
		\includegraphics[width=0.33\textwidth]{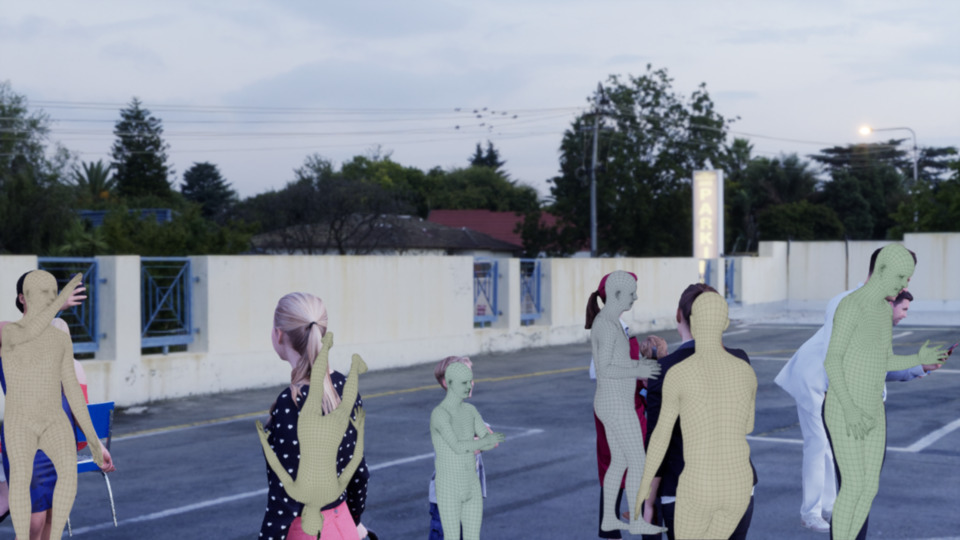}
	}
\vspace{0.1cm}
	
	\caption{Method evaluation. RGB images (row 1), FrankMocap (row 2), ExPose (row 3), CenterHMR (row 4), HMR (row 5), SMPLify-X (row 6) and SPIN (row 7).}
	\label{fig:baseline-visual}
	
\end{figure*}
\subsection{Evaluation of Methods}
A qualitative comparison of different methods is shown in Fig.~\ref{fig:baseline-visual}. 
In Sec.~\ref{baselines} of the main paper we analyse the dependence of errors with respect to occlusion and distance from the center of the image. Here, we analyze the dependence of errors on the body orientation.

\noindent\textbf{Orientation.}
We concentrate particularly on the yaw rotation with respect to the camera, and report the error in Fig.~\ref{fig:Baseline-orientation}, where $0^\circ$ corresponds to facing the camera. 
We observe that the error grows as the yaw angle increases, reaching the peak around $180^\circ$ and then decreases.
This suggests that the current 3D human-pose-and-shape methods perform worst when subjects are not facing the camera.

\begin{table}[t!]
	\scriptsize
	\resizebox{\columnwidth}{!}{
		\begin{tabular}{l|cc|cc}
			\toprule
			\multicolumn{1}{c|}{Models} & \multicolumn{2}{c|}{3DPW (14 joints) }&         
			\multicolumn{2}{c}{3DPW (24 joints)} \\
			\midrule
			& \multicolumn{1}{l}{\footnotesize MPJPE $\downarrow$} & \multicolumn{1}{l|}{\footnotesize PA-MPJPE $\downarrow$} & 
			\multicolumn{1}{l}{\footnotesize MPJPE $\downarrow$} & \multicolumn{1}{l}{\footnotesize PA-MPJPE $\downarrow$} \\
			\midrule
			Human3.6M \cite{h36m_pami}  &311.3&162.1 &286.2&178.1 \\
			\text{[MPII+LSPet+COCO]\textsubscript{EFT} \cite{joo2020exemplar}} & 125.0 & 77.4 & 121.9 & 86.1 \\
			AGORA & 147.4 & 81.0 & 141.3 & 88.8 \\
			\bottomrule															
		\end{tabular}
	}
	\caption{Training SPIN from scratch with Human3.6M vs. EFT vs. AGORA.}
	\label{tab:pretrain-comparison}
\end{table}
\noindent\textbf{From scratch training.}
In Table \ref{tab:pretrain-comparison} we report results for SPIN trained from scratch using different datasets.
For AGORA training, we report on-par PA-MPJPE compared to [MPII+LSPet+COCO]\textsubscript{EFT} but much better results than for training from scratch with Human3.6M \cite{h36m_pami}.
These experiments suggest that the AGORA training set is sufficiently realistic and large to support both finetuning and from-scratch training.
Pre-trained weights on AGORA will be made available for research purposes.

\noindent\textbf{Easy split experiment.}\label{easy-split}
To further show that the images in AGORA are comparable in complexity
to natural images, we create an easy test set of approximately 400 images as sanity check.  Each
image consists of only two people in the easy split, potentially with
some minor occlusion (5th row of Fig.~\ref{fig:teaser}). Despite having two person per image and minor occlusion, easy test set is still more challenging than 3DPW~\cite{vonMarcard20183dpw} because of varied lighting and complex clothing. We evaluate
FrankMocap~\cite{rong2020frankmocap} and ExPose on this set and report 118.0 and 111.6mm B-MPJPE error for 22 SMPL-X joints. ExPose error is comparable to
the reported error on 3DPW in original
work~\cite{choutas2020monocular} as 93.4. Since, FrankMocap only shows qualitative results, we can't compare the quantitative evaluation with original work.
The easy-split experiment along with the finetuning experiment
described in main paper suggest that our synthetic images are on par with natural images.

\end{document}